\documentclass[10pt,letterpaper]{article}

\usepackage{arxiv}
\usepackage{fullpage}
\usepackage[T1]{fontenc}
\usepackage[utf8]{inputenc}
\usepackage{charter}
\usepackage{microtype}
\usepackage[round,authoryear]{natbib}

\usepackage{amsmath,amsfonts,bm}
\newcommand{\fbestt}{f_{\textrm{best}_t}}
\newcommand{\fbest}{f_{\textrm{best}}}

\def\eqref#1{equation~\ref{#1}}

\def\1{\bm{1}}

\DeclareMathAlphabet{\mathsfit}{\encodingdefault}{\sfdefault}{m}{sl}
\SetMathAlphabet{\mathsfit}{bold}{\encodingdefault}{\sfdefault}{bx}{n}

\DeclareMathOperator*{\argmax}{arg\,max}
\DeclareMathOperator*{\argmin}{arg\,min}

\usepackage{mathtools}
\usepackage{graphicx}
\usepackage{xcolor}
\usepackage{xspace}
\usepackage{booktabs}
\usepackage{array,longtable,multirow}
\usepackage{float}
\usepackage{placeins}
\usepackage{algorithm}
\usepackage{algorithmicx}
\usepackage{algpseudocode}
\usepackage{titlesec}
\usepackage[font=small,labelfont=bf]{caption}
\usepackage{xurl}
\usepackage[hidelinks]{hyperref}
\usepackage{bookmark}

\hypersetup{
  pdftitle={Which LLM to pick? Online Active Model Selection for Large Language Models},
  pdfauthor={},
  pdfsubject={Online active model selection for large language models}
}

\titlespacing*{\section}{0pt}{12pt}{6pt}
\titlespacing*{\subsection}{0pt}{8pt}{4pt}
\titlespacing*{\subsubsection}{0pt}{7pt}{3pt}

\newcommand{\framework}{\textsc{Online LLM Picker}\xspace}

\title{Which LLM to pick?\\[0.3em]
Online Active Model Selection for Large Language Models}

\author{
  Alessandro Turrin \\
  \textnormal{TU Delft} \\
  \texttt{a.turring@tudelft.nl}
  \And
  Patrik Okanovic \\
  \textnormal{ETH Zurich} \\
  \texttt{patrik.okanovic@inf.ethz.ch}
  \And
  Torsten Hoefler \\
  \textnormal{ETH Zurich} \\
  \texttt{torsten.hoefler@inf.ethz.ch}
  \And
  Nezihe Merve Gürel \\
  \textnormal{TU Delft} \\
  \texttt{n.m.gurel@tudelft.nl}
}
\date{}

\begin{document}

\maketitle

\begin{abstract}
Large Language Models (LLMs) are increasingly applied to process streaming data, with practitioners relying on benchmarks to select the best model even though these signals only approximate real performance. 
While oracle annotations can provide reliable feedback, they are often costly and difficult to obtain at scale. 
To address this challenge, we propose \framework, the first framework for active model selection for LLMs in online settings. 
Given an arbitrary stream of queries and a limited annotation budget, \framework  selects the most informative prompts for annotation to identify the best LLM among candidate models. Across multiple tasks including 10 datasets, for over 130 language models, we show that \framework saves annotation cost by up to $71.67$\% while reliably identifying the best or near-best model for the stream.
We also show that using the returned model for sequential generation on unannotated prompts across the stream reduces regret by up to a factor of $2.51\times$, indicating that \framework can identify the best or near-best model well before processing all streaming prompts.
\end{abstract}

\section{Introduction}
Large Language Models (LLMs) are widely applied across various domains \citep{fan_bibliometric_2023, tan_unstructured_2023}, from medicine \citep{boll_distillnote_2025, veen_adapted_2024, biswas_intelligent_2024} and industry \citep{angelopoulos_industrial_2025, kok_when_2024, li_large_2024}, to education \citep{dan_educhat_2023, caines_application_2023} and law \citep{pont_legal_2023, lee_lexgpt_2023}, \citep{guha_legalbench_2023}.
Alongside this growing adoption, hundreds of publicly available LLMs have emerged \citep{google-t5_google-t5t5-small_2024, meta-llama_meta-llamallama-31-8b_2024, falcon_tiiuaefalcon-7b-instruct_2023, deepseek_deepseek-aideepseek-v31_2025, mistralai_mistralaimistral-7b-instruct-v03_nodate, qwen_qwenqwen3-next-80b-a3b-instruct_2025, nvidia_nvidianvidia-nemotron-nano-9b-v2_2025}, many demonstrating strong performance without the need for fine-tuning and the ability to follow instructions and adapt to new tasks with little or no task-specific supervision \citep{brown_language_2020, kojima_large_2023, dong_survey_2024, liu_pre-train_2021}.

Given the abundance of off-the-shelf models, selecting the most suitable LLM for a particular task or data stream is a non-trivial problem. Existing evaluation pipelines and benchmarks assess performance across diverse datasets and tasks using multiple metrics \citep{guo_evaluating_2023}, but no single model consistently excels across domains or datasets \citep{chang_survey_2023, liang_holistic_2023}. Because LLM effectiveness is highly context-dependent and varies substantially across scenarios, standard evaluation metrics alone cannot fully capture model quality or practical utility \citep{ouyang_training_2022, kocon_chatgpt_2023}. A common approach to model selection is therefore to rely on randomly or heuristically chosen small subsets of annotated data \citep{polo_tinybenchmarks_2024, vivek_anchor_2024}. However, such strategies often lead to inefficient use of resources and fail to reliably capture differences across models \citep{kossen_active_2021}. Active Model Selection \citep{madani_active_2012, karimi_online_2021, ashury-tahan_label-efficient_2024, okanovic_all_2024, kay_consensus-driven_2025, liu_contextual_2022, li_online_2024, hara_active_2024, gardner_bayesian_2015} addresses this limitation by selectively annotating a small set of queries to identify the best model for arbitrary data examples. Yet, prior work has largely focused on classification tasks rather than generation \citep{karimi_online_2021, liu_contextual_2022, okanovic_all_2024, kay_consensus-driven_2025, li_online_2024, hara_active_2024}, with the exception of \cite{ashury-tahan_label-efficient_2024}, which studies language model selection under the assumption that all prompts are available from the outset. To the best of our knowledge, no prior work has explored active model selection for LLMs in online settings to date.

\begin{figure}[t]
    \centering
    \includegraphics[width=1\textwidth]{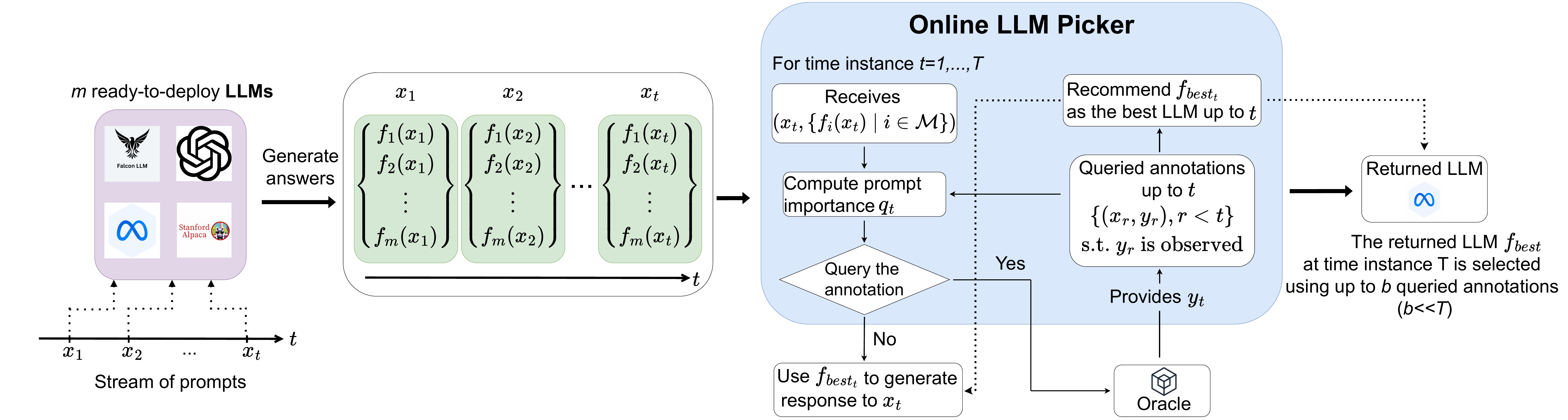}
    \caption{An overview of the \framework workflow.}
    \label{fig:ALPSov}
\end{figure}

\textbf{Contributions}: To fill this gap, in this work we propose \framework, an active model selection strategy designed to efficiently identify the best candidate LLM in online settings. 
Given an incoming stream of prompts and a small annotation budget, \framework selects the most informative prompts to annotate in order to reliably identify the best LLM for the stream. Our framework is based on the combined informativeness criterion that maximizes variance across model responses to unannotated prompts while also accounting for posterior uncertainty. \framework makes no assumptions about the underlying LLMs, allowing it to be applied to any generative task and any model collection.
\framework is depicted in Figure \ref{fig:ALPSov}.

We validate \framework through a comprehensive set of experiments spanning multiple LLM generation tasks, including open-ended question answering \citep{rajpurkar_squad_2016}, open-ended question answering with no correct response \citep{rajpurkar_know_2018}, grammar correction of natural language text \citep{loem_exploring_2023}, and both arithmetic \citep{gambardella_language_2024} and calculus problems expressed in code-like notation \citep{gao_pal_2023}. Our model collections across these tasks contain over 130 language models. 
We rank the models using both ROUGE \citep{lin_rouge_2004} and embedding-based similarity measures (e.g. BERTScore \citep{zhang_bertscore_2020}), so that we can capture lexical and semantic agreement with the reference outputs. 
Our results demonstrate that \framework identifies the best or near-best LLM in online environments while requiring up to $71.67$\% fewer annotations than competing baselines, and reduces regret by up to a factor of $2.51\times$ when applied to sequential generation on unannotated prompts across the stream.


\section{Related Work}
A wide range of metrics 
are proposed for the \textbf{evaluation of language models}, from early automatic metrics such as ROUGE \citep{lin_rouge_2004}, BLEU \citep{papineni_bleu_2002}, and METEOR \citep{lavie_meteor_2007} to benchmark-driven progress with GLUE \citep{wang_glue_2019}, SuperGLUE \citep{wang_superglue_2020}, MMLU \citep{hendrycks_measuring_2021}, and BIG-bench \citep{srivastava_beyond_2023}.
More recently, evaluation has shifted toward human feedback \citep{stiennon_learning_2022}; \citep{ouyang_training_2022}, but reliance on annotators is costly and limits scalability in large-scale or streaming settings. 
LLM-as-a-judge methods (\cite{gu_survey_2025}; \cite{bai_mt-bench-101_2024}; \cite{chiang_chatbot_2024}; \cite{dubois_length-controlled_2025}) offer more scalable pipelines, though they introduce bias. 
Efficiency is also identified as a central concern with Active Testing (\cite{berrada_scaling_2025}) which reduces annotation cost of evaluation through adaptive sampling. However, such methods focus on evaluating a fixed model or small set of models, whereas our setting requires identifying the best LLM for an arbitrary data stream under limited annotations.

The most relevant line of work for our setting is \textbf{active model selection}, which seeks to identify the best model from a set of candidates by querying informative examples \citep{madani_active_2012, karimi_online_2021, ashury-tahan_label-efficient_2024, okanovic_all_2024, kay_consensus-driven_2025, liu_contextual_2022, li_online_2024, hara_active_2024, gardner_bayesian_2015} serving as the primary motivation for our work. However, most of these approaches assume a pool-based setup where all queries are available upfront\citep{okanovic_all_2024, ashury-tahan_label-efficient_2024, kay_consensus-driven_2025} , or focus on classification tasks\citep{karimi_online_2021, liu_contextual_2022, okanovic_all_2024, kay_consensus-driven_2025, li_online_2024, hara_active_2024} , and are therefore not directly applicable to our setting.

In this work, we introduce the novel problem of online active model selection for large language models and present several strategies as baselines (introduced in Section~\ref{sec:experiments}) along with ours. We relegate discussions of related but less directly relevant frameworks to Appendix \ref{app:rel_work}.

\section{Online LLM Picker}

\subsection{Problem Statement and Background} 
Consider the inference phase with a stream of $T$ prompts 
$\{ x_t \in \mathcal{X} \mid t \in [T] \}$ drawn from an unknown distribution.  
Each prompt $x_t$ is associated with a reference annotation $y_t \in \mathcal{Y}$,  
which remains hidden unless queried from an oracle.

We consider $m$ pretrained language models 
$\mathcal{M} = \{ f_i : \mathcal{X} \to \mathcal{Y} \mid i \in [m] \}$.  
For each prompt $x_t$, model $i$ produces a response $f_i(x_t)$.  
At time step $t$, given the set of model responses 
$\{ f_i(x_t) \mid i \in \mathcal{M} \}$, we decide whether to query the oracle for the reference annotation $y_t$ or not.  
If $y_t$ is not queried, it remains hidden; otherwise, we compute the loss vector as
\begin{equation}
\bm{\ell}_t^{y} = [\ell_{t,i}^{y}]_{i\in \mathcal{M}} \quad \text{where} \hspace{.5em}  \ell_{t,i}^{y} = 1 - d\big(y_t, f_i(x_t)\big)\label{eq: loss definition}
\end{equation}
for some similarity score $d: \mathcal{Y}\mapsto \mathcal{Y}$. In our framework, we choose $d(\cdot, \cdot)$ to be ROUGE-L \citep{lin_rouge_2004} or BERTScore \citep{zhang_bertscore_2020} in our case \citep{rehman_evaluating_2025, sul_balancing_2023}.

Given an annotation budget $b \ll T$ for the entire stream, our objective is to select the best language model for the remaining $T-b$ prompts using only the annotated prompts. At each time step $t$, our method returns the language model $\fbestt$ based on the annotations observed up to time $t$ such that $\fbestt \sim \mathbf{p}_t$, where $\mathbf{p}_t$ represents our posterior belief over the language models. 
Our objective is to annotate the most informative examples across the stream so that $\fbestt$ closely approximates the \emph{true} best language model, defined as the model that incurs the minimum cumulative loss if all annotations in the stream were available: $\fbest \coloneq \argmin_{i \in \mathcal{M}} \sum_{t=1}^T \ell_{t,i}^y$.
Formally, we aim to minimize the cumulative loss of $\fbestt$ relative to $\fbest$, which is captured by the standard regret:  
\begin{equation}
   R_T = \sum_{t=1}^T \ell_{t,\fbestt}^y - \sum_{t=1}^T \ell_{t,\fbest}^y.\label{eq:regret}
\end{equation}

Since our ultimate goal is to identify the best model under the annotation budget $b$, we also evaluate the quality of the returned model using two complementary metrics. The first is \textit{identification probability}, which measures the probability of successfully recovering the true best model in a stochastic data stream. The second is \textit{annotation efficiency}, which quantifies the annotation cost saved by identifying the best or near-best models under budget $b$. We provide their formal definitions in Section~\ref{sec:experiments}.  

\subsection{The Algorithm}
In this section, we introduce \framework, which aims to identify the best language model for a given stream of prompts with limited reference annotations from the oracle.

At a high level, our algorithm operates as follows. At each time step $t$, we decide whether to query the annotation of the prompt $x_t$ via a random experiment: we draw 
a Bernoulli random variable $Q_t \sim \operatorname{Bernoulli}(q_t)$, 
with \emph{query probability} $q_t$, where the choice of $q_t$ depends on 
factors introduced later. If a query is made ($Q_t = 1$), we update our posterior belief over the models using the newly observed annotation. We denote this posterior by 
$\mathbf{p}_t = [p_{t, i}]_{i \in \mathcal{M}}$ where $p_{t, i}$ represents the probability that model $i$ is the true best language model $\fbest$ at time $t$. If no query is made ($Q_t = 0$), the reference annotation for $x_t$ remains 
hidden and the posterior remains unchanged. 

To perform the posterior update, we first construct an importance-weighted loss estimate using $\ell_{t,i}^y$ in Equation \ref{eq: loss definition}:  
\begin{equation*}
    \hat{\ell}_{t,i}^y \;=\; \frac{\ell_{t,i}^y}{q_t}\, Q_t 
    \quad \forall i \in \mathcal{M}
\end{equation*}
where $1/q_t$ serves as an importance weight, correcting sampling bias from the querying process by assigning greater weight to under-sampled instances.  

We then obtain the cumulative loss estimate up to time $t$ as  
\begin{equation*}
    \hat{L}_{t,i} 
    \;=\; \hat{L}_{t-1,i} \;+\; \hat{\ell}_{t,i}^y
    \quad \forall i \in \mathcal{M}
\end{equation*}
and update the posterior distribution $\mathbf{p}_t=[p_{t, i}]_{i\in\mathcal{M}}$ using the Exponential Weights (EW) algorithm by ~\cite{littlestone_weighted_1994} with an adaptive learning rate $\eta_t$:  
\begin{equation}\label{eq:posterior update}
    p_{t,i} \;\propto\; \exp\{-\eta_t \hat{L}_{t-1,i}\}.
\end{equation}
Appendix \ref{app:detailspt} details the derivation of $\bm{p}_t$.
We expand on our learning rate $\eta_t$ as follows.
\subsubsection{Adaptive Learning Rate}
While time-based decay schedules are common~\citep{karimi_online_2021}, 
inspired by AdaHedge~\citep{rooij_follow_2013}, we propose an adaptive learning rate $\eta_t$ that updates based on observed annotations.
In AdaHedge, the dynamic learning rate is defined in terms of the cumulative mixability gap, which measures the cumulative approximation error incurred when approximating the Hedge loss with the mix loss~\citep{rooij_follow_2013}.
By means of Bernstein's bound, the mixability gap at a certain round $t$ can be expressed in terms of variance of the losses. In addition, \cite{rooij_follow_2013} shows that the cumulative mixability gap grows at most as the square root of the cumulative loss variance up to $t$. Motivated by this, we introduce
an adaptive learning rate that depends on the variance of the losses over the queried examples. The variance of the loss can be computed as:
\begin{equation}\label{eq: variance}
    \underset{i \sim \bm{p}_t}{\mathrm{Var}} \, {\ell}_{t,i}^y
    =\langle\bm{p}_t, \bm{\ell}_t^y \odot \bm{\ell}_t^y\rangle - (\langle\bm{p}_t,\bm{\ell}_t^y\rangle)^2. \footnote{Here, $\odot$ denotes the element-wise product and $\langle \cdot, \cdot \rangle$ denotes the inner product.}
\end{equation}
Although variance-based strategies exist for adapting the learning rate~\citep{rooij_follow_2013}, they are less effective with limited annotations, where sparse queries and a few noisy examples can disproportionately distort the variance estimate. Following the idea of exponential moving averages in optimization \citep{kingma_adam_2017}, which keeps exponential moving averages of the gradient and its square (first and second moments), applies bias correction, and uses them to form a per-parameter adaptive step, we instead introduce an exponential moving average of the variance of the observed examples as a proxy. 
This proxy does not track the exact variance of the underlying losses, 
but provides a smoothed, adaptive estimate that emphasizes recent 
observations while discounting stale ones, thus making the learning rate more robust 
under sparse and noisy feedback.  

We compute the exponential moving average of the variance as:
\begin{equation}
\mathcal{V}_t =
    \begin{cases}
        \beta \cdot \mathcal{V}_{t-1} + (1-\beta)\cdot \underset{i \sim \bm{p}_t}{\mathrm{Var}} \, \ell_{t,i}^y, & \text{if } y_t \text{ is observed}, \\[6pt]
        \mathcal{V}_{t-1}, & \text{otherwise,}
    \end{cases}
    \label{eq:proxy_var}
\end{equation}
where $\beta\in[0,1]$ is the decay rate 
of the exponential moving average. The effective window size of this exponential moving average is approximately $(1-\beta)^{-1}$, meaning that values of $\beta\approx1$ place more weight on past observations and correspond to a longer memory.

At the beginning of the stream, we have $\mathcal{V}_0 = 0$ and no annotations are available yet.
This initialization biases the variance proxy toward zero, particularly in the early time instances. 
As discussed in \cite{kingma_adam_2017}, this bias can be corrected by a bias-adjusted proxy of the average variance:
\begin{equation}
    \hat{\mathcal{V}}_t = \frac{\mathcal{V}_t}{1-\beta^t}.
    \label{eq:proxyv}
\end{equation}
To connect back to the framework of \citet{rooij_follow_2013}, which 
relates the cumulative mixability gap to the cumulative variance, 
we estimate the cumulative variance up to time $t$ by scaling the number  of time steps with the bias-adjusted proxy of the average variance with $t \cdot \hat{\mathcal{V}}_t$ (\ref{eq:proxyv}). Eventually, at any time instance $t$, we compute the adaptive learning rate $\eta_t$ as follows:
\begin{equation}
    \eta_t = \sqrt{\frac{\log m}{t \cdot (\hat{\mathcal{V}_t}+\epsilon)}}
    \label{eq:ada_learn}
\end{equation}
where $\epsilon > 0$ is a small constant added for numerical stability. 
This makes sure that $\eta_t$ remains finite and well-defined for all $t$, 
including the case $\hat{\mathcal{V}}_t \to 0$, which corresponds to (near-) identical pairwise losses among the answers generated by language models.
With a uniform prior over the $m$ experts, the initial information cost of not knowing the best expert is $\log m$, which appears as $(\log m)/\eta$ in the regret bound \citep{rooij_follow_2013}.

Plugging Equation \ref{eq:ada_learn} into Equation \ref{eq:posterior update} completes the model posterior update. We now discuss the query probability $q_t$ (\ref{eq:query_prob}), which determines the outcome of the query decision at each time step $t$.

\subsubsection{Query Probability}
The query probability $q_t$ at each time step is designed to capture our uncertainty regarding the identity of the best model on the given prompt $x_t$, and subsequently guide our decision of whether to request the annotation of $x_t$ or not. Importantly, this probability should not only reflect predictive uncertainty but also integrate the current posterior belief over the language models, and balance the competing objectives of exploration and exploitation by doing so. A natural way to operationalize uncertainty in streaming settings is through variance-based measures\citep{karimi_online_2021}, which serve as a principled proxy for the potential information gain associated with querying an annotation. Intuitively, the goal is to estimate the expected value of acquiring the annotation for $x_t$, that is, the degree to which it would refine our estimate of the best model. More concretely, this quantity can be estimated from the losses incurred by the model responses $\{ f_i(x_t) \mid i \in \mathcal{M} \}$, together with the posterior distribution over models $\bm{p}_t$ at time $t$. Since the reference annotation $y_t$ is not available, this variance estimation is hypothetical.  
In the classification setting, such a hypothetical variance is well-defined because the true label at time $t$ is assumed to lie within a finite set of predefined classes.  
In contrast, in the generative setting there is no fixed label space: the set of valid responses is open-ended and potentially unbounded, which makes evaluation and comparison 
substantially more challenging.  
To address this, we approximate the response space at each time $t$ by considering 
the complete set of language model outputs $\{ f_i(x_t) \mid i \in \mathcal{M} \}$.

Towards that, we denote the (pairwise) loss between the response generated by the language models and that generated by the model $k$ by $\bm{\ell}_t^{k}=[{\ell}_{t,i}^k]_{i\in \mathcal{M}}$ where ${\ell}_{t,i}^k = 1-d(f_i(x_t), f_k(x_t))$.
At each time instance $t$, we treat the response of each language model $k\in \mathcal{M}$ as the hypothetical reference annotation. We then compute the maximum hypothetical variance among the losses with:
\begin{align}
    \max_{k\in \mathcal{M}} \underset{i \sim \bm{p}_t}{\mathrm{Var}} \, {\ell}^{k}_{t,i} 
    =\langle\bm{p}_t, \bm{\ell}_t^k \odot \bm{\ell}_t^k\rangle - (\langle\bm{p}_t,\bm{\ell}_t^k\rangle)^2\label{eqn: hypothetical var}
\end{align}
which represents our uncertainty about the annotation of $x_i$ given the model posterior $\bm{p}_t$. 
If the language model responses are nearly identical where $\bm{\ell}_t^{k} \approx 0$, or if they differ to the same degree such that $\bm{\ell}_t^{\,k}$ is similar for all $k \in \mathcal{M}$, then the dispersion under $\bm{p}_t$ is negligible. In this case, the annotation $y_t$ provides little information, contributes minimally to model 
selection, and has only a negligible effect on regret.  
Conversely, if the maximum hypothetical variance is large, for instance, when $\bm{\ell}_t^{k}$ exhibits substantial heterogeneity across model responses, then the annotation $y_t$ is informative.

Revisiting the model posterior, which our algorithm updates based on the observed annotations, we find that in some cases it concentrates around a single best model, whereas in others it remains diffuse across multiple candidates.
When the posterior distribution is already concentrated after some annotations, additional annotations have very little impact on the posterior update. Otherwise, a few strategically chosen annotations can shift the balance and determine the best model.
This perspective is rooted in the Bayesian experimental design, where the value of new information is measured by its expected reduction in posterior uncertainty. 
Shannon entropy of the posterior provides a natural quantification of `how much remains to be learned' about which model or parameter is best \citep{lindley_measure_1956, mackay_information-based_1992, sebastiani_maximum_2025}. 
A parallel also exists in online decision making: exploration is most valuable when the uncertainty expressed in terms of posterior entropy about the best choice is high, since this is when information can most effectively reduce future regret \citep{russo_information-theoretic_2016, russo_learning_2017}.

Combining multiple signals such as variance and entropy is a common design pattern in active learning and streaming settings, such as scaling committee disagreement by input density \citep{mccallum_employing_1998}, weighting predictive entropy by local density \citep{zhu_active_2008}, querying when both uncertainty and density are high \citep{ienco_high_2014}, and multiplying predictive entropy by a coverage-based representativeness factor \citep{katragadda_active_2023}.
Following this principle, we incorporate posterior entropy into our query probability. Specifically, we compute the normalized entropy of the posterior distribution over models:
\begin{equation}
\bar{\mathbb{H}}(\bm{p}_t) \;=\; \frac{\mathbb{H}(\bm{p}_t)}{\log m}\in[0,1]
\qquad
\textrm{where} \hspace{.5em} \mathbb{H}(\bm{p}_t) \;=\; -\sum_{i=1}^m {p}_{t,i}\,\log {p}_{t,i}.
\label{eq:norm_entr}
\end{equation}
We normalize entropy by $\log m$ to place uncertainty on a fixed $[0,1]$ scale, independent of the number of models $m$. This makes thresholds and query schedules comparable across settings and interpretable: $\bar{\mathbb{H}}=0$ when one model dominates (high confidence) and $\bar{\mathbb{H}}=1$ under a uniform posterior (maximal uncertainty).
We then incorporate this normalized entropy by scaling the variance in \ref{eqn: hypothetical var} such that $\underset{i \sim \bm{p}_t}{\mathrm{Var}} \, {\ell}_{t,i}^k\cdot\bar{\mathbb{H}}(\bm{p}_t)$.



Finally, we define the query probability as
\begin{align} q_t= 
\begin{cases} \max{\{\max_{k\in \mathcal{M}}\underset{i \sim \bm{p}_t}{\mathrm{Var}} \, {\ell}_{t,i}^k\cdot\bar{\mathbb{H}}(\bm{p}_t),\,\eta_t\}}, &\text{if } \max_{k\in \mathcal{M}}\underset{i \sim \bm{p}_t}{\mathrm{Var}} \, {\ell}_{t,i}^k\neq0 \\ 0,&\text{otherwise.} 
\end{cases} \label{eq:query_prob} 
\end{align}
where $\eta_t$ is a time-decaying lower bound based on the adaptive learning rate.
This bound prevents two problems: when variance is extremely small, importance-weighted loss estimates can become unstable and inflate regret; and when predictions are overly confident, the algorithm may skip prompts that are actually informative.

In essence, the query rule is based on the hypothetical loss variance across models under the current posterior as well as the direct posterior entropy. High variance and entropy trigger more queries, while a concentrated posterior reduces them. Early on, queries are mainly variance-driven, but over time $q_t$ balances both factors and reflects model disagreement and overall uncertainty about the annotation of the incoming queries. 

The pseudocode of \framework is depicted in Algorithm \ref{alg:onlineLLMPick}.

\algnewcommand\algorithmicReturn{\textbf{Return:}}
\algrenewcommand\Return{\item[\algorithmicReturn]}

\begin{algorithm}[htbp]
    \caption{Online LLM Picker}\label{algorithm}
    \begin{algorithmic}
        \Require Language models $\mathcal{M}$
        \State Set $\hat{{L}}_{0,i}=0 \quad \forall\, i\in\mathcal{M}, \, \mathcal{V}_0=0$
        \For{$t=1,2,...,T$}
            \State $\eta_t = \sqrt{\frac{\log m}{t \cdot (\hat{\mathcal{V}_t}+\epsilon)}}$
            \State ${p}_{t,i} \propto \exp\{-\eta_t \hat{{L}}_{t-1,i} \}\quad\forall\, i\in\mathcal{M}$ \Comment{update posterior over experts}
            \State Get the models' predictions $\{f_i (x_t)\mid i \in \mathcal{M}\}$ for $x_t$
            \State Recommend $\fbestt:=\arg\max\limits_{i\in\mathcal{M}}p_{t,i}$ as the best model up to time instance $t$
            \State Compute $q_t$ as in \emph{(\ref {eq:query_prob})} and sample $Q_t\sim \operatorname{Bernoulli}(q_t)$
            \If{Q=1} \Comment{query the annotation $y_t$}
                \State Update $\mathcal{V}_t$ as in \emph{(\ref{eq:proxy_var})}
                \State $\hat{{L}}_{t,i}=\hat{{L}}_{t-1,i}+\frac{{\ell}_{t,i}^y}{q_t} \quad\forall\,i\in \mathcal{M}$
            \Else
                \State Annotate $x_t$ with $\fbestt(x_i)$
                \State $\hat{{L}}_{t,i}=\hat{{L}}_{t-1,i}$
            \EndIf
        \EndFor\\
    \Return{ $\fbest = \argmax_{i\in\mathcal{M}} p_{i, T}$ }\Comment{return the best model for the stream}
    \end{algorithmic}
    \label{alg:onlineLLMPick}
\end{algorithm}
    
\section{Experiments}\label{sec:experiments}
We evaluate \framework\ for active LLM selection in streaming settings using several public generative datasets and model collections containing more than 130 language models. As this is the first study of its kind, we introduce baselines and compare performance across multiple metrics: \emph{identification probability}, \textit{regret}, and \textit{annotation efficiency} for identifying best or near-best models across the stream.

\subsection{Datasets and Models}
We benchmark \framework against competing baselines across datasets from different generative tasks.
First, we test the algorithms on collections of LLMs for the SQuAD \citep{rajpurkar_squad_2016} and SQuAD v2 \citep{rajpurkar_know_2018} datasets, which are designed for question answering tasks, with the latter including questions that may not have an answer in the given context.
Question answering is often framed as a non-generative task, especially in the extractive setting where models directly copy answer spans from the input text.
In contrast, our models generate answers in their own words, producing open-ended text instead of directly extracting it.
We also use the Comprehensive Arithmetic Problems dataset \citep{lee_garrethleecomprehensive-arithmetic-problems_2024} that features a range of algebraic expressions, the Calculus Datasets (FDU) \citep{zhang_di-zhang-fducalculus-dataset_2025} that consist of more advanced numerical and symbolic problems represented in LaTeX code, and the Grammar Correction dataset \citep{agentlans_agentlansgrammar-correction_2024}, which challenges models to fix grammar mistakes in English sentences. In addition, we also consider MT-Bench \citep{bai_mt-bench-101_2024}, a benchmark for evaluating multi-turn conversational ability of LLMs, and multiple datasets from the HELM Benchmark\citep{liang_holistic_2023}. In particular, we include two Longform Question Answering (LF-QA) datasets from classic HELM, LF-QA Canonical and LF-QA Prompt, respectively. From MedHELM \citep{bedi_medhelm_2025} we add MedCalc \citep{khandekar_medcalc-bench_2024} to our experiments. We also include FinQA\citep{chen_finqa_2022} from HELM Finance.

As for the language models, we include several open-source and proprietary LLMs (\cite{almazrouei_falcon_2023}, \cite{radford_language_2019}, \cite{brown_language_2020}, \cite{raffel_exploring_2023}, \cite{openai_gpt-4_2024}, \cite{deepseek-ai_deepseek-v3_2025}, \cite{touvron_llama_2023}, \cite{grattafiori_llama_2024}), ranging from those with a few million parameters to those with several billion. Some of these models are fine-tuned on the specific datasets of interest, while others are general-purpose models suitable for a wide range of generative tasks.
We also expand our model collection to DeepSeek \citep{deepseek-ai_deepseek-v3_2025} and various GPT-based models \citep{brown_language_2020, kocon_chatgpt_2023, ouyang_training_2022}. An overview of our datasets and models is provided in Table \ref{tab:models_datasets} as well as LLM scores on the test datasets in Figure \ref{fig:acc_hist_all} in Appendix \ref{app:datasets_and_llms}.
To simulate multiple LLMs using the same underlying model, we also apply prompt engineering techniques to instruction-tuned LLMs, encouraging them to behave in agent-specific ways. Prompt engineering is also used to improve performance on our benchmark datasets by reducing unnecessary verbosity and avoiding irrelevant explanations in the output. Our work is model-agnostic and makes no assumptions about LLM architectures or performance. To approximate real-world use, we evaluate across a diverse suite of datasets. More details on LLM collections and datasets can be found in Appendix \ref{app:datasets_and_llms}.

\begin{figure}[t]
\centering
\includegraphics[width=0.9\linewidth]{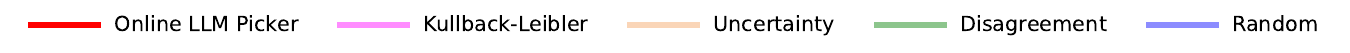}\\[-2pt]
\begin{minipage}[t]{.19\linewidth}\vspace{0pt}\centering
\includegraphics[width=\linewidth]{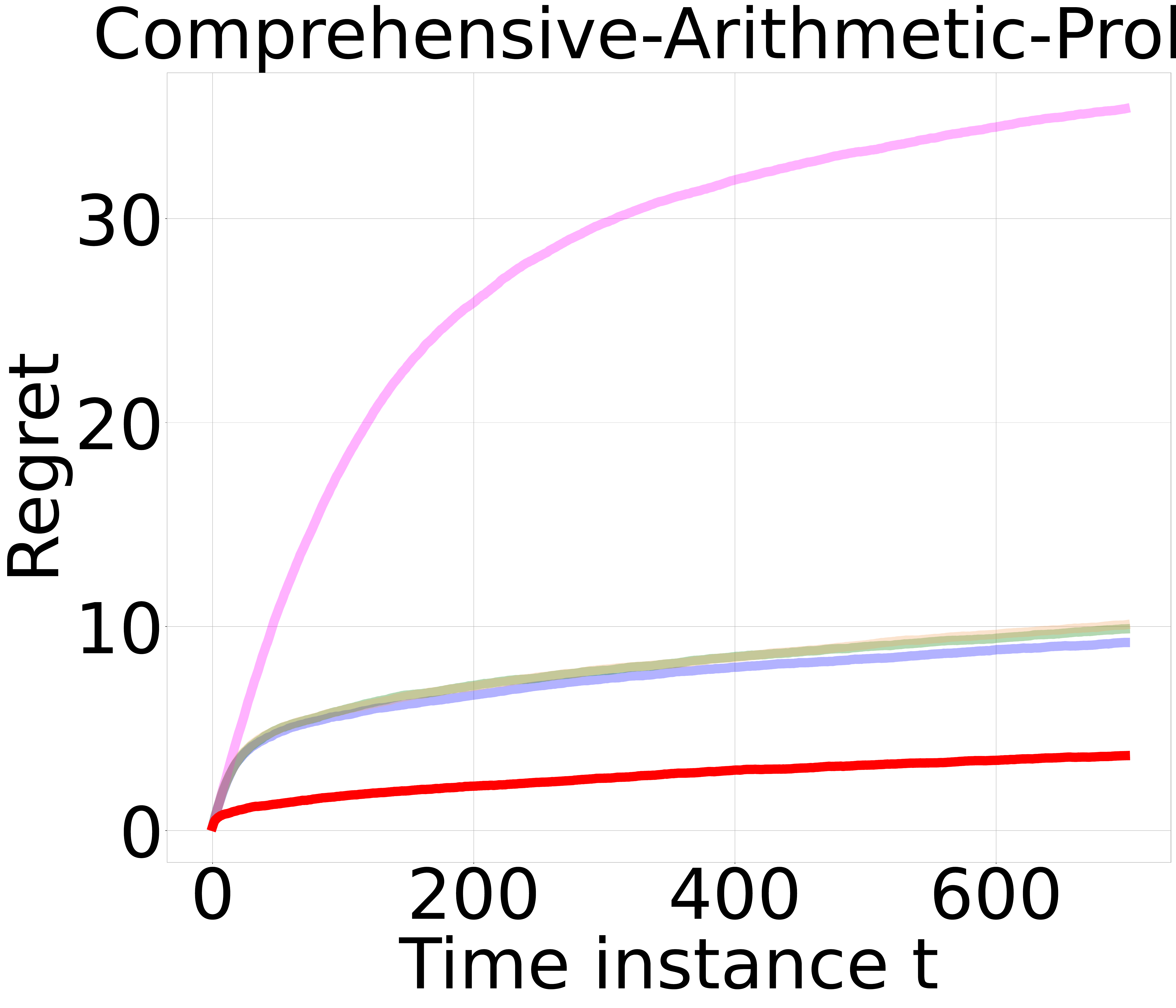}
\end{minipage}\hspace{0pt}%
\begin{minipage}[t]{.19\linewidth}\vspace{0pt}\centering
\includegraphics[width=\linewidth]{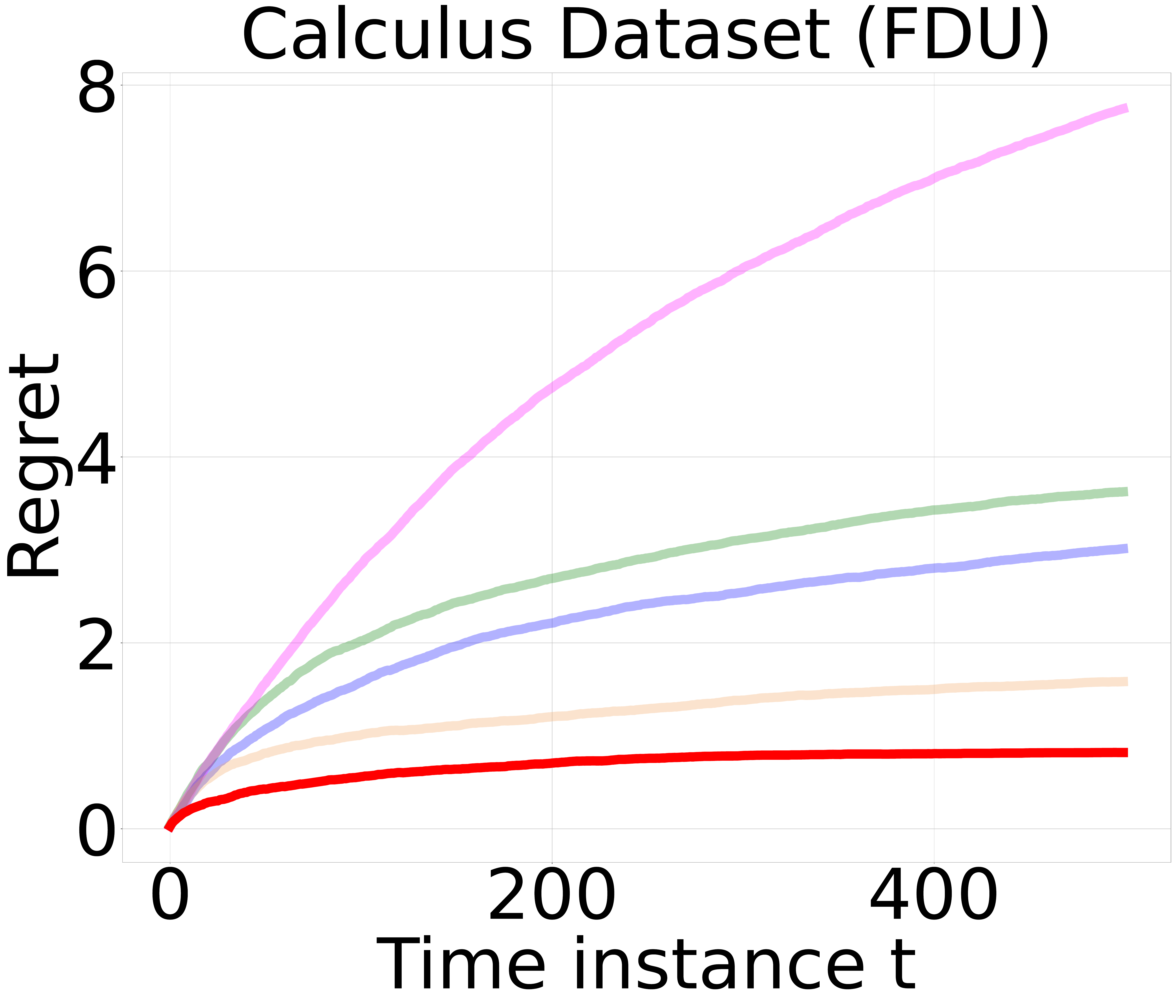}
\end{minipage}\hspace{0pt}%
\begin{minipage}[t]{.19\linewidth}\vspace{0pt}\centering
\includegraphics[width=\linewidth]{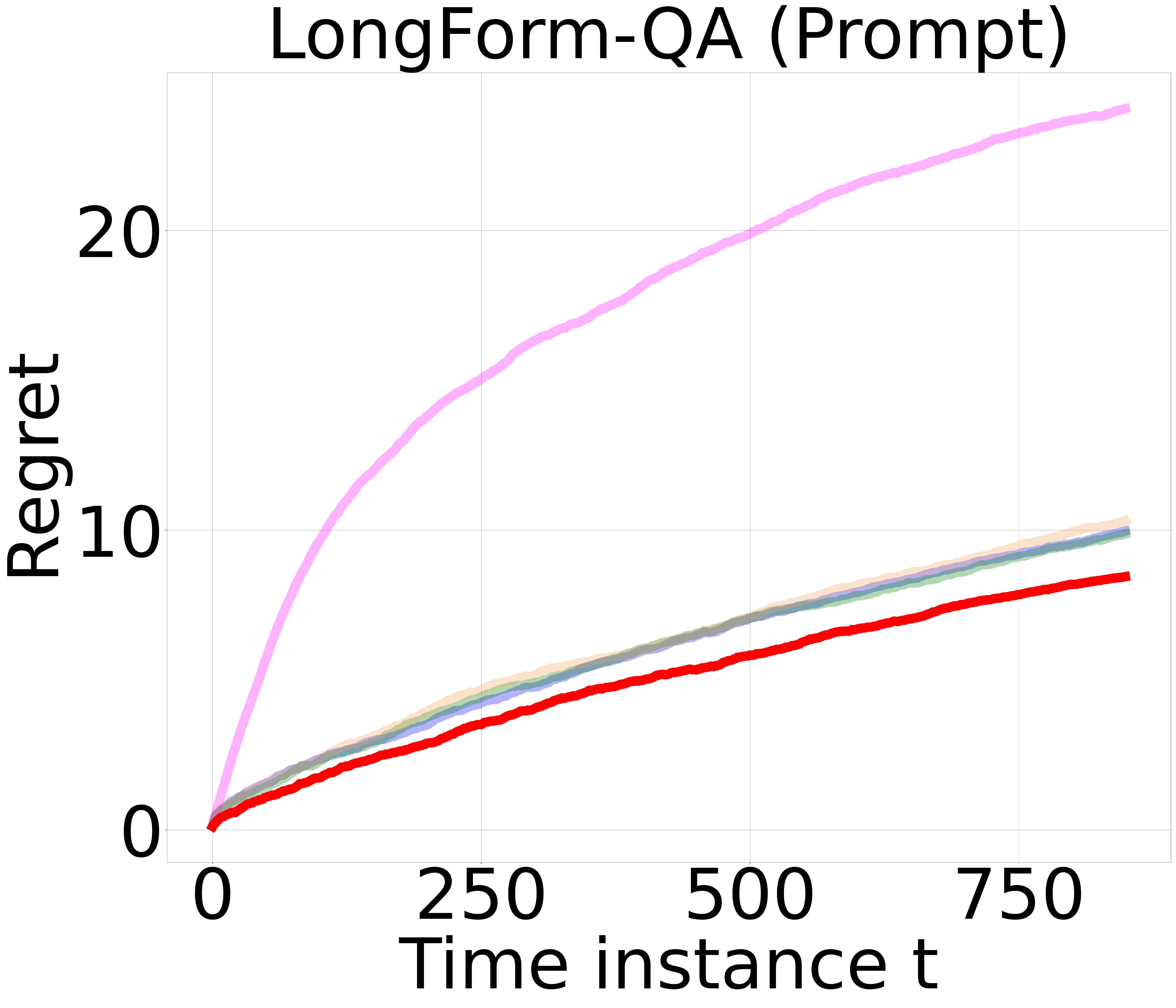}
\end{minipage}\hspace{0pt}%
\begin{minipage}[t]{.19\linewidth}\vspace{0pt}\centering
\includegraphics[width=\linewidth]{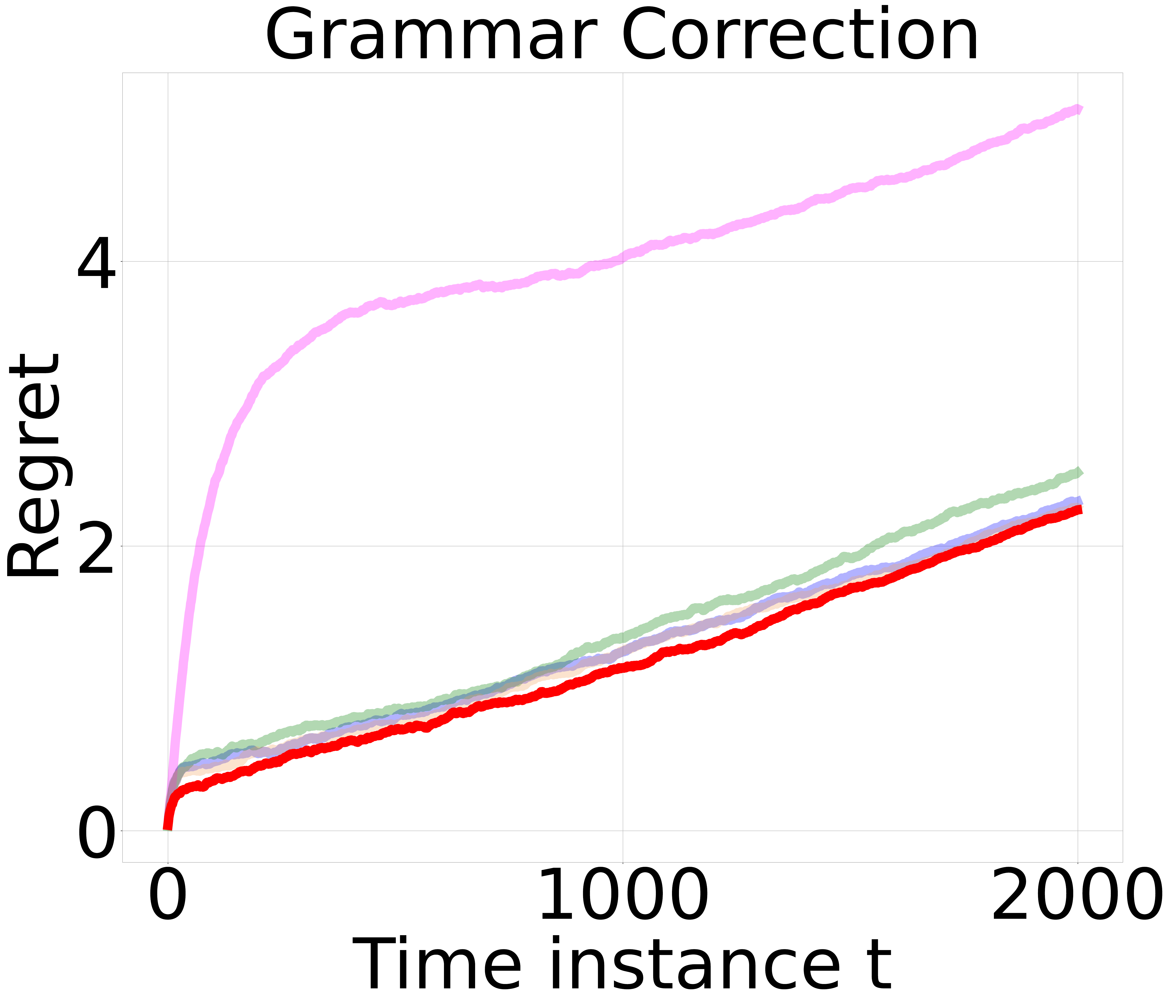}
\end{minipage}\hspace{0pt}%
\begin{minipage}[t]{.19\linewidth}\vspace{0pt}\centering
\includegraphics[width=\linewidth]{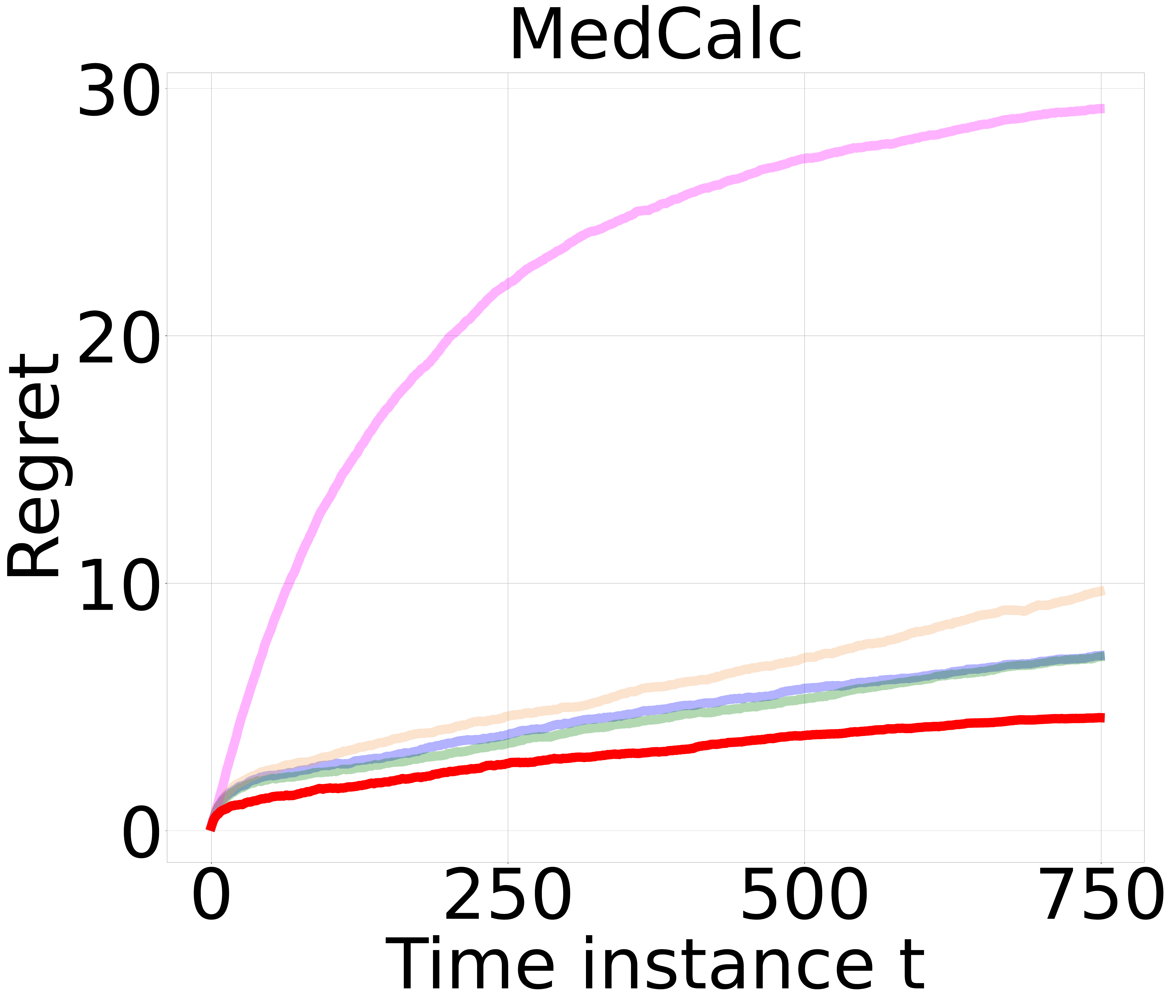}
\end{minipage}

\vspace{2pt}

\begin{minipage}[t]{.19\linewidth}\vspace{0pt}\centering
\includegraphics[width=\linewidth]{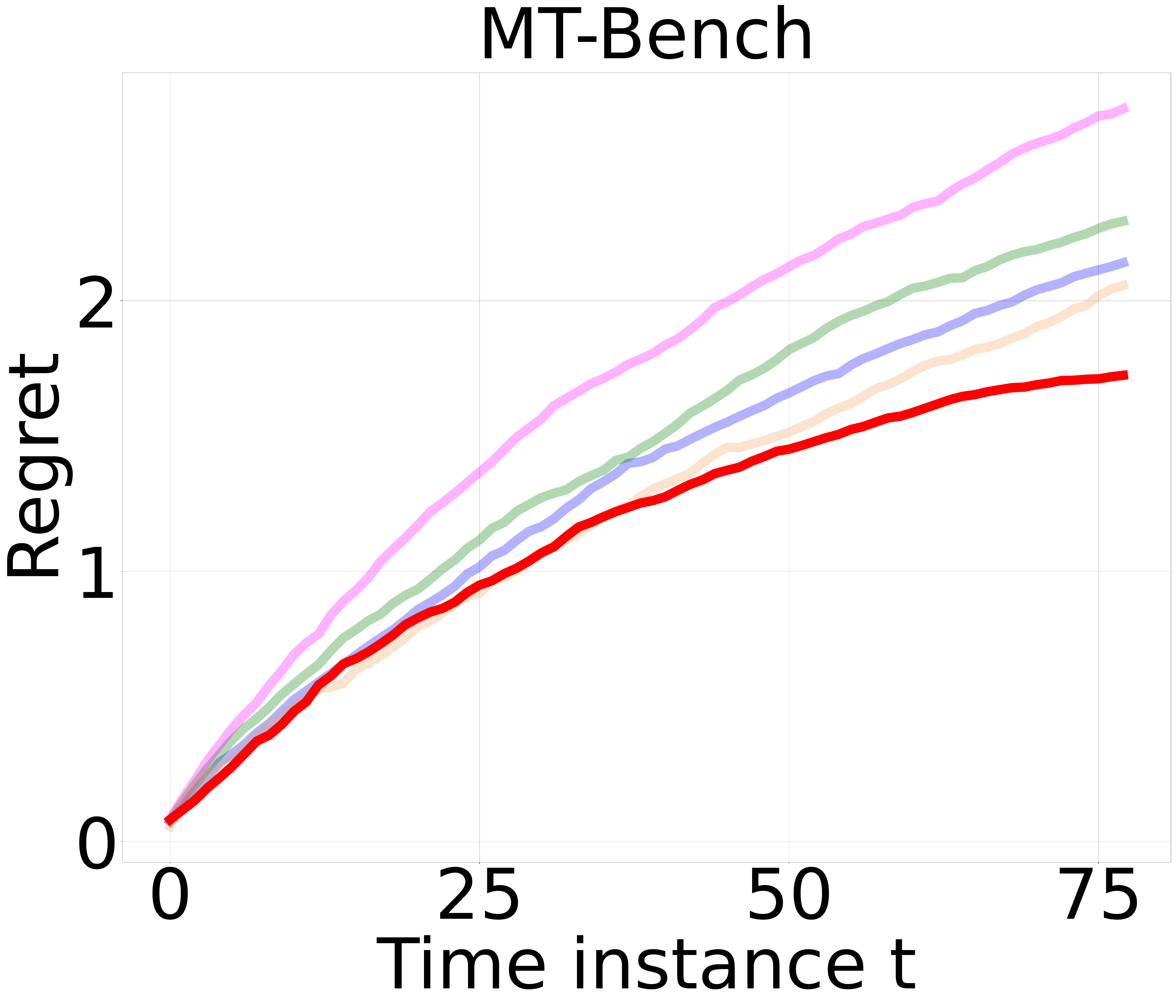}
\end{minipage}\hspace{0pt}%
\begin{minipage}[t]{.19\linewidth}\vspace{0pt}\centering
\includegraphics[width=\linewidth]{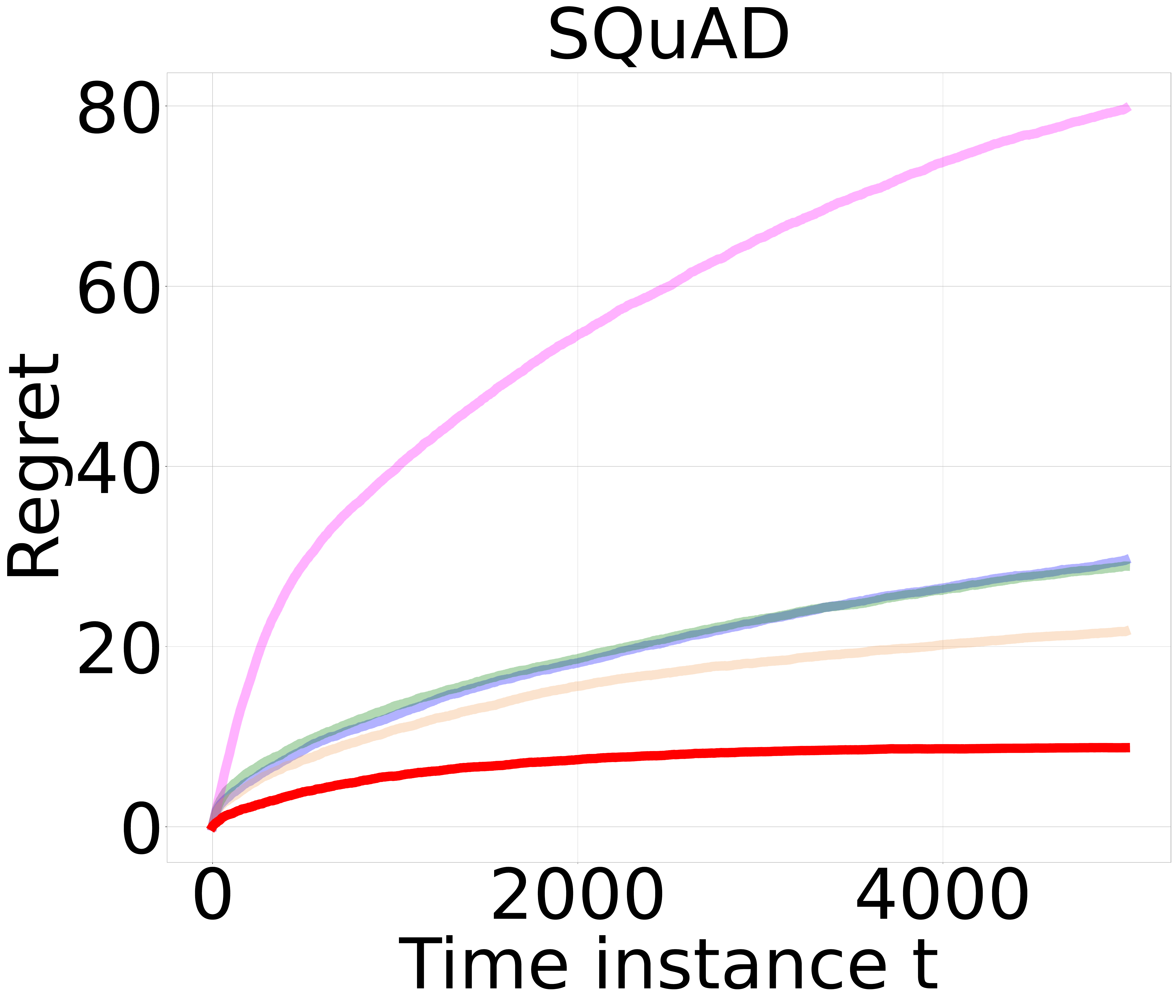}
\end{minipage}\hspace{0pt}%
\begin{minipage}[t]{.19\linewidth}\vspace{0pt}\centering
\includegraphics[width=\linewidth]{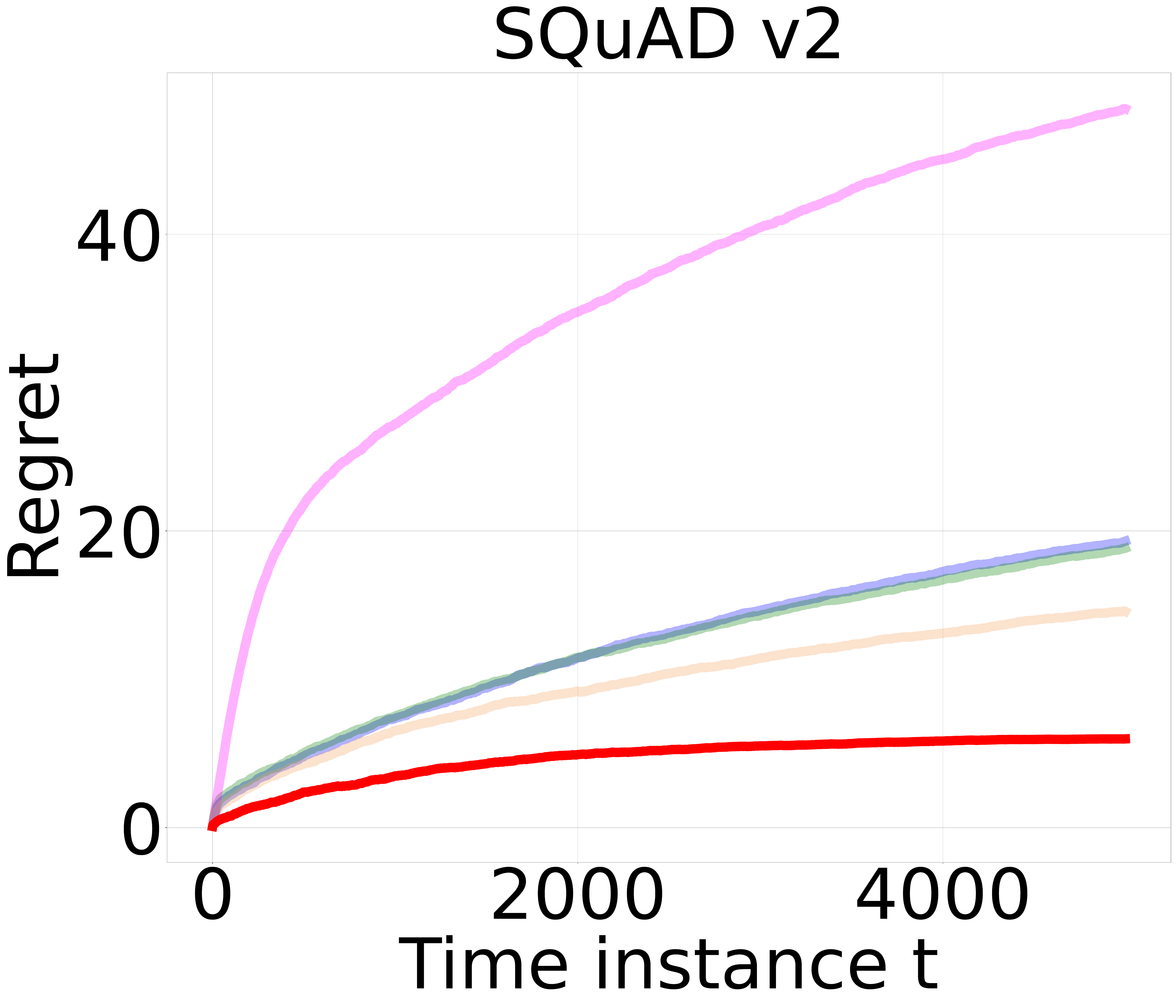}
\end{minipage}\hspace{0pt}%
\begin{minipage}[t]{.19\linewidth}\vspace{0pt}\centering
\includegraphics[width=\linewidth]{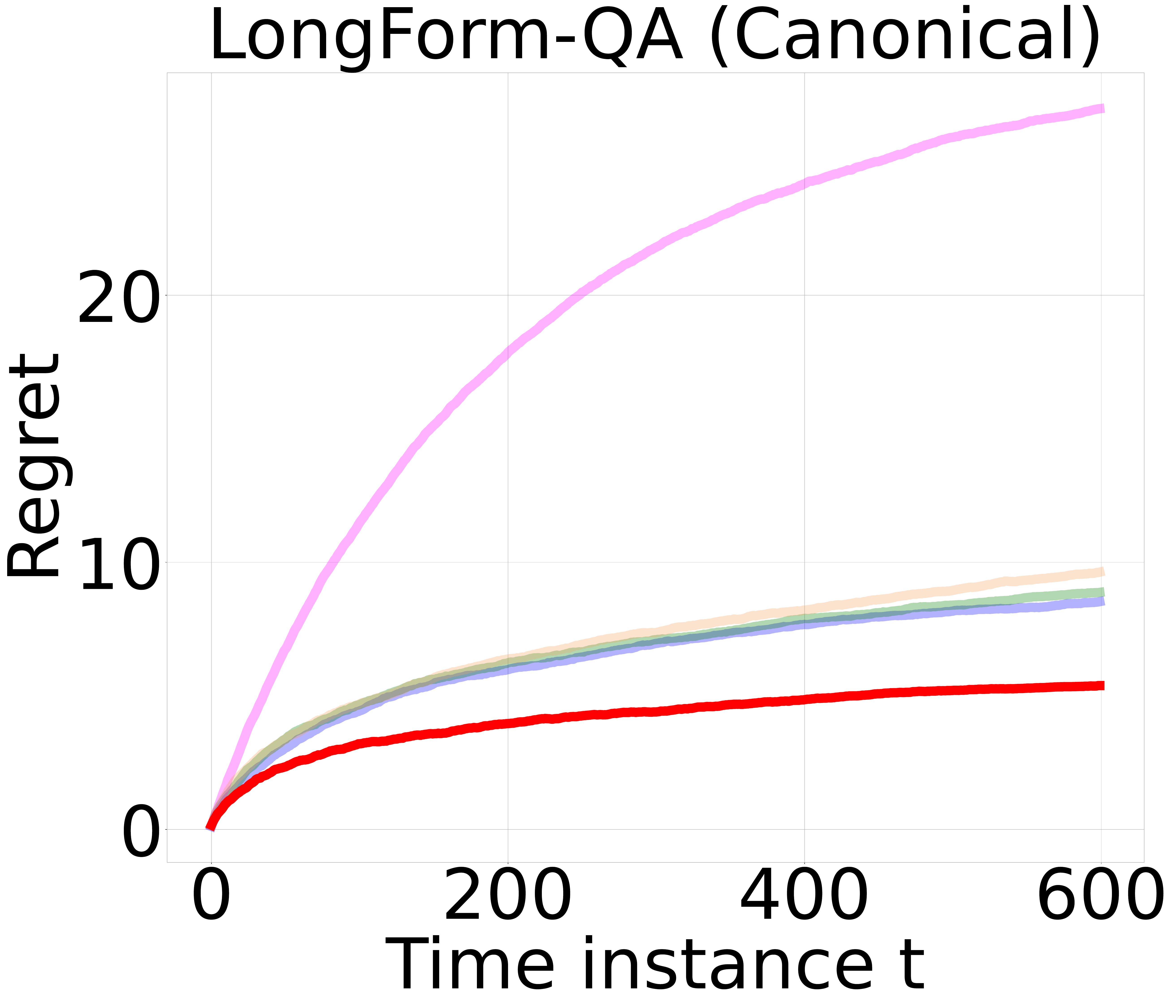}
\end{minipage}\hspace{0pt}%
\begin{minipage}[t]{.19\linewidth}\vspace{0pt}\centering
\includegraphics[width=\linewidth]{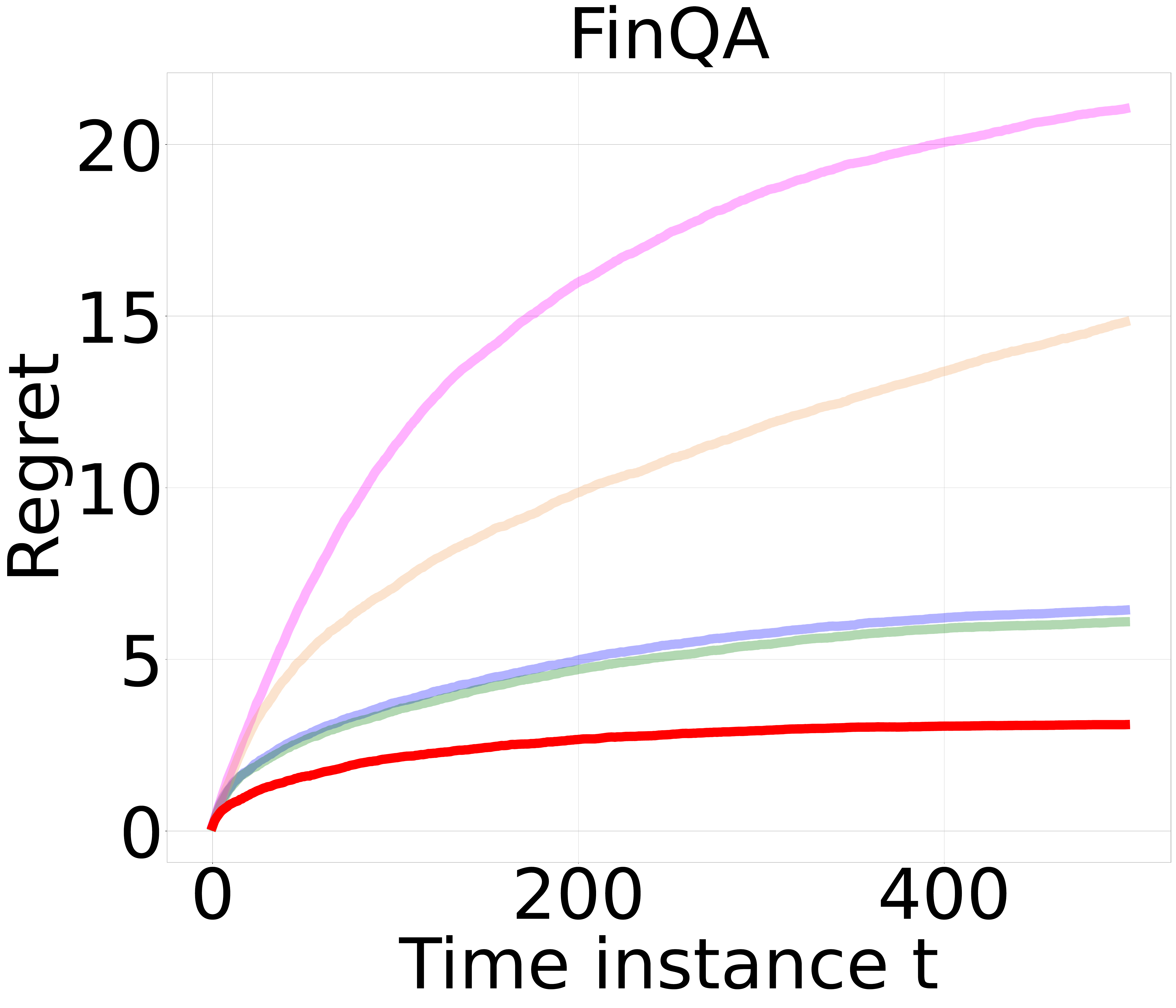}
\end{minipage}

\caption{Regret for \framework and baselines across 10 datasets.}
\end{figure}

\subsection{Baselines}
We introduce several strategies and baselines and evaluate the performance of \framework against them. 
These methods typically follow a coin-flipping strategy. At each time instance $t$, when a new prompt $x_t$ is received, the decision to query the reference annotation $y_t$ is made by sampling a Bernoulli random variable $Q_t$ with bias $q_t$, which is usually adaptive. The reference annotation $y_t$ is queried only if $Q_t=1$.

For \textbf{Random} (passive learning) baseline, we query annotation of each time instance with a fixed probability $q_t=b/T$, having an expected number of $b$ annotations queried over a stream of length $T$. For \textbf{Disagreement} baseline, we only consider prompts where the language models strongly disagree and use a fixed $q_t=b/T$ on those instances. Disagreement is quantified as the variance across experts in their average pairwise discrepancies, and a query is triggered when this measure exceeds a small threshold we introduced. We adapt the \textbf{Kullback–Leibler} baseline where KL queries with probability proportional to the Kullback–Leibler divergence \citep{shlens_notes_2014} between distribution of pairwise losses at time instance $t$ and the posterior belief, to encourage annotations when the loss distribution deviates strongly from the learned belief. We adapt \textbf{Uncertainty}\citep{dagan_committee-based_1995} to our setting to query annotations with probability similar to the Shannon entropy of the pairwise losses over candidate models.
It estimates how evenly the models support competing hypotheses and regards higher entropy as greater uncertainty.


\subsection{Evaluation Protocol}
We evaluate \framework\ separately on each dataset. For a given dataset, we generate a stream by drawing $T$ i.i.d.\ instances uniformly at random and feeding them to each method, referring to each such stream as a \emph{realization}. For each realization, we evaluate performance based on the language model returned by each method. We repeat this process over multiple independent realizations with fresh streams drawn from the test set, and average the results to estimate the expected performance for each metric.

We define the budget $b$ as the maximum number of annotations that a method can query within a given realization. We evaluate the methods across different budget levels. For a fair comparison under the same budget, we tune the hyperparameter(s) of each method to query the same number of annotations on average at the end of the stream and compare their performance. Hyperparameter tuning adjusts the query probability of a method by multiplying it with different up- or down-scaling factors to control how many annotations each method ends up requesting on average across all realizations. 

\subsection{Performance Metrics}
We evaluate each algorithm using the following metrics.
First, for a fixed annotation budget, we compute the \emph{regret} to measure how well the models returned by each method perform in sequential generation on the unannotated prompts across the stream. Second, we measure \emph{identification probability}, defined as the fraction of realizations in which the best model is correctly identified by the end of the stream. Finally, we assess \emph{annotation efficiency}, which captures how efficiently a method identifies the best or near-best language models relative to the number of annotations used.

\subsection{Experimental Results}
\subsubsection{Regret}
For a given budget where \framework returns a model with high confidence, we report the expected regret averaged over all realizations.
Our experiments are robust to the choice of budget and we evaluate \framework and baselines at very different budget levels.
In all cases, \framework consistently achieves a reduction in regret, in some cases reaching up to a factor of $2.51\times$ with respect to the best competing baseline, which shows its ability for sequential generation even well before exhausting its annotation budget. Details on annotation budget levels and further results are reported in Table~\ref{tab:regret_framework} of Appendix~\ref{app:ext_res}.

\subsubsection{Identification Probability}
\begin{figure}[t]
\centering
\includegraphics[width=\linewidth]{images/legend.pdf}\\[-2pt]

\begin{minipage}[t]{.19\linewidth}\vspace{0pt}\centering
\includegraphics[width=\linewidth]{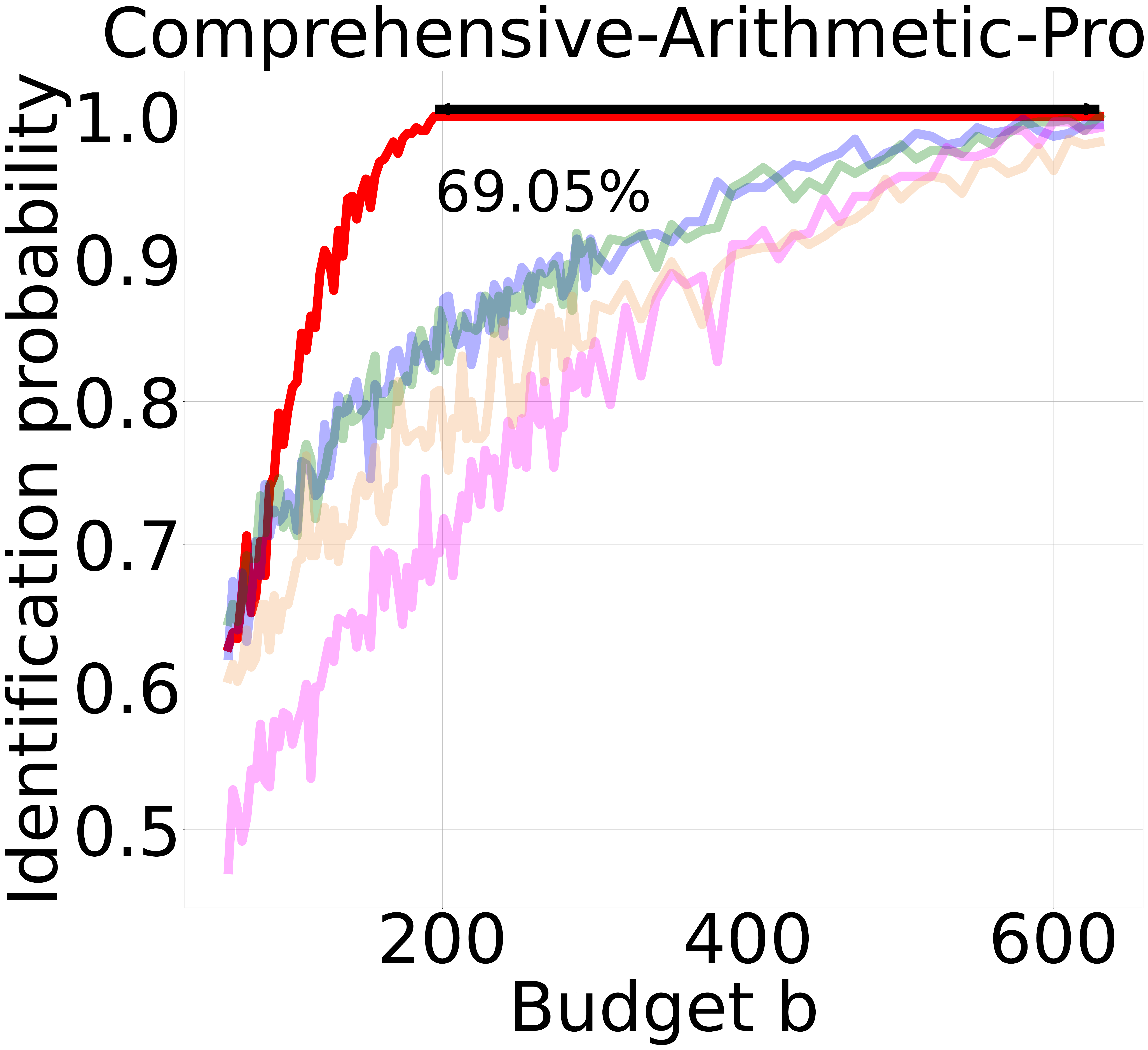}
\end{minipage}\hspace{0pt}%
\begin{minipage}[t]{.19\linewidth}\vspace{0pt}\centering
\includegraphics[width=\linewidth]{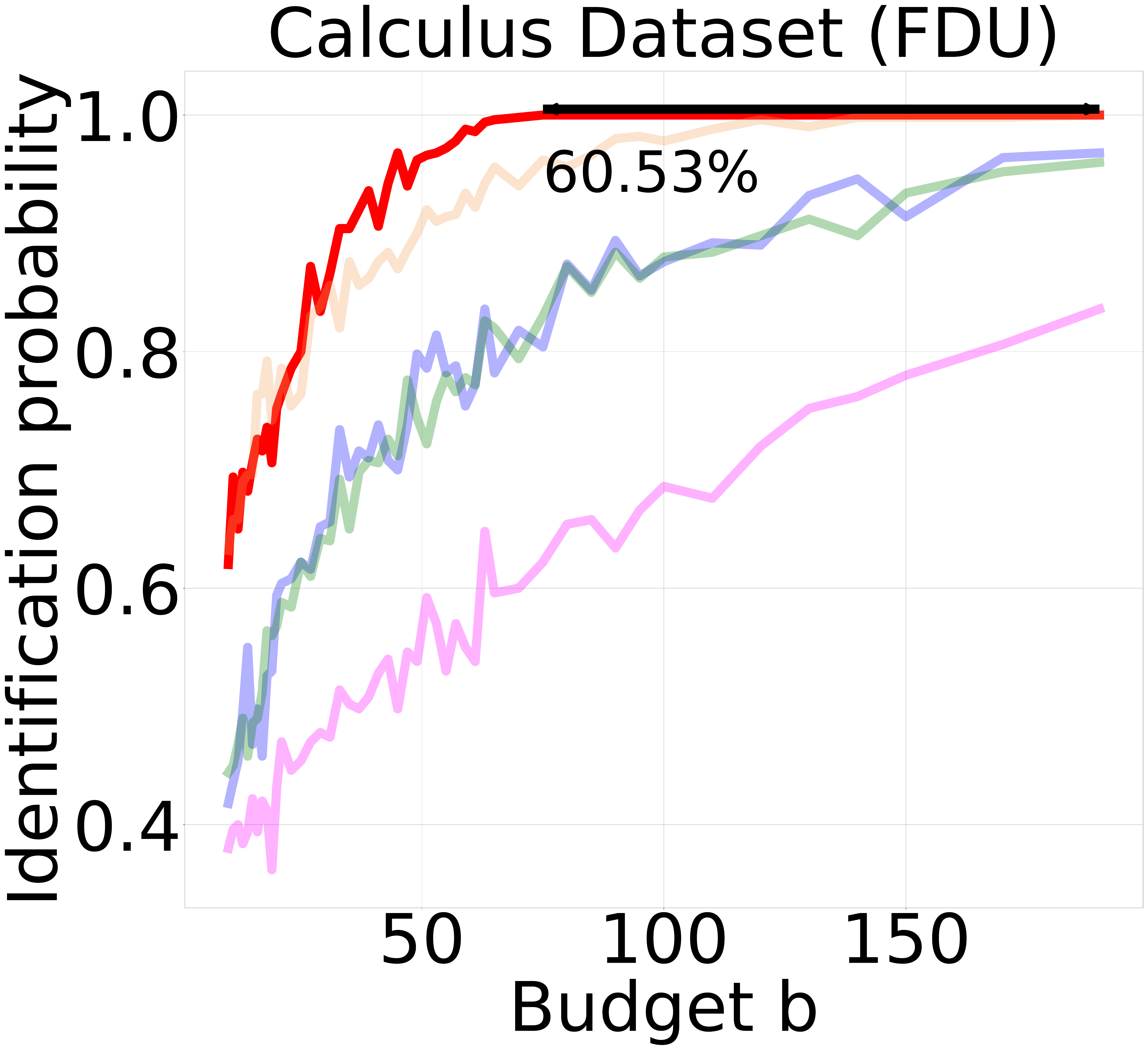}
\end{minipage}\hspace{0pt}%
\begin{minipage}[t]{.19\linewidth}\vspace{0pt}\centering
\includegraphics[width=\linewidth]{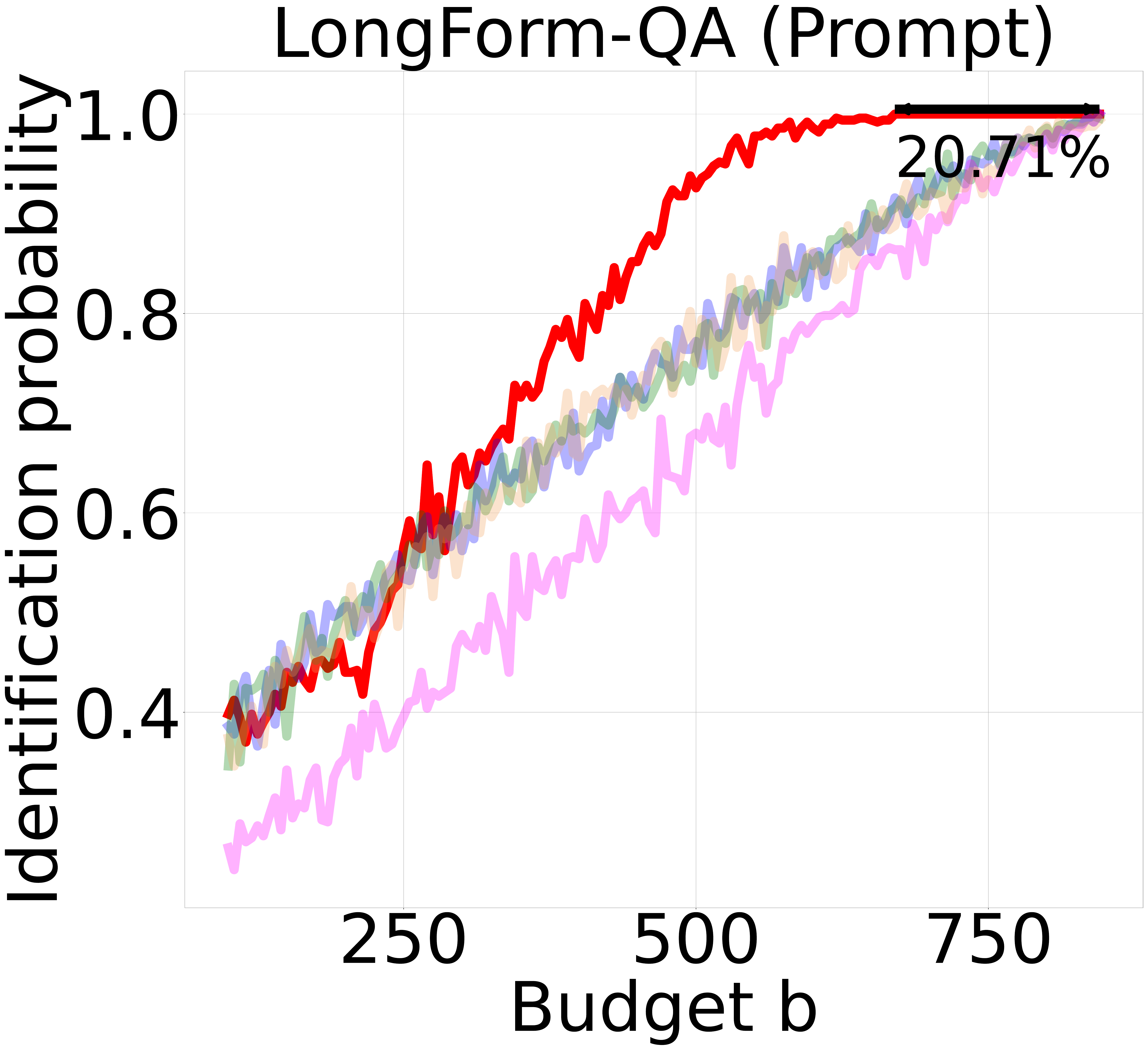}
\end{minipage}\hspace{0pt}%
\begin{minipage}[t]{.19\linewidth}\vspace{0pt}\centering
\includegraphics[width=\linewidth]{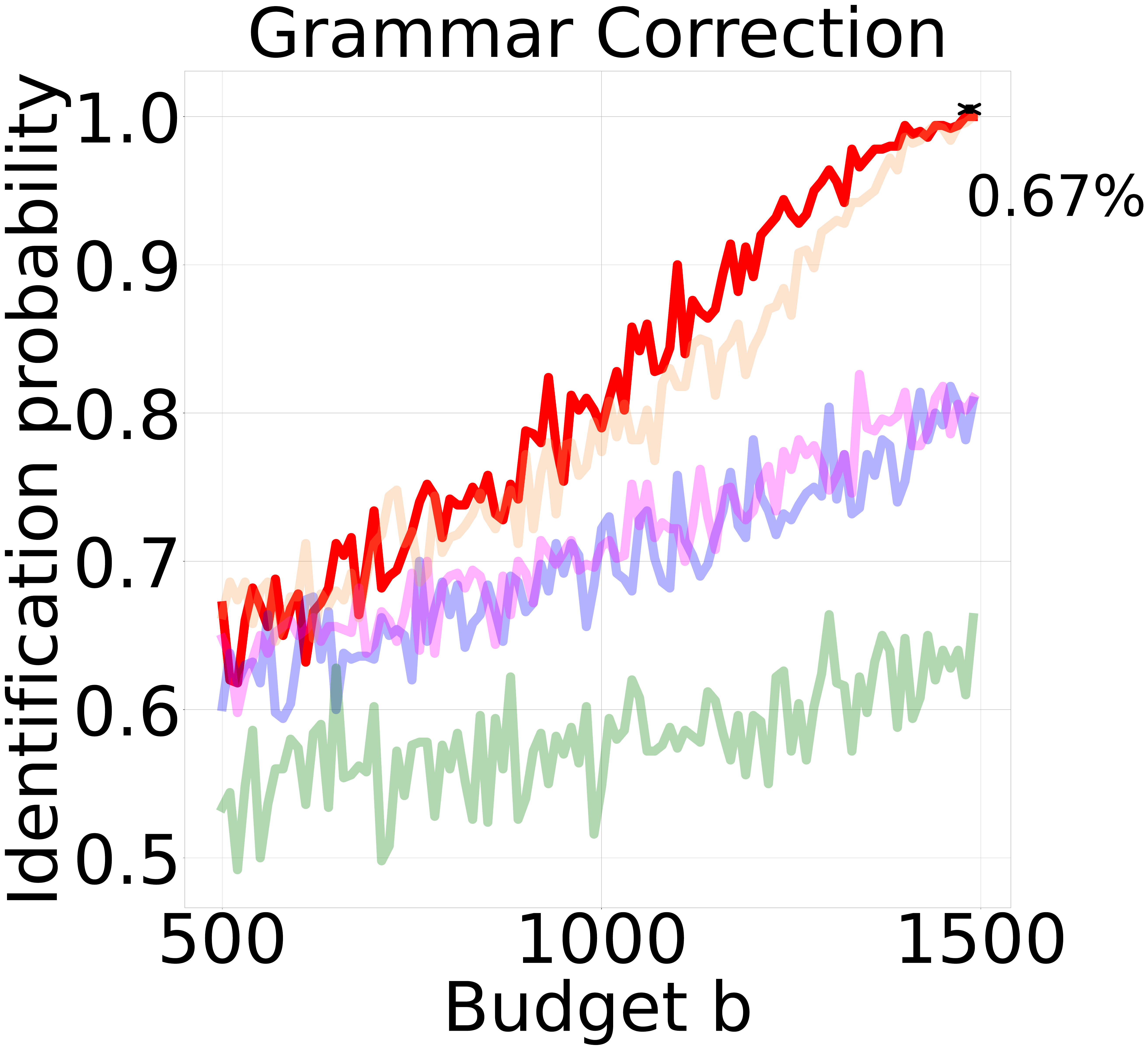}
\end{minipage}\hspace{0pt}%
\begin{minipage}[t]{.19\linewidth}\vspace{0pt}\centering
\includegraphics[width=\linewidth]{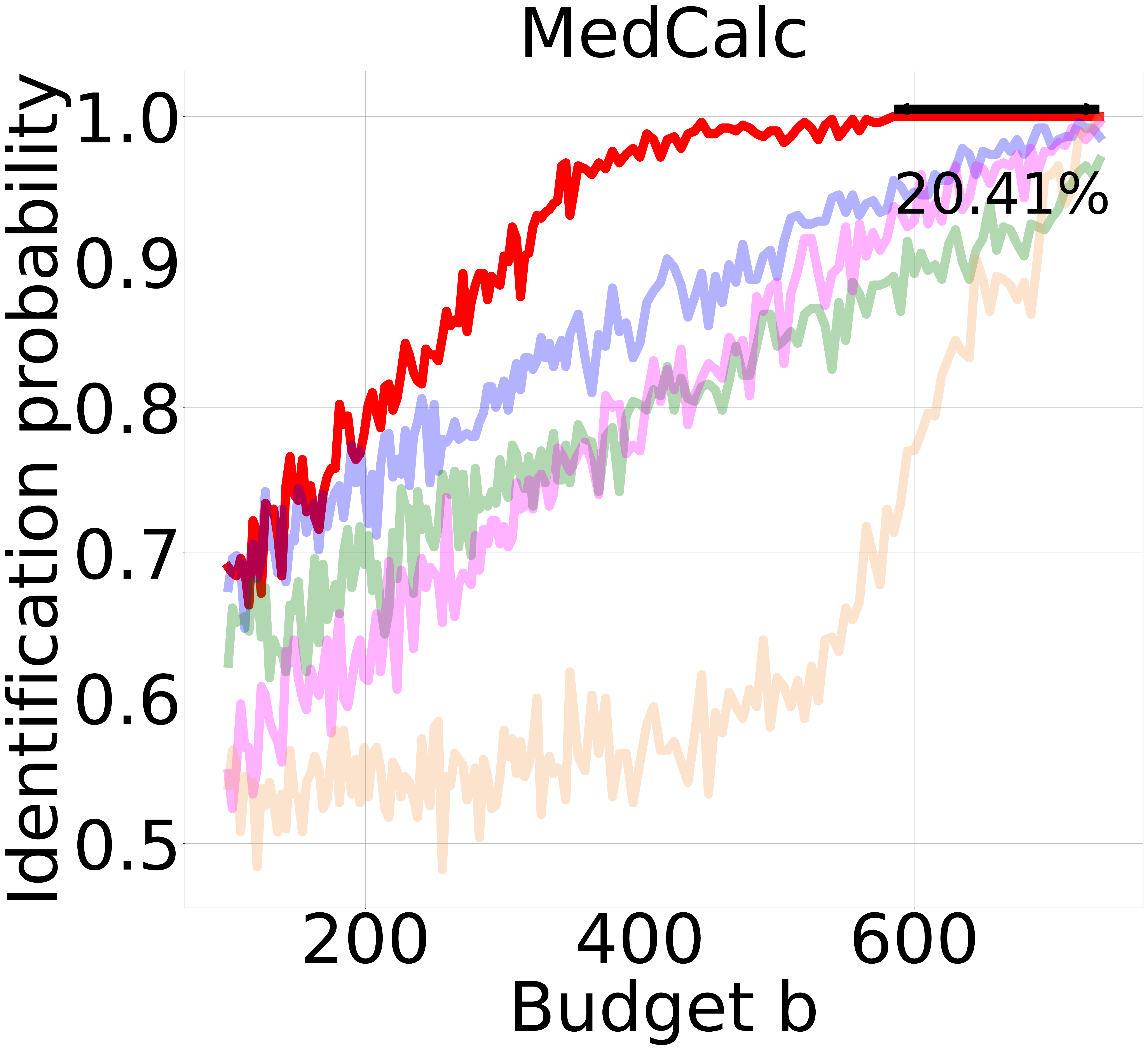}
\end{minipage}

\vspace{2pt}

\begin{minipage}[t]{.19\linewidth}\vspace{0pt}\centering
\includegraphics[width=\linewidth]{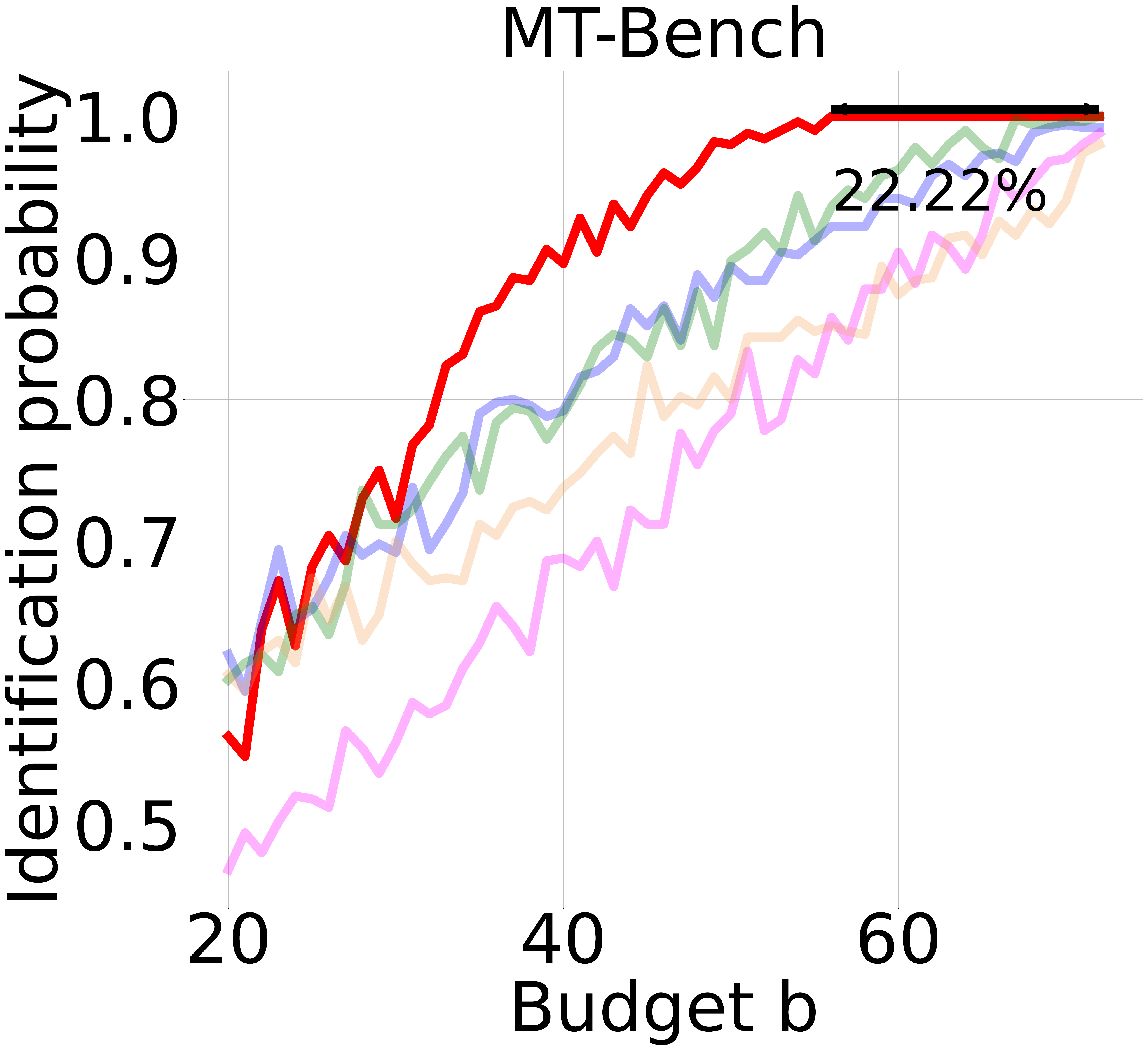}
\end{minipage}\hspace{0pt}%
\begin{minipage}[t]{.19\linewidth}\vspace{0pt}\centering
\includegraphics[width=\linewidth]{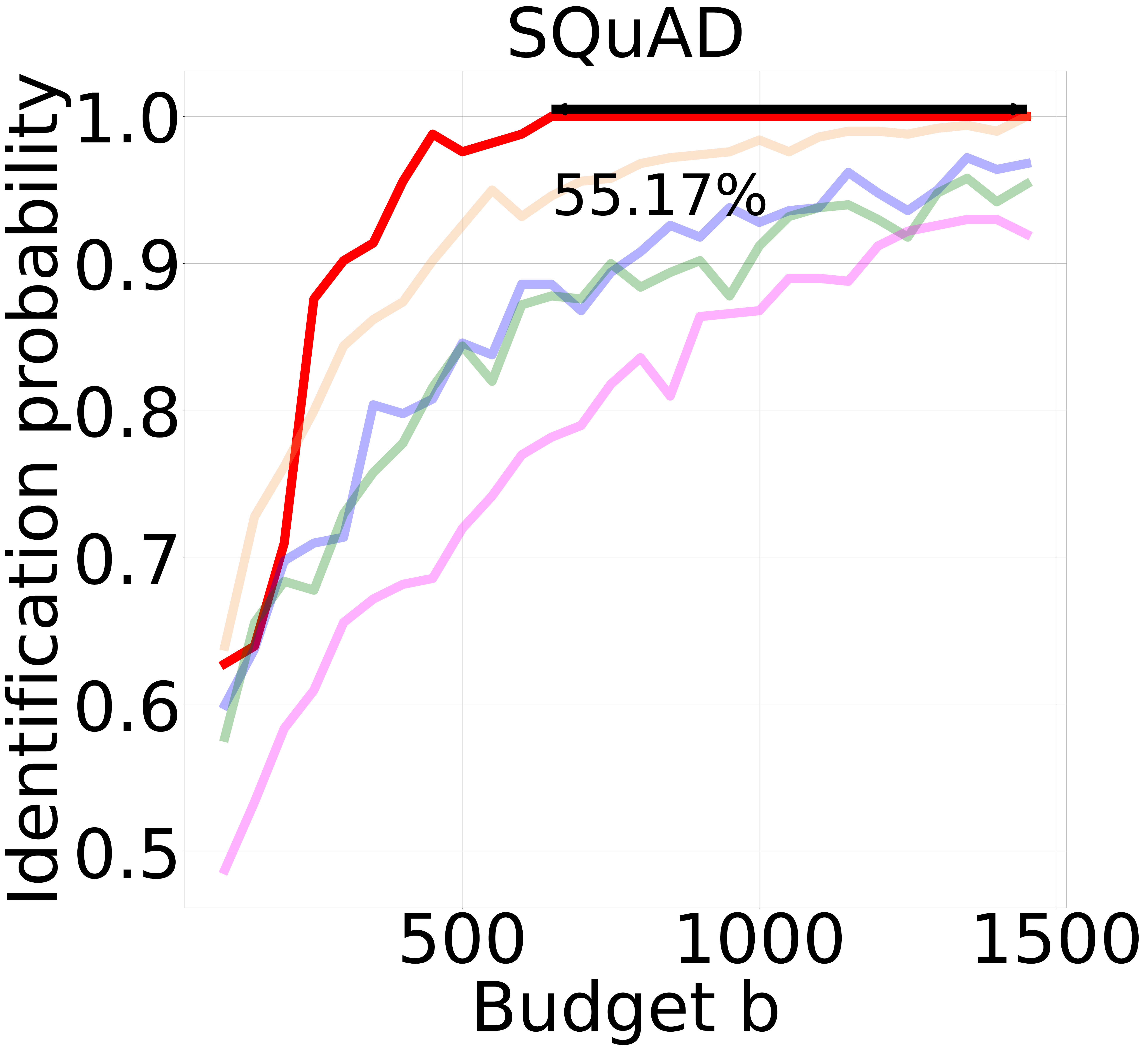}
\end{minipage}\hspace{0pt}%
\begin{minipage}[t]{.19\linewidth}\vspace{0pt}\centering
\includegraphics[width=\linewidth]{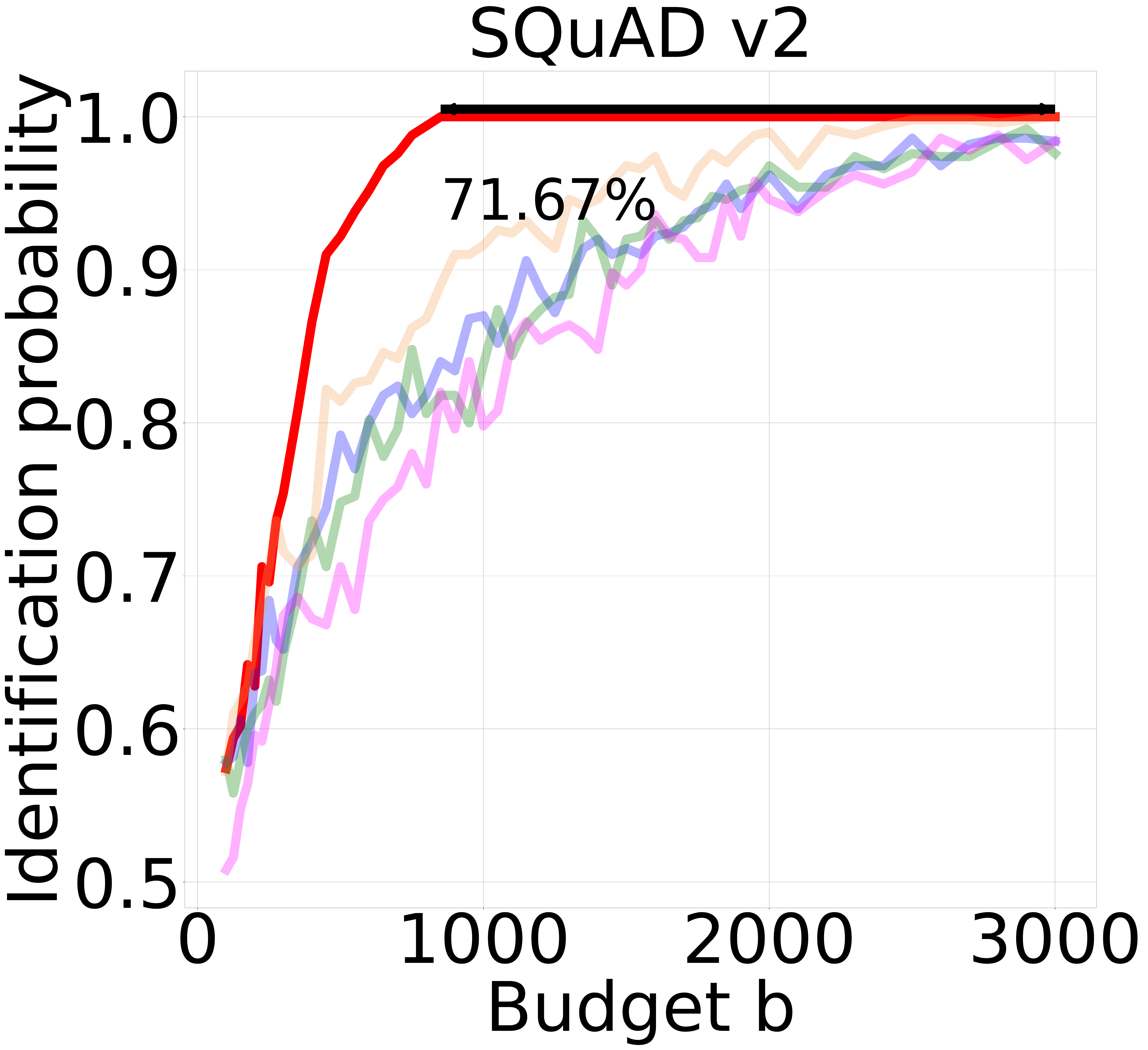}
\end{minipage}\hspace{0pt}%
\begin{minipage}[t]{.19\linewidth}\vspace{0pt}\centering
\includegraphics[width=\linewidth]{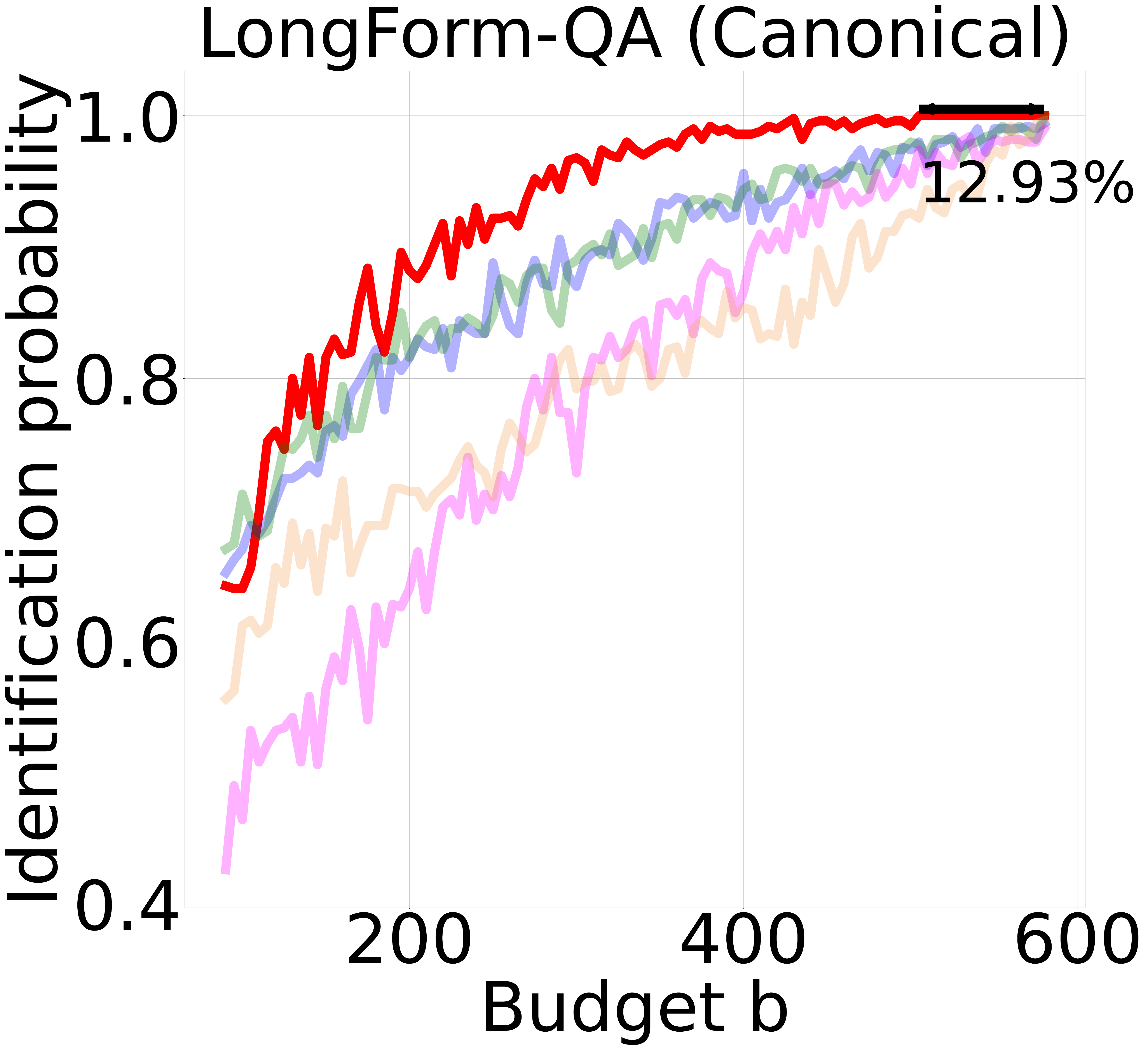}
\end{minipage}\hspace{0pt}%
\begin{minipage}[t]{.19\linewidth}\vspace{0pt}\centering
\includegraphics[width=\linewidth]{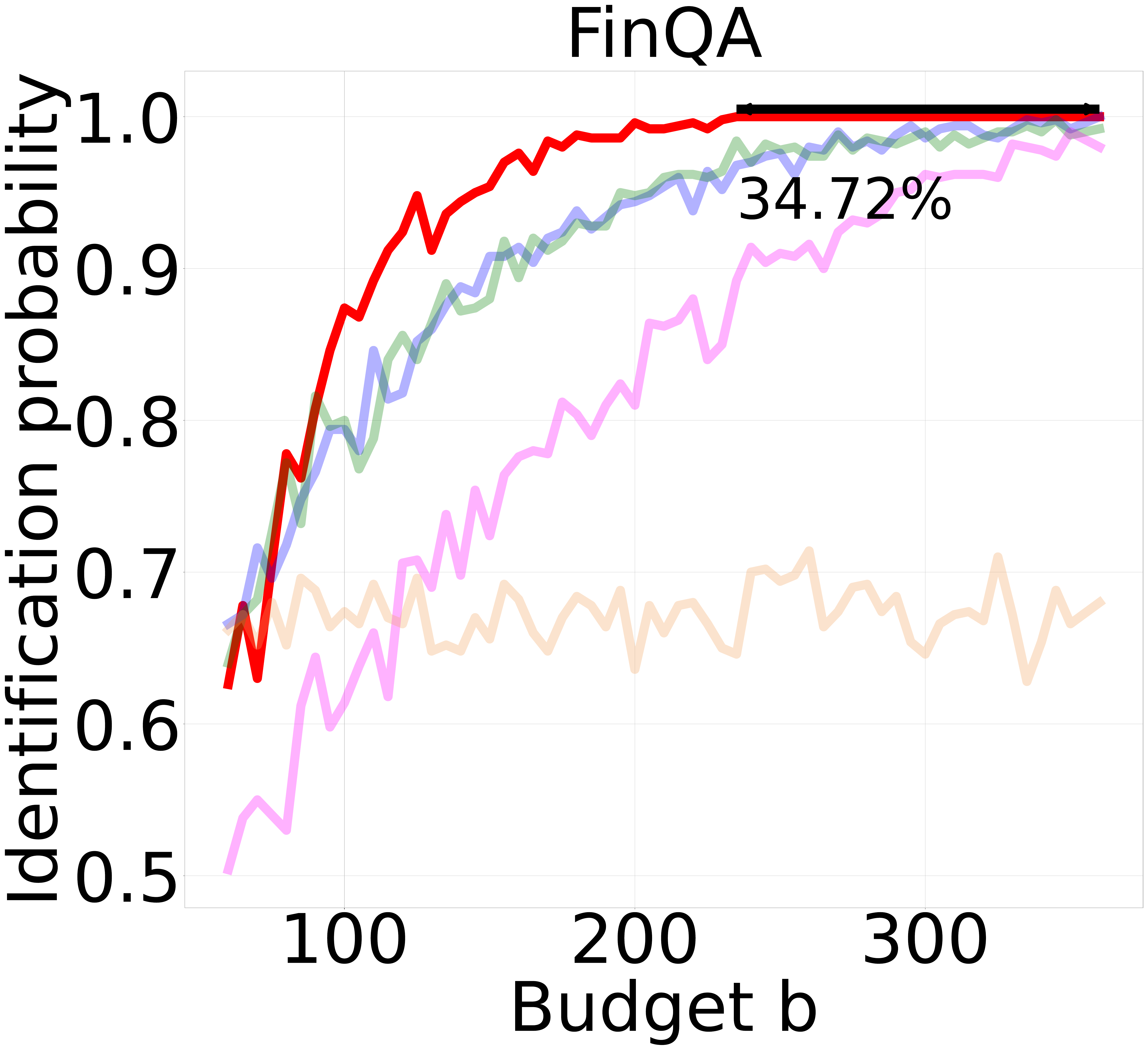}
\end{minipage}

\caption{Identification probabilities of \framework and baselines across different annotation budgets.}
\label{fig:identification_probs}

\end{figure}

Figure \ref{fig:identification_probs} shows the identification probabilities for \framework and the baselines. For each dataset, we extend the annotation budget until the best competing baseline achieves $100\%$ identification probability, and then report the percentage of annotation cost saved by \framework for confidently identifying the best model. Across all datasets and annotation budgets, \framework consistently requires the fewest annotations to reach maximum identification probability, showing its ability in correctly identifying the most informative examples to annotate, saving up to $71.67\%$ with respect to the budget needed by the best competing baseline.

\subsubsection{Annotation Efficiency}
We examine the annotation cost of selecting a best or near-best model with accuracy within the $\delta$-vicinity of that of the true best model across the entire stream. In particular, we focus on the required number of annotations where, in all realizations, the selected models are within 2.5\%, 1\%, 0.5\%, and 0.1\% of the true best model score. 
We evaluate the percentage reduction in annotation cost relative to the best competing baseline, reported separately for each dataset and each $\delta$-vicinity. Our results in Table \ref{tab:delta_vicin} show that \framework significantly reduces the number of annotations required by up to $70.69$\% to identify a near-best model whose accuracy is close to that of the best one. \framework is consistently more annotation-efficient in identifying both the best and a near-best model across different tasks and datasets.

\begin{table}[htbp]
\centering
\footnotesize
\setlength{\tabcolsep}{6pt}
\begin{tabular}{lrrrr}
\toprule
Dataset & $\delta{=}2.5\%$ & $\delta{=}1.0\%$ & $\delta{=}0.5\%$ & $\delta{=}0.1\%$ \\
\midrule
Comprehensive--Arithmetic--Problems & $\downarrow\,\mathbf{24.39}\%$ & $\downarrow\,\mathbf{64.89}\%$ & $\downarrow\,\mathbf{68.00}\%$ & $\downarrow\,\mathbf{69.05}\%$ \\
Calculus Dataset (FDU)            & $\uparrow28.00\%$ & $\downarrow\,\mathbf{53.75}\%$ & $\downarrow\,\mathbf{46.43}\%$ & $\downarrow\,\mathbf{60.53}\%$ \\
LongForm--QA (Prompt)             & $0.00\%$ & $\downarrow\,\mathbf{27.70}\%$ & $\downarrow\,\mathbf{25.79}\%$ & $\downarrow\,\mathbf{20.96}\%$ \\
Grammar Correction                & $0.00\%$ & $0.00\%$ & $0.00\%$ & $\downarrow\,\mathbf{0.67}\%$ \\
MedCalc                           & $\downarrow\,\mathbf{14.66}\%$ & $\downarrow\,\mathbf{38.46}\%$ & $\downarrow\,\mathbf{35.51}\%$ & $\downarrow\,\mathbf{21.09}\%$ \\
MT--Bench                         & $\downarrow\,\mathbf{22.22}\%$ & $\downarrow\,\mathbf{22.22}\%$ & $\downarrow\,\mathbf{22.22}\%$ & $\downarrow\,\mathbf{22.22}\%$ \\
SQuAD                             & $\uparrow20.00\%$ & $\downarrow\,\mathbf{52.00}\%$ & $\downarrow\,\mathbf{55.17}\%$ & $\downarrow\,\mathbf{55.17}\%$ \\
SQuAD v2                          & $0.00\%$ & $\downarrow\,\mathbf{33.33}\%$ & $\downarrow\,\mathbf{63.04}\%$ & $\downarrow\,\mathbf{70.69}\%$ \\
LongForm-QA (Canonical) & $\downarrow\mathbf{30.30}\%$ & $\downarrow\mathbf{27.00}\%$ & $\downarrow\mathbf{33.93}\%$ & $\downarrow\mathbf{23.28}\%$ \\
FinQA & $\downarrow\mathbf{34.69}\%$ & $\downarrow\mathbf{31.88}\%$ & $\downarrow\mathbf{34.72}\%$ & $\downarrow\mathbf{34.72}\%$ \\
\bottomrule

\end{tabular}
\caption{Annotation efficiency for the near-best model.}
\label{tab:delta_vicin}

\end{table}

\section{Discussions}
We introduce the novel problem of online model selection for language models with limited annotation evidence, which is a technically challenging setting given the open-ended nature of generative outputs. We propose \framework, a method tailored to this task that confidently identifies the best language model for the task in an annotation-efficient manner. As LLMs are increasingly deployed in domains where annotation budgets are limited and data distributions shift, \framework enables adaptive model selection that reduces annotation costs, maintains performance under evolving conditions, and improves the robustness of real-world LLM deployment.

\textbf{Ethics statement.} We do not foresee ethical concerns arising from this work. We introduce a novel active model selection algorithm that performs well under budget constraints and compare it with established baselines in the literature. The study uses public benchmarks and does not involve human subjects or sensitive data.

\textbf{Reproducibility statement.} Our results are completely reproducible. We present the rigorous workflow in the paper and report experiments in the main text and the appendix. We provide the complete source code and scripts as supplementary material to reproduce all experiments.

\bibliography{references}

@misc{karimi_online_2021,
	title = {Online {Active} {Model} {Selection} for {Pre}-trained {Classifiers}},
	url = {http://arxiv.org/abs/2010.09818},
	doi = {10.48550/arXiv.2010.09818},
	urldate = {2025-01-14},
	publisher = {arXiv},
	author = {Karimi, Mohammad Reza and Gürel, Nezihe Merve and Karlaš, Bojan and Rausch, Johannes and Zhang, Ce and Krause, Andreas},
	month = apr,
	year = {2021},
	note = {arXiv:2010.09818 [cs]},
}

@inproceedings{lin_rouge_2004,
	address = {Barcelona, Spain},
	title = {{ROUGE}: {A} {Package} for {Automatic} {Evaluation} of {Summaries}},
	shorttitle = {{ROUGE}},
	url = {https://aclanthology.org/W04-1013/},
	urldate = {2025-03-11},
	booktitle = {Text {Summarization} {Branches} {Out}},
	publisher = {Association for Computational Linguistics},
	author = {Lin, Chin-Yew},
	month = jul,
	year = {2004},
	pages = {74--81},
}

@article{littlestone_weighted_1994,
	title = {The {Weighted} {Majority} {Algorithm}},
	volume = {108},
	issn = {0890-5401},
	url = {https://www.sciencedirect.com/science/article/pii/S0890540184710091},
	doi = {10.1006/inco.1994.1009},
	number = {2},
	urldate = {2025-03-18},
	journal = {Information and Computation},
	author = {Littlestone, N. and Warmuth, M. K.},
	month = feb,
	year = {1994},
	pages = {212--261},
}

@misc{rajpurkar_squad_2016,
	title = {{SQuAD}: 100,000+ {Questions} for {Machine} {Comprehension} of {Text}},
	shorttitle = {{SQuAD}},
	url = {http://arxiv.org/abs/1606.05250},
	doi = {10.48550/arXiv.1606.05250},
	urldate = {2025-06-30},
	publisher = {arXiv},
	author = {Rajpurkar, Pranav and Zhang, Jian and Lopyrev, Konstantin and Liang, Percy},
	month = oct,
	year = {2016},
	note = {arXiv:1606.05250 [cs]},
}

@misc{rajpurkar_know_2018,
	title = {Know {What} {You} {Don}'t {Know}: {Unanswerable} {Questions} for {SQuAD}},
	shorttitle = {Know {What} {You} {Don}'t {Know}},
	url = {http://arxiv.org/abs/1806.03822},
	doi = {10.48550/arXiv.1806.03822},
	urldate = {2025-06-30},
	publisher = {arXiv},
	author = {Rajpurkar, Pranav and Jia, Robin and Liang, Percy},
	month = jun,
	year = {2018},
	note = {arXiv:1806.03822 [cs]},
}

@misc{brown_language_2020,
	title = {Language {Models} are {Few}-{Shot} {Learners}},
	url = {http://arxiv.org/abs/2005.14165},
	doi = {10.48550/arXiv.2005.14165},
	urldate = {2025-08-20},
	publisher = {arXiv},
	author = {Brown, Tom B. and Mann, Benjamin and Ryder, Nick and Subbiah, Melanie and Kaplan, Jared and Dhariwal, Prafulla and Neelakantan, Arvind and Shyam, Pranav and Sastry, Girish and Askell, Amanda and Agarwal, Sandhini and Herbert-Voss, Ariel and Krueger, Gretchen and Henighan, Tom and Child, Rewon and Ramesh, Aditya and Ziegler, Daniel M. and Wu, Jeffrey and Winter, Clemens and Hesse, Christopher and Chen, Mark and Sigler, Eric and Litwin, Mateusz and Gray, Scott and Chess, Benjamin and Clark, Jack and Berner, Christopher and McCandlish, Sam and Radford, Alec and Sutskever, Ilya and Amodei, Dario},
	month = jul,
	year = {2020},
	note = {arXiv:2005.14165 [cs]},
}

@misc{kojima_large_2023,
	title = {Large {Language} {Models} are {Zero}-{Shot} {Reasoners}},
	url = {http://arxiv.org/abs/2205.11916},
	doi = {10.48550/arXiv.2205.11916},
	urldate = {2025-08-20},
	publisher = {arXiv},
	author = {Kojima, Takeshi and Gu, Shixiang Shane and Reid, Machel and Matsuo, Yutaka and Iwasawa, Yusuke},
	month = jan,
	year = {2023},
	note = {arXiv:2205.11916 [cs]},
}

@misc{dong_survey_2024,
	title = {A {Survey} on {In}-context {Learning}},
	url = {http://arxiv.org/abs/2301.00234},
	doi = {10.48550/arXiv.2301.00234},
	urldate = {2025-08-20},
	publisher = {arXiv},
	author = {Dong, Qingxiu and Li, Lei and Dai, Damai and Zheng, Ce and Ma, Jingyuan and Li, Rui and Xia, Heming and Xu, Jingjing and Wu, Zhiyong and Liu, Tianyu and Chang, Baobao and Sun, Xu and Li, Lei and Sui, Zhifang},
	month = oct,
	year = {2024},
	note = {arXiv:2301.00234 [cs]},
}

@misc{liu_pre-train_2021,
	title = {Pre-train, {Prompt}, and {Predict}: {A} {Systematic} {Survey} of {Prompting} {Methods} in {Natural} {Language} {Processing}},
	shorttitle = {Pre-train, {Prompt}, and {Predict}},
	url = {http://arxiv.org/abs/2107.13586},
	doi = {10.48550/arXiv.2107.13586},
	urldate = {2025-08-20},
	publisher = {arXiv},
	author = {Liu, Pengfei and Yuan, Weizhe and Fu, Jinlan and Jiang, Zhengbao and Hayashi, Hiroaki and Neubig, Graham},
	month = jul,
	year = {2021},
	note = {arXiv:2107.13586 [cs]},
}

@misc{boll_distillnote_2025,
	title = {{DistillNote}: {LLM}-based clinical note summaries improve heart failure diagnosis},
	shorttitle = {{DistillNote}},
	url = {http://arxiv.org/abs/2506.16777},
	doi = {10.48550/arXiv.2506.16777},
	language = {en},
	urldate = {2025-08-20},
	publisher = {arXiv},
	author = {Boll, Heloisa Oss and Boll, Antonio Oss and Boll, Leticia Puttlitz and Hanna, Ameen Abu and Calixto, Iacer},
	month = jun,
	year = {2025},
	note = {arXiv:2506.16777 [cs]},
}

@article{veen_adapted_2024,
	title = {Adapted {Large} {Language} {Models} {Can} {Outperform} {Medical} {Experts} in {Clinical} {Text} {Summarization}},
	volume = {30},
	issn = {1078-8956, 1546-170X},
	url = {http://arxiv.org/abs/2309.07430},
	doi = {10.1038/s41591-024-02855-5},
	number = {4},
	urldate = {2025-08-20},
	journal = {Nature Medicine},
	author = {Veen, Dave Van and Uden, Cara Van and Blankemeier, Louis and Delbrouck, Jean-Benoit and Aali, Asad and Bluethgen, Christian and Pareek, Anuj and Polacin, Malgorzata and Reis, Eduardo Pontes and Seehofnerova, Anna and Rohatgi, Nidhi and Hosamani, Poonam and Collins, William and Ahuja, Neera and Langlotz, Curtis P. and Hom, Jason and Gatidis, Sergios and Pauly, John and Chaudhari, Akshay S.},
	month = apr,
	year = {2024},
	note = {arXiv:2309.07430 [cs]},
	pages = {1134--1142},
}

@article{biswas_intelligent_2024,
	title = {Intelligent {Clinical} {Documentation}: {Harnessing} {Generative} {AI} for {Patient}-{Centric} {Clinical} {Note} {Generation}},
	issn = {2456-2165},
	shorttitle = {Intelligent {Clinical} {Documentation}},
	url = {http://arxiv.org/abs/2405.18346},
	doi = {10.38124/ijisrt/IJISRT24MAY1483},
	urldate = {2025-08-20},
	journal = {International Journal of Innovative Science and Research Technology (IJISRT)},
	author = {Biswas, Anjanava and Talukdar, Wrick},
	month = may,
	year = {2024},
	note = {arXiv:2405.18346 [cs]},
	pages = {994--1008},
}

@inproceedings{angelopoulos_industrial_2025,
	address = {Cham},
	title = {Industrial {Maintenance} {Optimization} {Based} on the {Integration} of {Large} {Language} {Models} ({LLM}) and {Augmented} {Reality} ({AR})},
	isbn = {978-3-031-86489-6},
	doi = {10.1007/978-3-031-86489-6_20},
	language = {en},
	booktitle = {Advances in {Artificial} {Intelligence} in {Manufacturing} {II}},
	publisher = {Springer Nature Switzerland},
	author = {Angelopoulos, John and Manettas, Christos and Alexopoulos, Kosmas},
	editor = {Alexopoulos, Kosmas and Makris, Sotiris and Stavropoulos, Panagiotis},
	year = {2025},
	pages = {197--205},
}

@misc{kok_when_2024,
	title = {When {IoT} {Meet} {LLMs}: {Applications} and {Challenges}},
	shorttitle = {When {IoT} {Meet} {LLMs}},
	url = {http://arxiv.org/abs/2411.17722},
	doi = {10.48550/arXiv.2411.17722},
	urldate = {2025-08-20},
	publisher = {arXiv},
	author = {Kok, Ibrahim and Demirci, Orhan and Ozdemir, Suat},
	month = nov,
	year = {2024},
	note = {arXiv:2411.17722 [cs]},
}

@misc{li_large_2024,
	title = {Large {Language} {Models} for {Manufacturing}},
	url = {http://arxiv.org/abs/2410.21418},
	doi = {10.48550/arXiv.2410.21418},
	urldate = {2025-08-20},
	publisher = {arXiv},
	author = {Li, Yiwei and Zhao, Huaqin and Jiang, Hanqi and Pan, Yi and Liu, Zhengliang and Wu, Zihao and Shu, Peng and Tian, Jie and Yang, Tianze and Xu, Shaochen and Lyu, Yanjun and Blenk, Parker and Pence, Jacob and Rupram, Jason and Banu, Eliza and Liu, Ninghao and Wang, Linbing and Song, Wenzhan and Zhai, Xiaoming and Song, Kenan and Zhu, Dajiang and Li, Beiwen and Wang, Xianqiao and Liu, Tianming},
	month = oct,
	year = {2024},
	note = {arXiv:2410.21418 [cs]
version: 1},
}

@misc{dan_educhat_2023,
	title = {{EduChat}: {A} {Large}-{Scale} {Language} {Model}-based {Chatbot} {System} for {Intelligent} {Education}},
	shorttitle = {{EduChat}},
	url = {http://arxiv.org/abs/2308.02773},
	doi = {10.48550/arXiv.2308.02773},
	urldate = {2025-08-20},
	publisher = {arXiv},
	author = {Dan, Yuhao and Lei, Zhikai and Gu, Yiyang and Li, Yong and Yin, Jianghao and Lin, Jiaju and Ye, Linhao and Tie, Zhiyan and Zhou, Yougen and Wang, Yilei and Zhou, Aimin and Zhou, Ze and Chen, Qin and Zhou, Jie and He, Liang and Qiu, Xipeng},
	month = aug,
	year = {2023},
	note = {arXiv:2308.02773 [cs]},
}

@misc{caines_application_2023,
	title = {On the application of {Large} {Language} {Models} for language teaching and assessment technology},
	url = {http://arxiv.org/abs/2307.08393},
	doi = {10.48550/arXiv.2307.08393},
	urldate = {2025-08-20},
	publisher = {arXiv},
	author = {Caines, Andrew and Benedetto, Luca and Taslimipoor, Shiva and Davis, Christopher and Gao, Yuan and Andersen, Oeistein and Yuan, Zheng and Elliott, Mark and Moore, Russell and Bryant, Christopher and Rei, Marek and Yannakoudakis, Helen and Mullooly, Andrew and Nicholls, Diane and Buttery, Paula},
	month = jul,
	year = {2023},
	note = {arXiv:2307.08393 [cs]},
}

@misc{pont_legal_2023,
	title = {Legal {Summarisation} through {LLMs}: {The} {PRODIGIT} {Project}},
	shorttitle = {Legal {Summarisation} through {LLMs}},
	url = {http://arxiv.org/abs/2308.04416},
	doi = {10.48550/arXiv.2308.04416},
	urldate = {2025-08-20},
	publisher = {arXiv},
	author = {Pont, Thiago Dal and Galli, Federico and Loreggia, Andrea and Pisano, Giuseppe and Rovatti, Riccardo and Sartor, Giovanni},
	month = aug,
	year = {2023},
	note = {arXiv:2308.04416 [cs]},
}

@misc{lee_lexgpt_2023,
	title = {{LexGPT} 0.1: pre-trained {GPT}-{J} models with {Pile} of {Law}},
	shorttitle = {{LexGPT} 0.1},
	url = {http://arxiv.org/abs/2306.05431},
	doi = {10.48550/arXiv.2306.05431},
	urldate = {2025-08-20},
	publisher = {arXiv},
	author = {Lee, Jieh-Sheng},
	month = jun,
	year = {2023},
	note = {arXiv:2306.05431 [cs]},
}

@misc{guha_legalbench_2023,
	title = {{LegalBench}: {A} {Collaboratively} {Built} {Benchmark} for {Measuring} {Legal} {Reasoning} in {Large} {Language} {Models}},
	shorttitle = {{LegalBench}},
	url = {http://arxiv.org/abs/2308.11462},
	doi = {10.48550/arXiv.2308.11462},
	urldate = {2025-08-20},
	publisher = {arXiv},
	author = {Guha, Neel and Nyarko, Julian and Ho, Daniel E. and Ré, Christopher and Chilton, Adam and Narayana, Aditya and Chohlas-Wood, Alex and Peters, Austin and Waldon, Brandon and Rockmore, Daniel N. and Zambrano, Diego and Talisman, Dmitry and Hoque, Enam and Surani, Faiz and Fagan, Frank and Sarfaty, Galit and Dickinson, Gregory M. and Porat, Haggai and Hegland, Jason and Wu, Jessica and Nudell, Joe and Niklaus, Joel and Nay, John and Choi, Jonathan H. and Tobia, Kevin and Hagan, Margaret and Ma, Megan and Livermore, Michael and Rasumov-Rahe, Nikon and Holzenberger, Nils and Kolt, Noam and Henderson, Peter and Rehaag, Sean and Goel, Sharad and Gao, Shang and Williams, Spencer and Gandhi, Sunny and Zur, Tom and Iyer, Varun and Li, Zehua},
	month = aug,
	year = {2023},
	note = {arXiv:2308.11462 [cs]},
}

@misc{liu_contextual_2022,
	title = {Contextual {Active} {Model} {Selection}},
	url = {http://arxiv.org/abs/2207.06030},
	doi = {10.48550/arXiv.2207.06030},
	urldate = {2025-08-20},
	publisher = {arXiv},
	author = {Liu, Xuefeng and Xia, Fangfang and Stevens, Rick L. and Chen, Yuxin},
	month = jul,
	year = {2022},
	note = {arXiv:2207.06030 [cs]},
}

@misc{li_online_2024,
	title = {Online {Foundation} {Model} {Selection} in {Robotics}},
	url = {http://arxiv.org/abs/2402.08570},
	doi = {10.48550/arXiv.2402.08570},
	urldate = {2025-08-20},
	publisher = {arXiv},
	author = {Li, Po-han and Toprak, Oyku Selin and Narayanan, Aditya and Topcu, Ufuk and Chinchali, Sandeep},
	month = feb,
	year = {2024},
	note = {arXiv:2402.08570 [cs]},
}

@misc{ouyang_training_2022,
	title = {Training language models to follow instructions with human feedback},
	url = {http://arxiv.org/abs/2203.02155},
	doi = {10.48550/arXiv.2203.02155},
	urldate = {2025-08-20},
	publisher = {arXiv},
	author = {Ouyang, Long and Wu, Jeff and Jiang, Xu and Almeida, Diogo and Wainwright, Carroll L. and Mishkin, Pamela and Zhang, Chong and Agarwal, Sandhini and Slama, Katarina and Ray, Alex and Schulman, John and Hilton, Jacob and Kelton, Fraser and Miller, Luke and Simens, Maddie and Askell, Amanda and Welinder, Peter and Christiano, Paul and Leike, Jan and Lowe, Ryan},
	month = mar,
	year = {2022},
	note = {arXiv:2203.02155 [cs]},
}

@article{kocon_chatgpt_2023,
	title = {{ChatGPT}: {Jack} of all trades, master of none},
	volume = {99},
	issn = {15662535},
	shorttitle = {{ChatGPT}},
	url = {http://arxiv.org/abs/2302.10724},
	doi = {10.1016/j.inffus.2023.101861},
	urldate = {2025-08-20},
	journal = {Information Fusion},
	author = {Kocoń, Jan and Cichecki, Igor and Kaszyca, Oliwier and Kochanek, Mateusz and Szydło, Dominika and Baran, Joanna and Bielaniewicz, Julita and Gruza, Marcin and Janz, Arkadiusz and Kanclerz, Kamil and Kocoń, Anna and Koptyra, Bartłomiej and Mieleszczenko-Kowszewicz, Wiktoria and Miłkowski, Piotr and Oleksy, Marcin and Piasecki, Maciej and Radliński, Łukasz and Wojtasik, Konrad and Woźniak, Stanisław and Kazienko, Przemysław},
	month = nov,
	year = {2023},
	note = {arXiv:2302.10724 [cs]},
	pages = {101861},
}

@inproceedings{gardner_bayesian_2015,
	title = {Bayesian {Active} {Model} {Selection} with an {Application} to {Automated} {Audiometry}},
	volume = {28},
	url = {https://papers.nips.cc/paper_files/paper/2015/hash/d9731321ef4e063ebbee79298fa36f56-Abstract.html},
	urldate = {2025-08-20},
	booktitle = {Advances in {Neural} {Information} {Processing} {Systems}},
	publisher = {Curran Associates, Inc.},
	author = {Gardner, Jacob and Malkomes, Gustavo and Garnett, Roman and Weinberger, Kilian Q and Barbour, Dennis and Cunningham, John P},
	year = {2015},
}

@misc{russo_learning_2017,
	title = {Learning to {Optimize} via {Information}-{Directed} {Sampling}},
	url = {http://arxiv.org/abs/1403.5556},
	doi = {10.48550/arXiv.1403.5556},
	urldate = {2025-08-20},
	publisher = {arXiv},
	author = {Russo, Daniel and Roy, Benjamin Van},
	month = jul,
	year = {2017},
	note = {arXiv:1403.5556 [cs]},
}

@article{lindley_measure_1956,
	title = {On a {Measure} of the {Information} {Provided} by an {Experiment}},
	volume = {27},
	issn = {0003-4851},
	url = {http://projecteuclid.org/euclid.aoms/1177728069},
	doi = {10.1214/aoms/1177728069},
	language = {en},
	number = {4},
	urldate = {2025-08-30},
	journal = {The Annals of Mathematical Statistics},
	author = {Lindley, D. V.},
	month = dec,
	year = {1956},
	pages = {986--1005},
}

@article{mackay_information-based_1992,
	title = {Information-{Based} {Objective} {Functions} for {Active} {Data} {Selection}},
	volume = {4},
	issn = {0899-7667},
	url = {https://doi.org/10.1162/neco.1992.4.4.590},
	doi = {10.1162/neco.1992.4.4.590},
	number = {4},
	urldate = {2025-08-30},
	journal = {Neural Computation},
	author = {MacKay, David J. C.},
	month = jul,
	year = {1992},
	pages = {590--604},
}

@article{russo_information-theoretic_2016,
	title = {An {Information}-{Theoretic} {Analysis} of {Thompson} {Sampling}},
	volume = {17},
	issn = {1533-7928},
	url = {http://jmlr.org/papers/v17/14-087.html},
	number = {68},
	urldate = {2025-08-30},
	journal = {Journal of Machine Learning Research},
	author = {Russo, Daniel and Roy, Benjamin Van},
	year = {2016},
	pages = {1--30},
}

@misc{rooij_follow_2013,
	title = {Follow the {Leader} {If} {You} {Can}, {Hedge} {If} {You} {Must}},
	url = {http://arxiv.org/abs/1301.0534},
	doi = {10.48550/arXiv.1301.0534},
	urldate = {2025-08-30},
	publisher = {arXiv},
	author = {Rooij, Steven de and Erven, Tim van and Grünwald, Peter D. and Koolen, Wouter M.},
	month = jan,
	year = {2013},
	note = {arXiv:1301.0534 [cs]},
}

@misc{kingma_adam_2017,
	title = {Adam: {A} {Method} for {Stochastic} {Optimization}},
	shorttitle = {Adam},
	url = {http://arxiv.org/abs/1412.6980},
	doi = {10.48550/arXiv.1412.6980},
	urldate = {2025-08-30},
	publisher = {arXiv},
	author = {Kingma, Diederik P. and Ba, Jimmy},
	month = jan,
	year = {2017},
	note = {arXiv:1412.6980 [cs]},
}

@misc{zhang_bertscore_2020,
	title = {{BERTScore}: {Evaluating} {Text} {Generation} with {BERT}},
	shorttitle = {{BERTScore}},
	url = {http://arxiv.org/abs/1904.09675},
	doi = {10.48550/arXiv.1904.09675},
	urldate = {2025-08-30},
	publisher = {arXiv},
	author = {Zhang, Tianyi and Kishore, Varsha and Wu, Felix and Weinberger, Kilian Q. and Artzi, Yoav},
	month = feb,
	year = {2020},
	note = {arXiv:1904.09675 [cs]},
}

@article{sebastiani_maximum_2025,
	title = {Maximum {Entropy} {Sampling} and {Optimal} {Bayesian} {Experimental} {Design} {\textbar} {Request} {PDF}},
	issn = {1467-9868},
	url = {https://www.researchgate.net/publication/4771982_Maximum_Entropy_Sampling_and_Optimal_Bayesian_Experimental_Design},
	language = {en},
	urldate = {2025-08-31},
	journal = {ResearchGate},
	author = {Sebastiani, Paola and Wynn, Henry P},
	month = aug,
	year = {2025},
}

@inproceedings{ienco_high_2014,
	title = {High density-focused uncertainty sampling for active learning over evolving stream data},
	url = {https://proceedings.mlr.press/v36/ienco14.html},
	language = {en},
	urldate = {2025-08-31},
	booktitle = {Proceedings of the 3rd {International} {Workshop} on {Big} {Data}, {Streams} and {Heterogeneous} {Source} {Mining}: {Algorithms}, {Systems}, {Programming} {Models} and {Applications}},
	publisher = {PMLR},
	author = {Ienco, Dino and Žliobaitė, Indrė and Pfahringer, Bernhard},
	month = aug,
	year = {2014},
	note = {ISSN: 1938-7228},
	pages = {133--148},
}

@misc{katragadda_active_2023,
	title = {Active {Learning} with {Combinatorial} {Coverage}},
	url = {http://arxiv.org/abs/2302.14567},
	doi = {10.48550/arXiv.2302.14567},
	urldate = {2025-08-31},
	publisher = {arXiv},
	author = {Katragadda, Sai Prathyush and Cody, Tyler and Beling, Peter and Freeman, Laura},
	month = feb,
	year = {2023},
	note = {arXiv:2302.14567 [cs]},
}

@inproceedings{mccallum_employing_1998,
	address = {San Francisco, CA, USA},
	series = {{ICML} '98},
	title = {Employing {EM} and {Pool}-{Based} {Active} {Learning} for {Text} {Classification}},
	isbn = {978-1-55860-556-5},
	urldate = {2025-08-31},
	booktitle = {Proceedings of the {Fifteenth} {International} {Conference} on {Machine} {Learning}},
	publisher = {Morgan Kaufmann Publishers Inc.},
	author = {McCallum, Andrew and Nigam, Kamal},
	year = {1998},
	pages = {350--358},
}

@inproceedings{zhu_active_2008,
	address = {USA},
	series = {{COLING} '08},
	title = {Active learning with sampling by uncertainty and density for word sense disambiguation and text classification},
	isbn = {978-1-905593-44-6},
	urldate = {2025-08-31},
	booktitle = {Proceedings of the 22nd {International} {Conference} on {Computational} {Linguistics} - {Volume} 1},
	publisher = {Association for Computational Linguistics},
	author = {Zhu, Jingbo and Wang, Huizhen and Yao, Tianshun and Tsou, Benjamin K.},
	year = {2008},
	pages = {1137--1144},
}

@misc{almazrouei_falcon_2023,
	title = {The {Falcon} {Series} of {Open} {Language} {Models}},
	url = {http://arxiv.org/abs/2311.16867},
	doi = {10.48550/arXiv.2311.16867},
	urldate = {2025-08-31},
	publisher = {arXiv},
	author = {Almazrouei, Ebtesam and Alobeidli, Hamza and Alshamsi, Abdulaziz and Cappelli, Alessandro and Cojocaru, Ruxandra and Debbah, Mérouane and Goffinet, Étienne and Hesslow, Daniel and Launay, Julien and Malartic, Quentin and Mazzotta, Daniele and Noune, Badreddine and Pannier, Baptiste and Penedo, Guilherme},
	month = nov,
	year = {2023},
	note = {arXiv:2311.16867 [cs]},
}

@inproceedings{radford_language_2019,
	title = {Language {Models} are {Unsupervised} {Multitask} {Learners}},
	url = {https://www.semanticscholar.org/paper/Language-Models-are-Unsupervised-Multitask-Learners-Radford-Wu/9405cc0d6169988371b2755e573cc28650d14dfe},
	urldate = {2025-08-31},
	author = {Radford, Alec and Wu, Jeff and Child, R. and Luan, D. and Amodei, Dario and Sutskever, I.},
	year = {2019},
}

@misc{raffel_exploring_2023,
	title = {Exploring the {Limits} of {Transfer} {Learning} with a {Unified} {Text}-to-{Text} {Transformer}},
	url = {http://arxiv.org/abs/1910.10683},
	doi = {10.48550/arXiv.1910.10683},
	urldate = {2025-08-31},
	publisher = {arXiv},
	author = {Raffel, Colin and Shazeer, Noam and Roberts, Adam and Lee, Katherine and Narang, Sharan and Matena, Michael and Zhou, Yanqi and Li, Wei and Liu, Peter J.},
	month = sep,
	year = {2023},
	note = {arXiv:1910.10683 [cs]},
}

@misc{openai_gpt-4_2024,
	title = {{GPT}-4 {Technical} {Report}},
	url = {http://arxiv.org/abs/2303.08774},
	doi = {10.48550/arXiv.2303.08774},
	urldate = {2025-08-31},
	publisher = {arXiv},
	author = {OpenAI and Achiam, Josh and Adler, Steven and Agarwal, Sandhini and Ahmad, Lama and Akkaya, Ilge and Aleman, Florencia Leoni and Almeida, Diogo and Altenschmidt, Janko and Altman, Sam and Anadkat, Shyamal and Avila, Red and Babuschkin, Igor and Balaji, Suchir and Balcom, Valerie and Baltescu, Paul and Bao, Haiming and Bavarian, Mohammad and Belgum, Jeff and Bello, Irwan and Berdine, Jake and Bernadett-Shapiro, Gabriel and Berner, Christopher and Bogdonoff, Lenny and Boiko, Oleg and Boyd, Madelaine and Brakman, Anna-Luisa and Brockman, Greg and Brooks, Tim and Brundage, Miles and Button, Kevin and Cai, Trevor and Campbell, Rosie and Cann, Andrew and Carey, Brittany and Carlson, Chelsea and Carmichael, Rory and Chan, Brooke and Chang, Che and Chantzis, Fotis and Chen, Derek and Chen, Sully and Chen, Ruby and Chen, Jason and Chen, Mark and Chess, Ben and Cho, Chester and Chu, Casey and Chung, Hyung Won and Cummings, Dave and Currier, Jeremiah and Dai, Yunxing and Decareaux, Cory and Degry, Thomas and Deutsch, Noah and Deville, Damien and Dhar, Arka and Dohan, David and Dowling, Steve and Dunning, Sheila and Ecoffet, Adrien and Eleti, Atty and Eloundou, Tyna and Farhi, David and Fedus, Liam and Felix, Niko and Fishman, Simón Posada and Forte, Juston and Fulford, Isabella and Gao, Leo and Georges, Elie and Gibson, Christian and Goel, Vik and Gogineni, Tarun and Goh, Gabriel and Gontijo-Lopes, Rapha and Gordon, Jonathan and Grafstein, Morgan and Gray, Scott and Greene, Ryan and Gross, Joshua and Gu, Shixiang Shane and Guo, Yufei and Hallacy, Chris and Han, Jesse and Harris, Jeff and He, Yuchen and Heaton, Mike and Heidecke, Johannes and Hesse, Chris and Hickey, Alan and Hickey, Wade and Hoeschele, Peter and Houghton, Brandon and Hsu, Kenny and Hu, Shengli and Hu, Xin and Huizinga, Joost and Jain, Shantanu and Jain, Shawn and Jang, Joanne and Jiang, Angela and Jiang, Roger and Jin, Haozhun and Jin, Denny and Jomoto, Shino and Jonn, Billie and Jun, Heewoo and Kaftan, Tomer and Kaiser, Łukasz and Kamali, Ali and Kanitscheider, Ingmar and Keskar, Nitish Shirish and Khan, Tabarak and Kilpatrick, Logan and Kim, Jong Wook and Kim, Christina and Kim, Yongjik and Kirchner, Jan Hendrik and Kiros, Jamie and Knight, Matt and Kokotajlo, Daniel and Kondraciuk, Łukasz and Kondrich, Andrew and Konstantinidis, Aris and Kosic, Kyle and Krueger, Gretchen and Kuo, Vishal and Lampe, Michael and Lan, Ikai and Lee, Teddy and Leike, Jan and Leung, Jade and Levy, Daniel and Li, Chak Ming and Lim, Rachel and Lin, Molly and Lin, Stephanie and Litwin, Mateusz and Lopez, Theresa and Lowe, Ryan and Lue, Patricia and Makanju, Anna and Malfacini, Kim and Manning, Sam and Markov, Todor and Markovski, Yaniv and Martin, Bianca and Mayer, Katie and Mayne, Andrew and McGrew, Bob and McKinney, Scott Mayer and McLeavey, Christine and McMillan, Paul and McNeil, Jake and Medina, David and Mehta, Aalok and Menick, Jacob and Metz, Luke and Mishchenko, Andrey and Mishkin, Pamela and Monaco, Vinnie and Morikawa, Evan and Mossing, Daniel and Mu, Tong and Murati, Mira and Murk, Oleg and Mély, David and Nair, Ashvin and Nakano, Reiichiro and Nayak, Rajeev and Neelakantan, Arvind and Ngo, Richard and Noh, Hyeonwoo and Ouyang, Long and O'Keefe, Cullen and Pachocki, Jakub and Paino, Alex and Palermo, Joe and Pantuliano, Ashley and Parascandolo, Giambattista and Parish, Joel and Parparita, Emy and Passos, Alex and Pavlov, Mikhail and Peng, Andrew and Perelman, Adam and Peres, Filipe de Avila Belbute and Petrov, Michael and Pinto, Henrique Ponde de Oliveira and Michael and Pokorny and Pokrass, Michelle and Pong, Vitchyr H. and Powell, Tolly and Power, Alethea and Power, Boris and Proehl, Elizabeth and Puri, Raul and Radford, Alec and Rae, Jack and Ramesh, Aditya and Raymond, Cameron and Real, Francis and Rimbach, Kendra and Ross, Carl and Rotsted, Bob and Roussez, Henri and Ryder, Nick and Saltarelli, Mario and Sanders, Ted and Santurkar, Shibani and Sastry, Girish and Schmidt, Heather and Schnurr, David and Schulman, John and Selsam, Daniel and Sheppard, Kyla and Sherbakov, Toki and Shieh, Jessica and Shoker, Sarah and Shyam, Pranav and Sidor, Szymon and Sigler, Eric and Simens, Maddie and Sitkin, Jordan and Slama, Katarina and Sohl, Ian and Sokolowsky, Benjamin and Song, Yang and Staudacher, Natalie and Such, Felipe Petroski and Summers, Natalie and Sutskever, Ilya and Tang, Jie and Tezak, Nikolas and Thompson, Madeleine B. and Tillet, Phil and Tootoonchian, Amin and Tseng, Elizabeth and Tuggle, Preston and Turley, Nick and Tworek, Jerry and Uribe, Juan Felipe Cerón and Vallone, Andrea and Vijayvergiya, Arun and Voss, Chelsea and Wainwright, Carroll and Wang, Justin Jay and Wang, Alvin and Wang, Ben and Ward, Jonathan and Wei, Jason and Weinmann, C. J. and Welihinda, Akila and Welinder, Peter and Weng, Jiayi and Weng, Lilian and Wiethoff, Matt and Willner, Dave and Winter, Clemens and Wolrich, Samuel and Wong, Hannah and Workman, Lauren and Wu, Sherwin and Wu, Jeff and Wu, Michael and Xiao, Kai and Xu, Tao and Yoo, Sarah and Yu, Kevin and Yuan, Qiming and Zaremba, Wojciech and Zellers, Rowan and Zhang, Chong and Zhang, Marvin and Zhao, Shengjia and Zheng, Tianhao and Zhuang, Juntang and Zhuk, William and Zoph, Barret},
	month = mar,
	year = {2024},
	note = {arXiv:2303.08774 [cs]},
}

@misc{deepseek-ai_deepseek-v3_2025,
	title = {{DeepSeek}-{V3} {Technical} {Report}},
	url = {http://arxiv.org/abs/2412.19437},
	doi = {10.48550/arXiv.2412.19437},
	urldate = {2025-08-31},
	publisher = {arXiv},
	author = {DeepSeek-AI and Liu, Aixin and Feng, Bei and Xue, Bing and Wang, Bingxuan and Wu, Bochao and Lu, Chengda and Zhao, Chenggang and Deng, Chengqi and Zhang, Chenyu and Ruan, Chong and Dai, Damai and Guo, Daya and Yang, Dejian and Chen, Deli and Ji, Dongjie and Li, Erhang and Lin, Fangyun and Dai, Fucong and Luo, Fuli and Hao, Guangbo and Chen, Guanting and Li, Guowei and Zhang, H. and Bao, Han and Xu, Hanwei and Wang, Haocheng and Zhang, Haowei and Ding, Honghui and Xin, Huajian and Gao, Huazuo and Li, Hui and Qu, Hui and Cai, J. L. and Liang, Jian and Guo, Jianzhong and Ni, Jiaqi and Li, Jiashi and Wang, Jiawei and Chen, Jin and Chen, Jingchang and Yuan, Jingyang and Qiu, Junjie and Li, Junlong and Song, Junxiao and Dong, Kai and Hu, Kai and Gao, Kaige and Guan, Kang and Huang, Kexin and Yu, Kuai and Wang, Lean and Zhang, Lecong and Xu, Lei and Xia, Leyi and Zhao, Liang and Wang, Litong and Zhang, Liyue and Li, Meng and Wang, Miaojun and Zhang, Mingchuan and Zhang, Minghua and Tang, Minghui and Li, Mingming and Tian, Ning and Huang, Panpan and Wang, Peiyi and Zhang, Peng and Wang, Qiancheng and Zhu, Qihao and Chen, Qinyu and Du, Qiushi and Chen, R. J. and Jin, R. L. and Ge, Ruiqi and Zhang, Ruisong and Pan, Ruizhe and Wang, Runji and Xu, Runxin and Zhang, Ruoyu and Chen, Ruyi and Li, S. S. and Lu, Shanghao and Zhou, Shangyan and Chen, Shanhuang and Wu, Shaoqing and Ye, Shengfeng and Ye, Shengfeng and Ma, Shirong and Wang, Shiyu and Zhou, Shuang and Yu, Shuiping and Zhou, Shunfeng and Pan, Shuting and Wang, T. and Yun, Tao and Pei, Tian and Sun, Tianyu and Xiao, W. L. and Zeng, Wangding and Zhao, Wanjia and An, Wei and Liu, Wen and Liang, Wenfeng and Gao, Wenjun and Yu, Wenqin and Zhang, Wentao and Li, X. Q. and Jin, Xiangyue and Wang, Xianzu and Bi, Xiao and Liu, Xiaodong and Wang, Xiaohan and Shen, Xiaojin and Chen, Xiaokang and Zhang, Xiaokang and Chen, Xiaosha and Nie, Xiaotao and Sun, Xiaowen and Wang, Xiaoxiang and Cheng, Xin and Liu, Xin and Xie, Xin and Liu, Xingchao and Yu, Xingkai and Song, Xinnan and Shan, Xinxia and Zhou, Xinyi and Yang, Xinyu and Li, Xinyuan and Su, Xuecheng and Lin, Xuheng and Li, Y. K. and Wang, Y. Q. and Wei, Y. X. and Zhu, Y. X. and Zhang, Yang and Xu, Yanhong and Xu, Yanhong and Huang, Yanping and Li, Yao and Zhao, Yao and Sun, Yaofeng and Li, Yaohui and Wang, Yaohui and Yu, Yi and Zheng, Yi and Zhang, Yichao and Shi, Yifan and Xiong, Yiliang and He, Ying and Tang, Ying and Piao, Yishi and Wang, Yisong and Tan, Yixuan and Ma, Yiyang and Liu, Yiyuan and Guo, Yongqiang and Wu, Yu and Ou, Yuan and Zhu, Yuchen and Wang, Yuduan and Gong, Yue and Zou, Yuheng and He, Yujia and Zha, Yukun and Xiong, Yunfan and Ma, Yunxian and Yan, Yuting and Luo, Yuxiang and You, Yuxiang and Liu, Yuxuan and Zhou, Yuyang and Wu, Z. F. and Ren, Z. Z. and Ren, Zehui and Sha, Zhangli and Fu, Zhe and Xu, Zhean and Huang, Zhen and Zhang, Zhen and Xie, Zhenda and Zhang, Zhengyan and Hao, Zhewen and Gou, Zhibin and Ma, Zhicheng and Yan, Zhigang and Shao, Zhihong and Xu, Zhipeng and Wu, Zhiyu and Zhang, Zhongyu and Li, Zhuoshu and Gu, Zihui and Zhu, Zijia and Liu, Zijun and Li, Zilin and Xie, Ziwei and Song, Ziyang and Gao, Ziyi and Pan, Zizheng},
	month = feb,
	year = {2025},
	note = {arXiv:2412.19437 [cs]},
}

@misc{touvron_llama_2023,
	title = {{LLaMA}: {Open} and {Efficient} {Foundation} {Language} {Models}},
	shorttitle = {{LLaMA}},
	url = {http://arxiv.org/abs/2302.13971},
	doi = {10.48550/arXiv.2302.13971},
	urldate = {2025-08-31},
	publisher = {arXiv},
	author = {Touvron, Hugo and Lavril, Thibaut and Izacard, Gautier and Martinet, Xavier and Lachaux, Marie-Anne and Lacroix, Timothée and Rozière, Baptiste and Goyal, Naman and Hambro, Eric and Azhar, Faisal and Rodriguez, Aurelien and Joulin, Armand and Grave, Edouard and Lample, Guillaume},
	month = feb,
	year = {2023},
	note = {arXiv:2302.13971 [cs]},
}

@misc{grattafiori_llama_2024,
	title = {The {Llama} 3 {Herd} of {Models}},
	url = {http://arxiv.org/abs/2407.21783},
	doi = {10.48550/arXiv.2407.21783},
	urldate = {2025-08-31},
	publisher = {arXiv},
	author = {Grattafiori, Aaron and Dubey, Abhimanyu and Jauhri, Abhinav and Pandey, Abhinav and Kadian, Abhishek and Al-Dahle, Ahmad and Letman, Aiesha and Mathur, Akhil and Schelten, Alan and Vaughan, Alex and Yang, Amy and Fan, Angela and Goyal, Anirudh and Hartshorn, Anthony and Yang, Aobo and Mitra, Archi and Sravankumar, Archie and Korenev, Artem and Hinsvark, Arthur and Rao, Arun and Zhang, Aston and Rodriguez, Aurelien and Gregerson, Austen and Spataru, Ava and Roziere, Baptiste and Biron, Bethany and Tang, Binh and Chern, Bobbie and Caucheteux, Charlotte and Nayak, Chaya and Bi, Chloe and Marra, Chris and McConnell, Chris and Keller, Christian and Touret, Christophe and Wu, Chunyang and Wong, Corinne and Ferrer, Cristian Canton and Nikolaidis, Cyrus and Allonsius, Damien and Song, Daniel and Pintz, Danielle and Livshits, Danny and Wyatt, Danny and Esiobu, David and Choudhary, Dhruv and Mahajan, Dhruv and Garcia-Olano, Diego and Perino, Diego and Hupkes, Dieuwke and Lakomkin, Egor and AlBadawy, Ehab and Lobanova, Elina and Dinan, Emily and Smith, Eric Michael and Radenovic, Filip and Guzmán, Francisco and Zhang, Frank and Synnaeve, Gabriel and Lee, Gabrielle and Anderson, Georgia Lewis and Thattai, Govind and Nail, Graeme and Mialon, Gregoire and Pang, Guan and Cucurell, Guillem and Nguyen, Hailey and Korevaar, Hannah and Xu, Hu and Touvron, Hugo and Zarov, Iliyan and Ibarra, Imanol Arrieta and Kloumann, Isabel and Misra, Ishan and Evtimov, Ivan and Zhang, Jack and Copet, Jade and Lee, Jaewon and Geffert, Jan and Vranes, Jana and Park, Jason and Mahadeokar, Jay and Shah, Jeet and Linde, Jelmer van der and Billock, Jennifer and Hong, Jenny and Lee, Jenya and Fu, Jeremy and Chi, Jianfeng and Huang, Jianyu and Liu, Jiawen and Wang, Jie and Yu, Jiecao and Bitton, Joanna and Spisak, Joe and Park, Jongsoo and Rocca, Joseph and Johnstun, Joshua and Saxe, Joshua and Jia, Junteng and Alwala, Kalyan Vasuden and Prasad, Karthik and Upasani, Kartikeya and Plawiak, Kate and Li, Ke and Heafield, Kenneth and Stone, Kevin and El-Arini, Khalid and Iyer, Krithika and Malik, Kshitiz and Chiu, Kuenley and Bhalla, Kunal and Lakhotia, Kushal and Rantala-Yeary, Lauren and Maaten, Laurens van der and Chen, Lawrence and Tan, Liang and Jenkins, Liz and Martin, Louis and Madaan, Lovish and Malo, Lubo and Blecher, Lukas and Landzaat, Lukas and Oliveira, Luke de and Muzzi, Madeline and Pasupuleti, Mahesh and Singh, Mannat and Paluri, Manohar and Kardas, Marcin and Tsimpoukelli, Maria and Oldham, Mathew and Rita, Mathieu and Pavlova, Maya and Kambadur, Melanie and Lewis, Mike and Si, Min and Singh, Mitesh Kumar and Hassan, Mona and Goyal, Naman and Torabi, Narjes and Bashlykov, Nikolay and Bogoychev, Nikolay and Chatterji, Niladri and Zhang, Ning and Duchenne, Olivier and Çelebi, Onur and Alrassy, Patrick and Zhang, Pengchuan and Li, Pengwei and Vasic, Petar and Weng, Peter and Bhargava, Prajjwal and Dubal, Pratik and Krishnan, Praveen and Koura, Punit Singh and Xu, Puxin and He, Qing and Dong, Qingxiao and Srinivasan, Ragavan and Ganapathy, Raj and Calderer, Ramon and Cabral, Ricardo Silveira and Stojnic, Robert and Raileanu, Roberta and Maheswari, Rohan and Girdhar, Rohit and Patel, Rohit and Sauvestre, Romain and Polidoro, Ronnie and Sumbaly, Roshan and Taylor, Ross and Silva, Ruan and Hou, Rui and Wang, Rui and Hosseini, Saghar and Chennabasappa, Sahana and Singh, Sanjay and Bell, Sean and Kim, Seohyun Sonia and Edunov, Sergey and Nie, Shaoliang and Narang, Sharan and Raparthy, Sharath and Shen, Sheng and Wan, Shengye and Bhosale, Shruti and Zhang, Shun and Vandenhende, Simon and Batra, Soumya and Whitman, Spencer and Sootla, Sten and Collot, Stephane and Gururangan, Suchin and Borodinsky, Sydney and Herman, Tamar and Fowler, Tara and Sheasha, Tarek and Georgiou, Thomas and Scialom, Thomas and Speckbacher, Tobias and Mihaylov, Todor and Xiao, Tong and Karn, Ujjwal and Goswami, Vedanuj and Gupta, Vibhor and Ramanathan, Vignesh and Kerkez, Viktor and Gonguet, Vincent and Do, Virginie and Vogeti, Vish and Albiero, Vítor and Petrovic, Vladan and Chu, Weiwei and Xiong, Wenhan and Fu, Wenyin and Meers, Whitney and Martinet, Xavier and Wang, Xiaodong and Wang, Xiaofang and Tan, Xiaoqing Ellen and Xia, Xide and Xie, Xinfeng and Jia, Xuchao and Wang, Xuewei and Goldschlag, Yaelle and Gaur, Yashesh and Babaei, Yasmine and Wen, Yi and Song, Yiwen and Zhang, Yuchen and Li, Yue and Mao, Yuning and Coudert, Zacharie Delpierre and Yan, Zheng and Chen, Zhengxing and Papakipos, Zoe and Singh, Aaditya and Srivastava, Aayushi and Jain, Abha and Kelsey, Adam and Shajnfeld, Adam and Gangidi, Adithya and Victoria, Adolfo and Goldstand, Ahuva and Menon, Ajay and Sharma, Ajay and Boesenberg, Alex and Baevski, Alexei and Feinstein, Allie and Kallet, Amanda and Sangani, Amit and Teo, Amos and Yunus, Anam and Lupu, Andrei and Alvarado, Andres and Caples, Andrew and Gu, Andrew and Ho, Andrew and Poulton, Andrew and Ryan, Andrew and Ramchandani, Ankit and Dong, Annie and Franco, Annie and Goyal, Anuj and Saraf, Aparajita and Chowdhury, Arkabandhu and Gabriel, Ashley and Bharambe, Ashwin and Eisenman, Assaf and Yazdan, Azadeh and James, Beau and Maurer, Ben and Leonhardi, Benjamin and Huang, Bernie and Loyd, Beth and Paola, Beto De and Paranjape, Bhargavi and Liu, Bing and Wu, Bo and Ni, Boyu and Hancock, Braden and Wasti, Bram and Spence, Brandon and Stojkovic, Brani and Gamido, Brian and Montalvo, Britt and Parker, Carl and Burton, Carly and Mejia, Catalina and Liu, Ce and Wang, Changhan and Kim, Changkyu and Zhou, Chao and Hu, Chester and Chu, Ching-Hsiang and Cai, Chris and Tindal, Chris and Feichtenhofer, Christoph and Gao, Cynthia and Civin, Damon and Beaty, Dana and Kreymer, Daniel and Li, Daniel and Adkins, David and Xu, David and Testuggine, Davide and David, Delia and Parikh, Devi and Liskovich, Diana and Foss, Didem and Wang, Dingkang and Le, Duc and Holland, Dustin and Dowling, Edward and Jamil, Eissa and Montgomery, Elaine and Presani, Eleonora and Hahn, Emily and Wood, Emily and Le, Eric-Tuan and Brinkman, Erik and Arcaute, Esteban and Dunbar, Evan and Smothers, Evan and Sun, Fei and Kreuk, Felix and Tian, Feng and Kokkinos, Filippos and Ozgenel, Firat and Caggioni, Francesco and Kanayet, Frank and Seide, Frank and Florez, Gabriela Medina and Schwarz, Gabriella and Badeer, Gada and Swee, Georgia and Halpern, Gil and Herman, Grant and Sizov, Grigory and Guangyi and Zhang and Lakshminarayanan, Guna and Inan, Hakan and Shojanazeri, Hamid and Zou, Han and Wang, Hannah and Zha, Hanwen and Habeeb, Haroun and Rudolph, Harrison and Suk, Helen and Aspegren, Henry and Goldman, Hunter and Zhan, Hongyuan and Damlaj, Ibrahim and Molybog, Igor and Tufanov, Igor and Leontiadis, Ilias and Veliche, Irina-Elena and Gat, Itai and Weissman, Jake and Geboski, James and Kohli, James and Lam, Janice and Asher, Japhet and Gaya, Jean-Baptiste and Marcus, Jeff and Tang, Jeff and Chan, Jennifer and Zhen, Jenny and Reizenstein, Jeremy and Teboul, Jeremy and Zhong, Jessica and Jin, Jian and Yang, Jingyi and Cummings, Joe and Carvill, Jon and Shepard, Jon and McPhie, Jonathan and Torres, Jonathan and Ginsburg, Josh and Wang, Junjie and Wu, Kai and U, Kam Hou and Saxena, Karan and Khandelwal, Kartikay and Zand, Katayoun and Matosich, Kathy and Veeraraghavan, Kaushik and Michelena, Kelly and Li, Keqian and Jagadeesh, Kiran and Huang, Kun and Chawla, Kunal and Huang, Kyle and Chen, Lailin and Garg, Lakshya and A, Lavender and Silva, Leandro and Bell, Lee and Zhang, Lei and Guo, Liangpeng and Yu, Licheng and Moshkovich, Liron and Wehrstedt, Luca and Khabsa, Madian and Avalani, Manav and Bhatt, Manish and Mankus, Martynas and Hasson, Matan and Lennie, Matthew and Reso, Matthias and Groshev, Maxim and Naumov, Maxim and Lathi, Maya and Keneally, Meghan and Liu, Miao and Seltzer, Michael L. and Valko, Michal and Restrepo, Michelle and Patel, Mihir and Vyatskov, Mik and Samvelyan, Mikayel and Clark, Mike and Macey, Mike and Wang, Mike and Hermoso, Miquel Jubert and Metanat, Mo and Rastegari, Mohammad and Bansal, Munish and Santhanam, Nandhini and Parks, Natascha and White, Natasha and Bawa, Navyata and Singhal, Nayan and Egebo, Nick and Usunier, Nicolas and Mehta, Nikhil and Laptev, Nikolay Pavlovich and Dong, Ning and Cheng, Norman and Chernoguz, Oleg and Hart, Olivia and Salpekar, Omkar and Kalinli, Ozlem and Kent, Parkin and Parekh, Parth and Saab, Paul and Balaji, Pavan and Rittner, Pedro and Bontrager, Philip and Roux, Pierre and Dollar, Piotr and Zvyagina, Polina and Ratanchandani, Prashant and Yuvraj, Pritish and Liang, Qian and Alao, Rachad and Rodriguez, Rachel and Ayub, Rafi and Murthy, Raghotham and Nayani, Raghu and Mitra, Rahul and Parthasarathy, Rangaprabhu and Li, Raymond and Hogan, Rebekkah and Battey, Robin and Wang, Rocky and Howes, Russ and Rinott, Ruty and Mehta, Sachin and Siby, Sachin and Bondu, Sai Jayesh and Datta, Samyak and Chugh, Sara and Hunt, Sara and Dhillon, Sargun and Sidorov, Sasha and Pan, Satadru and Mahajan, Saurabh and Verma, Saurabh and Yamamoto, Seiji and Ramaswamy, Sharadh and Lindsay, Shaun and Lindsay, Shaun and Feng, Sheng and Lin, Shenghao and Zha, Shengxin Cindy and Patil, Shishir and Shankar, Shiva and Zhang, Shuqiang and Zhang, Shuqiang and Wang, Sinong and Agarwal, Sneha and Sajuyigbe, Soji and Chintala, Soumith and Max, Stephanie and Chen, Stephen and Kehoe, Steve and Satterfield, Steve and Govindaprasad, Sudarshan and Gupta, Sumit and Deng, Summer and Cho, Sungmin and Virk, Sunny and Subramanian, Suraj and Choudhury, Sy and Goldman, Sydney and Remez, Tal and Glaser, Tamar and Best, Tamara and Koehler, Thilo and Robinson, Thomas and Li, Tianhe and Zhang, Tianjun and Matthews, Tim and Chou, Timothy and Shaked, Tzook and Vontimitta, Varun and Ajayi, Victoria and Montanez, Victoria and Mohan, Vijai and Kumar, Vinay Satish and Mangla, Vishal and Ionescu, Vlad and Poenaru, Vlad and Mihailescu, Vlad Tiberiu and Ivanov, Vladimir and Li, Wei and Wang, Wenchen and Jiang, Wenwen and Bouaziz, Wes and Constable, Will and Tang, Xiaocheng and Wu, Xiaojian and Wang, Xiaolan and Wu, Xilun and Gao, Xinbo and Kleinman, Yaniv and Chen, Yanjun and Hu, Ye and Jia, Ye and Qi, Ye and Li, Yenda and Zhang, Yilin and Zhang, Ying and Adi, Yossi and Nam, Youngjin and Yu and Wang and Zhao, Yu and Hao, Yuchen and Qian, Yundi and Li, Yunlu and He, Yuzi and Rait, Zach and DeVito, Zachary and Rosnbrick, Zef and Wen, Zhaoduo and Yang, Zhenyu and Zhao, Zhiwei and Ma, Zhiyu},
	month = nov,
	year = {2024},
	note = {arXiv:2407.21783 [cs]},
}

@article{hara_active_2024,
	title = {Active model selection: {A} variance minimization approach},
	volume = {113},
	issn = {1573-0565},
	shorttitle = {Active model selection},
	url = {https://doi.org/10.1007/s10994-024-06603-1},
	doi = {10.1007/s10994-024-06603-1},
	language = {en},
	number = {11},
	urldate = {2025-09-01},
	journal = {Machine Learning},
	author = {Hara, Satoshi and Matsuura, Mitsuru and Honda, Junya and Ito, Shinji},
	month = dec,
	year = {2024},
	pages = {8327--8345},
}

@misc{fan_bibliometric_2023,
	title = {A {Bibliometric} {Review} of {Large} {Language} {Models} {Research} from 2017 to 2023},
	url = {http://arxiv.org/abs/2304.02020},
	doi = {10.48550/arXiv.2304.02020},
	urldate = {2025-09-05},
	publisher = {arXiv},
	author = {Fan, Lizhou and Li, Lingyao and Ma, Zihui and Lee, Sanggyu and Yu, Huizi and Hemphill, Libby},
	month = apr,
	year = {2023},
	note = {arXiv:2304.02020 [cs]},
}

@misc{tan_unstructured_2023,
	title = {Unstructured and structured data: {Can} we have the best of both worlds with large language models?},
	shorttitle = {Unstructured and structured data},
	url = {http://arxiv.org/abs/2304.13010},
	doi = {10.48550/arXiv.2304.13010},
	urldate = {2025-09-05},
	publisher = {arXiv},
	author = {Tan, Wang-Chiew},
	month = jul,
	year = {2023},
	note = {arXiv:2304.13010 [cs]},
}

@misc{guo_evaluating_2023,
	title = {Evaluating {Large} {Language} {Models}: {A} {Comprehensive} {Survey}},
	shorttitle = {Evaluating {Large} {Language} {Models}},
	url = {http://arxiv.org/abs/2310.19736},
	doi = {10.48550/arXiv.2310.19736},
	urldate = {2025-09-06},
	publisher = {arXiv},
	author = {Guo, Zishan and Jin, Renren and Liu, Chuang and Huang, Yufei and Shi, Dan and Supryadi and Yu, Linhao and Liu, Yan and Li, Jiaxuan and Xiong, Bojian and Xiong, Deyi},
	month = nov,
	year = {2023},
	note = {arXiv:2310.19736 [cs]},
}

@misc{loem_exploring_2023,
	title = {Exploring {Effectiveness} of {GPT}-3 in {Grammatical} {Error} {Correction}: {A} {Study} on {Performance} and {Controllability} in {Prompt}-{Based} {Methods}},
	shorttitle = {Exploring {Effectiveness} of {GPT}-3 in {Grammatical} {Error} {Correction}},
	url = {http://arxiv.org/abs/2305.18156},
	doi = {10.48550/arXiv.2305.18156},
	urldate = {2025-09-06},
	publisher = {arXiv},
	author = {Loem, Mengsay and Kaneko, Masahiro and Takase, Sho and Okazaki, Naoaki},
	month = may,
	year = {2023},
	note = {arXiv:2305.18156 [cs]},
}

@misc{gao_pal_2023,
	title = {{PAL}: {Program}-aided {Language} {Models}},
	shorttitle = {{PAL}},
	url = {http://arxiv.org/abs/2211.10435},
	doi = {10.48550/arXiv.2211.10435},
	urldate = {2025-09-06},
	publisher = {arXiv},
	author = {Gao, Luyu and Madaan, Aman and Zhou, Shuyan and Alon, Uri and Liu, Pengfei and Yang, Yiming and Callan, Jamie and Neubig, Graham},
	month = jan,
	year = {2023},
	note = {arXiv:2211.10435 [cs]},
}

@misc{gambardella_language_2024,
	title = {Language {Models} {Do} {Hard} {Arithmetic} {Tasks} {Easily} and {Hardly} {Do} {Easy} {Arithmetic} {Tasks}},
	url = {http://arxiv.org/abs/2406.02356},
	doi = {10.48550/arXiv.2406.02356},
	urldate = {2025-09-06},
	publisher = {arXiv},
	author = {Gambardella, Andrew and Iwasawa, Yusuke and Matsuo, Yutaka},
	month = jun,
	year = {2024},
	note = {arXiv:2406.02356 [cs]},
}

@misc{madani_active_2012,
	title = {Active {Model} {Selection}},
	url = {http://arxiv.org/abs/1207.4138},
	doi = {10.48550/arXiv.1207.4138},
	urldate = {2025-09-06},
	publisher = {arXiv},
	author = {Madani, Omid and Lizotte, Daniel J. and Greiner, Russell},
	month = jul,
	year = {2012},
	note = {arXiv:1207.4138 [cs]},
}

@misc{kay_consensus-driven_2025,
	title = {Consensus-{Driven} {Active} {Model} {Selection}},
	url = {http://arxiv.org/abs/2507.23771},
	doi = {10.48550/arXiv.2507.23771},
	urldate = {2025-09-06},
	publisher = {arXiv},
	author = {Kay, Justin and Horn, Grant Van and Maji, Subhransu and Sheldon, Daniel and Beery, Sara},
	month = jul,
	year = {2025},
	note = {arXiv:2507.23771 [cs]},
}

@misc{okanovic_all_2024,
	title = {All models are wrong, some are useful: {Model} {Selection} with {Limited} {Labels}},
	shorttitle = {All models are wrong, some are useful},
	url = {http://arxiv.org/abs/2410.13609},
	doi = {10.48550/arXiv.2410.13609},
	urldate = {2025-09-06},
	publisher = {arXiv},
	author = {Okanovic, Patrik and Kirsch, Andreas and Kasper, Jannes and Hoefler, Torsten and Krause, Andreas and Gürel, Nezihe Merve},
	month = oct,
	year = {2024},
	note = {arXiv:2410.13609 [cs]},
}

@inproceedings{papineni_bleu_2002,
	address = {Philadelphia, Pennsylvania, USA},
	title = {Bleu: a {Method} for {Automatic} {Evaluation} of {Machine} {Translation}},
	shorttitle = {Bleu},
	url = {https://aclanthology.org/P02-1040/},
	doi = {10.3115/1073083.1073135},
	urldate = {2025-09-06},
	booktitle = {Proceedings of the 40th {Annual} {Meeting} of the {Association} for {Computational} {Linguistics}},
	publisher = {Association for Computational Linguistics},
	author = {Papineni, Kishore and Roukos, Salim and Ward, Todd and Zhu, Wei-Jing},
	editor = {Isabelle, Pierre and Charniak, Eugene and Lin, Dekang},
	month = jul,
	year = {2002},
	pages = {311--318},
}

@inproceedings{lavie_meteor_2007,
	address = {USA},
	series = {{StatMT} '07},
	title = {Meteor: an automatic metric for {MT} evaluation with high levels of correlation with human judgments},
	shorttitle = {Meteor},
	urldate = {2025-09-06},
	booktitle = {Proceedings of the {Second} {Workshop} on {Statistical} {Machine} {Translation}},
	publisher = {Association for Computational Linguistics},
	author = {Lavie, Alon and Agarwal, Abhaya},
	year = {2007},
	pages = {228--231},
}

@misc{wang_superglue_2020,
	title = {{SuperGLUE}: {A} {Stickier} {Benchmark} for {General}-{Purpose} {Language} {Understanding} {Systems}},
	shorttitle = {{SuperGLUE}},
	url = {http://arxiv.org/abs/1905.00537},
	doi = {10.48550/arXiv.1905.00537},
	urldate = {2025-09-06},
	publisher = {arXiv},
	author = {Wang, Alex and Pruksachatkun, Yada and Nangia, Nikita and Singh, Amanpreet and Michael, Julian and Hill, Felix and Levy, Omer and Bowman, Samuel R.},
	month = feb,
	year = {2020},
	note = {arXiv:1905.00537 [cs]},
}

@misc{hendrycks_measuring_2021,
	title = {Measuring {Massive} {Multitask} {Language} {Understanding}},
	url = {http://arxiv.org/abs/2009.03300},
	doi = {10.48550/arXiv.2009.03300},
	urldate = {2025-09-06},
	publisher = {arXiv},
	author = {Hendrycks, Dan and Burns, Collin and Basart, Steven and Zou, Andy and Mazeika, Mantas and Song, Dawn and Steinhardt, Jacob},
	month = jan,
	year = {2021},
	note = {arXiv:2009.03300 [cs]},
}

@misc{srivastava_beyond_2023,
	title = {Beyond the {Imitation} {Game}: {Quantifying} and extrapolating the capabilities of language models},
	shorttitle = {Beyond the {Imitation} {Game}},
	url = {http://arxiv.org/abs/2206.04615},
	doi = {10.48550/arXiv.2206.04615},
	urldate = {2025-09-06},
	publisher = {arXiv},
	author = {Srivastava, Aarohi and Rastogi, Abhinav and Rao, Abhishek and Shoeb, Abu Awal Md and Abid, Abubakar and Fisch, Adam and Brown, Adam R. and Santoro, Adam and Gupta, Aditya and Garriga-Alonso, Adrià and Kluska, Agnieszka and Lewkowycz, Aitor and Agarwal, Akshat and Power, Alethea and Ray, Alex and Warstadt, Alex and Kocurek, Alexander W. and Safaya, Ali and Tazarv, Ali and Xiang, Alice and Parrish, Alicia and Nie, Allen and Hussain, Aman and Askell, Amanda and Dsouza, Amanda and Slone, Ambrose and Rahane, Ameet and Iyer, Anantharaman S. and Andreassen, Anders and Madotto, Andrea and Santilli, Andrea and Stuhlmüller, Andreas and Dai, Andrew and La, Andrew and Lampinen, Andrew and Zou, Andy and Jiang, Angela and Chen, Angelica and Vuong, Anh and Gupta, Animesh and Gottardi, Anna and Norelli, Antonio and Venkatesh, Anu and Gholamidavoodi, Arash and Tabassum, Arfa and Menezes, Arul and Kirubarajan, Arun and Mullokandov, Asher and Sabharwal, Ashish and Herrick, Austin and Efrat, Avia and Erdem, Aykut and Karakaş, Ayla and Roberts, B. Ryan and Loe, Bao Sheng and Zoph, Barret and Bojanowski, Bartłomiej and Özyurt, Batuhan and Hedayatnia, Behnam and Neyshabur, Behnam and Inden, Benjamin and Stein, Benno and Ekmekci, Berk and Lin, Bill Yuchen and Howald, Blake and Orinion, Bryan and Diao, Cameron and Dour, Cameron and Stinson, Catherine and Argueta, Cedrick and Ramírez, César Ferri and Singh, Chandan and Rathkopf, Charles and Meng, Chenlin and Baral, Chitta and Wu, Chiyu and Callison-Burch, Chris and Waites, Chris and Voigt, Christian and Manning, Christopher D. and Potts, Christopher and Ramirez, Cindy and Rivera, Clara E. and Siro, Clemencia and Raffel, Colin and Ashcraft, Courtney and Garbacea, Cristina and Sileo, Damien and Garrette, Dan and Hendrycks, Dan and Kilman, Dan and Roth, Dan and Freeman, Daniel and Khashabi, Daniel and Levy, Daniel and González, Daniel Moseguí and Perszyk, Danielle and Hernandez, Danny and Chen, Danqi and Ippolito, Daphne and Gilboa, Dar and Dohan, David and Drakard, David and Jurgens, David and Datta, Debajyoti and Ganguli, Deep and Emelin, Denis and Kleyko, Denis and Yuret, Deniz and Chen, Derek and Tam, Derek and Hupkes, Dieuwke and Misra, Diganta and Buzan, Dilyar and Mollo, Dimitri Coelho and Yang, Diyi and Lee, Dong-Ho and Schrader, Dylan and Shutova, Ekaterina and Cubuk, Ekin Dogus and Segal, Elad and Hagerman, Eleanor and Barnes, Elizabeth and Donoway, Elizabeth and Pavlick, Ellie and Rodola, Emanuele and Lam, Emma and Chu, Eric and Tang, Eric and Erdem, Erkut and Chang, Ernie and Chi, Ethan A. and Dyer, Ethan and Jerzak, Ethan and Kim, Ethan and Manyasi, Eunice Engefu and Zheltonozhskii, Evgenii and Xia, Fanyue and Siar, Fatemeh and Martínez-Plumed, Fernando and Happé, Francesca and Chollet, Francois and Rong, Frieda and Mishra, Gaurav and Winata, Genta Indra and Melo, Gerard de and Kruszewski, Germán and Parascandolo, Giambattista and Mariani, Giorgio and Wang, Gloria and Jaimovitch-López, Gonzalo and Betz, Gregor and Gur-Ari, Guy and Galijasevic, Hana and Kim, Hannah and Rashkin, Hannah and Hajishirzi, Hannaneh and Mehta, Harsh and Bogar, Hayden and Shevlin, Henry and Schütze, Hinrich and Yakura, Hiromu and Zhang, Hongming and Wong, Hugh Mee and Ng, Ian and Noble, Isaac and Jumelet, Jaap and Geissinger, Jack and Kernion, Jackson and Hilton, Jacob and Lee, Jaehoon and Fisac, Jaime Fernández and Simon, James B. and Koppel, James and Zheng, James and Zou, James and Kocoń, Jan and Thompson, Jana and Wingfield, Janelle and Kaplan, Jared and Radom, Jarema and Sohl-Dickstein, Jascha and Phang, Jason and Wei, Jason and Yosinski, Jason and Novikova, Jekaterina and Bosscher, Jelle and Marsh, Jennifer and Kim, Jeremy and Taal, Jeroen and Engel, Jesse and Alabi, Jesujoba and Xu, Jiacheng and Song, Jiaming and Tang, Jillian and Waweru, Joan and Burden, John and Miller, John and Balis, John U. and Batchelder, Jonathan and Berant, Jonathan and Frohberg, Jörg and Rozen, Jos and Hernandez-Orallo, Jose and Boudeman, Joseph and Guerr, Joseph and Jones, Joseph and Tenenbaum, Joshua B. and Rule, Joshua S. and Chua, Joyce and Kanclerz, Kamil and Livescu, Karen and Krauth, Karl and Gopalakrishnan, Karthik and Ignatyeva, Katerina and Markert, Katja and Dhole, Kaustubh D. and Gimpel, Kevin and Omondi, Kevin and Mathewson, Kory and Chiafullo, Kristen and Shkaruta, Ksenia and Shridhar, Kumar and McDonell, Kyle and Richardson, Kyle and Reynolds, Laria and Gao, Leo and Zhang, Li and Dugan, Liam and Qin, Lianhui and Contreras-Ochando, Lidia and Morency, Louis-Philippe and Moschella, Luca and Lam, Lucas and Noble, Lucy and Schmidt, Ludwig and He, Luheng and Colón, Luis Oliveros and Metz, Luke and Şenel, Lütfi Kerem and Bosma, Maarten and Sap, Maarten and Hoeve, Maartje ter and Farooqi, Maheen and Faruqui, Manaal and Mazeika, Mantas and Baturan, Marco and Marelli, Marco and Maru, Marco and Quintana, Maria Jose Ramírez and Tolkiehn, Marie and Giulianelli, Mario and Lewis, Martha and Potthast, Martin and Leavitt, Matthew L. and Hagen, Matthias and Schubert, Mátyás and Baitemirova, Medina Orduna and Arnaud, Melody and McElrath, Melvin and Yee, Michael A. and Cohen, Michael and Gu, Michael and Ivanitskiy, Michael and Starritt, Michael and Strube, Michael and Swędrowski, Michał and Bevilacqua, Michele and Yasunaga, Michihiro and Kale, Mihir and Cain, Mike and Xu, Mimee and Suzgun, Mirac and Walker, Mitch and Tiwari, Mo and Bansal, Mohit and Aminnaseri, Moin and Geva, Mor and Gheini, Mozhdeh and T, Mukund Varma and Peng, Nanyun and Chi, Nathan A. and Lee, Nayeon and Krakover, Neta Gur-Ari and Cameron, Nicholas and Roberts, Nicholas and Doiron, Nick and Martinez, Nicole and Nangia, Nikita and Deckers, Niklas and Muennighoff, Niklas and Keskar, Nitish Shirish and Iyer, Niveditha S. and Constant, Noah and Fiedel, Noah and Wen, Nuan and Zhang, Oliver and Agha, Omar and Elbaghdadi, Omar and Levy, Omer and Evans, Owain and Casares, Pablo Antonio Moreno and Doshi, Parth and Fung, Pascale and Liang, Paul Pu and Vicol, Paul and Alipoormolabashi, Pegah and Liao, Peiyuan and Liang, Percy and Chang, Peter and Eckersley, Peter and Htut, Phu Mon and Hwang, Pinyu and Miłkowski, Piotr and Patil, Piyush and Pezeshkpour, Pouya and Oli, Priti and Mei, Qiaozhu and Lyu, Qing and Chen, Qinlang and Banjade, Rabin and Rudolph, Rachel Etta and Gabriel, Raefer and Habacker, Rahel and Risco, Ramon and Millière, Raphaël and Garg, Rhythm and Barnes, Richard and Saurous, Rif A. and Arakawa, Riku and Raymaekers, Robbe and Frank, Robert and Sikand, Rohan and Novak, Roman and Sitelew, Roman and LeBras, Ronan and Liu, Rosanne and Jacobs, Rowan and Zhang, Rui and Salakhutdinov, Ruslan and Chi, Ryan and Lee, Ryan and Stovall, Ryan and Teehan, Ryan and Yang, Rylan and Singh, Sahib and Mohammad, Saif M. and Anand, Sajant and Dillavou, Sam and Shleifer, Sam and Wiseman, Sam and Gruetter, Samuel and Bowman, Samuel R. and Schoenholz, Samuel S. and Han, Sanghyun and Kwatra, Sanjeev and Rous, Sarah A. and Ghazarian, Sarik and Ghosh, Sayan and Casey, Sean and Bischoff, Sebastian and Gehrmann, Sebastian and Schuster, Sebastian and Sadeghi, Sepideh and Hamdan, Shadi and Zhou, Sharon and Srivastava, Shashank and Shi, Sherry and Singh, Shikhar and Asaadi, Shima and Gu, Shixiang Shane and Pachchigar, Shubh and Toshniwal, Shubham and Upadhyay, Shyam and Shyamolima and Debnath and Shakeri, Siamak and Thormeyer, Simon and Melzi, Simone and Reddy, Siva and Makini, Sneha Priscilla and Lee, Soo-Hwan and Torene, Spencer and Hatwar, Sriharsha and Dehaene, Stanislas and Divic, Stefan and Ermon, Stefano and Biderman, Stella and Lin, Stephanie and Prasad, Stephen and Piantadosi, Steven T. and Shieber, Stuart M. and Misherghi, Summer and Kiritchenko, Svetlana and Mishra, Swaroop and Linzen, Tal and Schuster, Tal and Li, Tao and Yu, Tao and Ali, Tariq and Hashimoto, Tatsu and Wu, Te-Lin and Desbordes, Théo and Rothschild, Theodore and Phan, Thomas and Wang, Tianle and Nkinyili, Tiberius and Schick, Timo and Kornev, Timofei and Tunduny, Titus and Gerstenberg, Tobias and Chang, Trenton and Neeraj, Trishala and Khot, Tushar and Shultz, Tyler and Shaham, Uri and Misra, Vedant and Demberg, Vera and Nyamai, Victoria and Raunak, Vikas and Ramasesh, Vinay and Prabhu, Vinay Uday and Padmakumar, Vishakh and Srikumar, Vivek and Fedus, William and Saunders, William and Zhang, William and Vossen, Wout and Ren, Xiang and Tong, Xiaoyu and Zhao, Xinran and Wu, Xinyi and Shen, Xudong and Yaghoobzadeh, Yadollah and Lakretz, Yair and Song, Yangqiu and Bahri, Yasaman and Choi, Yejin and Yang, Yichi and Hao, Yiding and Chen, Yifu and Belinkov, Yonatan and Hou, Yu and Hou, Yufang and Bai, Yuntao and Seid, Zachary and Zhao, Zhuoye and Wang, Zijian and Wang, Zijie J. and Wang, Zirui and Wu, Ziyi},
	month = jun,
	year = {2023},
	note = {arXiv:2206.04615 [cs]},
}

@misc{stiennon_learning_2022,
	title = {Learning to summarize from human feedback},
	url = {http://arxiv.org/abs/2009.01325},
	doi = {10.48550/arXiv.2009.01325},
	urldate = {2025-09-06},
	publisher = {arXiv},
	author = {Stiennon, Nisan and Ouyang, Long and Wu, Jeff and Ziegler, Daniel M. and Lowe, Ryan and Voss, Chelsea and Radford, Alec and Amodei, Dario and Christiano, Paul},
	month = feb,
	year = {2022},
	note = {arXiv:2009.01325 [cs]},
}

@inproceedings{bai_mt-bench-101_2024,
	title = {{MT}-{Bench}-101: {A} {Fine}-{Grained} {Benchmark} for {Evaluating} {Large} {Language} {Models} in {Multi}-{Turn} {Dialogues}},
	shorttitle = {{MT}-{Bench}-101},
	url = {http://arxiv.org/abs/2402.14762},
	doi = {10.18653/v1/2024.acl-long.401},
	urldate = {2025-09-06},
	booktitle = {Proceedings of the 62nd {Annual} {Meeting} of the {Association} for {Computational} {Linguistics} ({Volume} 1: {Long} {Papers})},
	author = {Bai, Ge and Liu, Jie and Bu, Xingyuan and He, Yancheng and Liu, Jiaheng and Zhou, Zhanhui and Lin, Zhuoran and Su, Wenbo and Ge, Tiezheng and Zheng, Bo and Ouyang, Wanli},
	year = {2024},
	note = {arXiv:2402.14762 [cs]},
	pages = {7421--7454},
}

@misc{chiang_chatbot_2024,
	title = {Chatbot {Arena}: {An} {Open} {Platform} for {Evaluating} {LLMs} by {Human} {Preference}},
	shorttitle = {Chatbot {Arena}},
	url = {http://arxiv.org/abs/2403.04132},
	doi = {10.48550/arXiv.2403.04132},
	urldate = {2025-09-06},
	publisher = {arXiv},
	author = {Chiang, Wei-Lin and Zheng, Lianmin and Sheng, Ying and Angelopoulos, Anastasios Nikolas and Li, Tianle and Li, Dacheng and Zhang, Hao and Zhu, Banghua and Jordan, Michael and Gonzalez, Joseph E. and Stoica, Ion},
	month = mar,
	year = {2024},
	note = {arXiv:2403.04132 [cs]},
}

@misc{dubois_length-controlled_2025,
	title = {Length-{Controlled} {AlpacaEval}: {A} {Simple} {Way} to {Debias} {Automatic} {Evaluators}},
	shorttitle = {Length-{Controlled} {AlpacaEval}},
	url = {http://arxiv.org/abs/2404.04475},
	doi = {10.48550/arXiv.2404.04475},
	urldate = {2025-09-06},
	publisher = {arXiv},
	author = {Dubois, Yann and Galambosi, Balázs and Liang, Percy and Hashimoto, Tatsunori B.},
	month = mar,
	year = {2025},
	note = {arXiv:2404.04475 [cs]},
}

@misc{liang_holistic_2023,
	title = {Holistic {Evaluation} of {Language} {Models}},
	url = {http://arxiv.org/abs/2211.09110},
	doi = {10.48550/arXiv.2211.09110},
	urldate = {2025-09-06},
	publisher = {arXiv},
	author = {Liang, Percy and Bommasani, Rishi and Lee, Tony and Tsipras, Dimitris and Soylu, Dilara and Yasunaga, Michihiro and Zhang, Yian and Narayanan, Deepak and Wu, Yuhuai and Kumar, Ananya and Newman, Benjamin and Yuan, Binhang and Yan, Bobby and Zhang, Ce and Cosgrove, Christian and Manning, Christopher D. and Ré, Christopher and Acosta-Navas, Diana and Hudson, Drew A. and Zelikman, Eric and Durmus, Esin and Ladhak, Faisal and Rong, Frieda and Ren, Hongyu and Yao, Huaxiu and Wang, Jue and Santhanam, Keshav and Orr, Laurel and Zheng, Lucia and Yuksekgonul, Mert and Suzgun, Mirac and Kim, Nathan and Guha, Neel and Chatterji, Niladri and Khattab, Omar and Henderson, Peter and Huang, Qian and Chi, Ryan and Xie, Sang Michael and Santurkar, Shibani and Ganguli, Surya and Hashimoto, Tatsunori and Icard, Thomas and Zhang, Tianyi and Chaudhary, Vishrav and Wang, William and Li, Xuechen and Mai, Yifan and Zhang, Yuhui and Koreeda, Yuta},
	month = oct,
	year = {2023},
	note = {arXiv:2211.09110 [cs]},
}

@misc{berrada_scaling_2025,
	title = {Scaling {Up} {Active} {Testing} to {Large} {Language} {Models}},
	url = {http://arxiv.org/abs/2508.09093},
	doi = {10.48550/arXiv.2508.09093},
	urldate = {2025-09-06},
	publisher = {arXiv},
	author = {Berrada, Gabrielle and Kossen, Jannik and Razzak, Muhammed and Smith, Freddie Bickford and Gal, Yarin and Rainforth, Tom},
	month = aug,
	year = {2025},
	note = {arXiv:2508.09093 [cs]},
}

@misc{wang_glue_2019,
	title = {{GLUE}: {A} {Multi}-{Task} {Benchmark} and {Analysis} {Platform} for {Natural} {Language} {Understanding}},
	shorttitle = {{GLUE}},
	url = {http://arxiv.org/abs/1804.07461},
	doi = {10.48550/arXiv.1804.07461},
	urldate = {2025-09-06},
	publisher = {arXiv},
	author = {Wang, Alex and Singh, Amanpreet and Michael, Julian and Hill, Felix and Levy, Omer and Bowman, Samuel R.},
	month = feb,
	year = {2019},
	note = {arXiv:1804.07461 [cs]},
}

@misc{gu_survey_2025,
	title = {A {Survey} on {LLM}-as-a-{Judge}},
	url = {http://arxiv.org/abs/2411.15594},
	doi = {10.48550/arXiv.2411.15594},
	urldate = {2025-09-06},
	publisher = {arXiv},
	author = {Gu, Jiawei and Jiang, Xuhui and Shi, Zhichao and Tan, Hexiang and Zhai, Xuehao and Xu, Chengjin and Li, Wei and Shen, Yinghan and Ma, Shengjie and Liu, Honghao and Wang, Saizhuo and Zhang, Kun and Wang, Yuanzhuo and Gao, Wen and Ni, Lionel and Guo, Jian},
	month = mar,
	year = {2025},
	note = {arXiv:2411.15594 [cs]},
}

@misc{ashury-tahan_label-efficient_2024,
	title = {Label-{Efficient} {Model} {Selection} for {Text} {Generation}},
	url = {http://arxiv.org/abs/2402.07891},
	doi = {10.48550/arXiv.2402.07891},
	urldate = {2025-09-13},
	publisher = {arXiv},
	author = {Ashury-Tahan, Shir and Gera, Ariel and Sznajder, Benjamin and Choshen, Leshem and Ein-Dor, Liat and Shnarch, Eyal},
	month = jun,
	year = {2024},
	note = {arXiv:2402.07891 [cs]},
}

@misc{shlens_notes_2014,
	title = {Notes on {Kullback}-{Leibler} {Divergence} and {Likelihood}},
	url = {http://arxiv.org/abs/1404.2000},
	doi = {10.48550/arXiv.1404.2000},
	urldate = {2025-09-13},
	publisher = {arXiv},
	author = {Shlens, Jonathon},
	month = apr,
	year = {2014},
	note = {arXiv:1404.2000 [cs]},
}

@inproceedings{kossen_active_2021,
	title = {Active {Testing}: {Sample}-{Efficient} {Model} {Evaluation}},
	shorttitle = {Active {Testing}},
	url = {https://proceedings.mlr.press/v139/kossen21a.html},
	language = {en},
	urldate = {2025-09-16},
	booktitle = {Proceedings of the 38th {International} {Conference} on {Machine} {Learning}},
	publisher = {PMLR},
	author = {Kossen, Jannik and Farquhar, Sebastian and Gal, Yarin and Rainforth, Tom},
	month = jul,
	year = {2021},
	note = {ISSN: 2640-3498},
	pages = {5753--5763},
}

@inproceedings{polo_tinybenchmarks_2024,
	address = {Vienna, Austria},
	series = {{ICML}'24},
	title = {{tinyBenchmarks}: evaluating {LLMs} with fewer examples},
	volume = {235},
	shorttitle = {{tinyBenchmarks}},
	urldate = {2025-09-16},
	booktitle = {Proceedings of the 41st {International} {Conference} on {Machine} {Learning}},
	publisher = {JMLR.org},
	author = {Polo, Felipe Maia and Weber, Lucas and Choshen, Leshem and Sun, Yuekai and Xu, Gongjun and Yurochkin, Mikhail},
	year = {2024},
	pages = {34303--34326},
}

@inproceedings{vivek_anchor_2024,
	address = {St. Julian's, Malta},
	title = {Anchor {Points}: {Benchmarking} {Models} with {Much} {Fewer} {Examples}},
	shorttitle = {Anchor {Points}},
	url = {https://aclanthology.org/2024.eacl-long.95/},
	doi = {10.18653/v1/2024.eacl-long.95},
	urldate = {2025-09-16},
	booktitle = {Proceedings of the 18th {Conference} of the {European} {Chapter} of the {Association} for {Computational} {Linguistics} ({Volume} 1: {Long} {Papers})},
	publisher = {Association for Computational Linguistics},
	author = {Vivek, Rajan and Ethayarajh, Kawin and Yang, Diyi and Kiela, Douwe},
	editor = {Graham, Yvette and Purver, Matthew},
	month = mar,
	year = {2024},
	pages = {1576--1601},
}

@misc{google-t5_google-t5t5-small_2024,
	title = {google-t5/t5-small · {Hugging} {Face}},
	url = {https://huggingface.co/google-t5/t5-small},
	urldate = {2025-09-17},
	author = {Google-t5},
	month = mar,
	year = {2024},
}

@misc{meta-llama_meta-llamallama-31-8b_2024,
	title = {meta-llama/{Llama}-3.1-{8B} · {Hugging} {Face}},
	url = {https://huggingface.co/meta-llama/Llama-3.1-8B},
	urldate = {2025-09-17},
	author = {Meta-Llama},
	month = dec,
	year = {2024},
}

@misc{deepseek_deepseek-aideepseek-v31_2025,
	title = {deepseek-ai/{DeepSeek}-{V3}.1 · {Hugging} {Face}},
	url = {https://huggingface.co/deepseek-ai/DeepSeek-V3.1},
	urldate = {2025-09-17},
	author = {DeepSeek},
	month = aug,
	year = {2025},
}

@misc{falcon_tiiuaefalcon-7b-instruct_2023,
	title = {tiiuae/falcon-7b-instruct · {Hugging} {Face}},
	url = {https://huggingface.co/tiiuae/falcon-7b-instruct},
	urldate = {2025-09-17},
	author = {Falcon},
	month = jun,
	year = {2023},
}

@misc{lee_garrethleecomprehensive-arithmetic-problems_2024,
	title = {garrethlee/comprehensive-arithmetic-problems · {Datasets} at {Hugging} {Face}},
	url = {https://huggingface.co/datasets/garrethlee/comprehensive-arithmetic-problems},
	urldate = {2025-09-17},
	author = {Lee, Garreth},
	month = dec,
	year = {2024},
}

@misc{agentlans_agentlansgrammar-correction_2024,
	title = {agentlans/grammar-correction · {Datasets} at {Hugging} {Face}},
	url = {https://huggingface.co/datasets/agentlans/grammar-correction},
	urldate = {2025-09-17},
	author = {Agentlans},
	month = dec,
	year = {2024},
}

@misc{zhang_di-zhang-fducalculus-dataset_2025,
	title = {di-zhang-fdu/calculus-dataset · {Datasets} at {Hugging} {Face}},
	url = {https://huggingface.co/datasets/di-zhang-fdu/calculus-dataset},
	urldate = {2025-09-17},
	author = {Zhang, Di},
	month = jan,
	year = {2025},
}

@misc{khandekar_medcalc-bench_2024,
	title = {{MedCalc}-{Bench}: {Evaluating} {Large} {Language} {Models} for {Medical} {Calculations}},
	shorttitle = {{MedCalc}-{Bench}},
	url = {http://arxiv.org/abs/2406.12036},
	doi = {10.48550/arXiv.2406.12036},
	urldate = {2025-09-17},
	publisher = {arXiv},
	author = {Khandekar, Nikhil and Jin, Qiao and Xiong, Guangzhi and Dunn, Soren and Applebaum, Serina S. and Anwar, Zain and Sarfo-Gyamfi, Maame and Safranek, Conrad W. and Anwar, Abid A. and Zhang, Andrew and Gilson, Aidan and Singer, Maxwell B. and Dave, Amisha and Taylor, Andrew and Zhang, Aidong and Chen, Qingyu and Lu, Zhiyong},
	month = jun,
	year = {2024},
	note = {arXiv:2406.12036 [cs]},
}

@misc{bedi_medhelm_2025,
	title = {{MedHELM}: {Holistic} {Evaluation} of {Large} {Language} {Models} for {Medical} {Tasks}},
	shorttitle = {{MedHELM}},
	url = {http://arxiv.org/abs/2505.23802},
	doi = {10.48550/arXiv.2505.23802},
	urldate = {2025-09-17},
	publisher = {arXiv},
	author = {Bedi, Suhana and Cui, Hejie and Fuentes, Miguel and Unell, Alyssa and Wornow, Michael and Banda, Juan M. and Kotecha, Nikesh and Keyes, Timothy and Mai, Yifan and Oez, Mert and Qiu, Hao and Jain, Shrey and Schettini, Leonardo and Kashyap, Mehr and Fries, Jason Alan and Swaminathan, Akshay and Chung, Philip and Nateghi, Fateme and Aali, Asad and Nayak, Ashwin and Vedak, Shivam and Jain, Sneha S. and Patel, Birju and Fayanju, Oluseyi and Shah, Shreya and Goh, Ethan and Yao, Dong-han and Soetikno, Brian and Reis, Eduardo and Gatidis, Sergios and Divi, Vasu and Capasso, Robson and Saralkar, Rachna and Chiang, Chia-Chun and Jindal, Jenelle and Pham, Tho and Ghoddusi, Faraz and Lin, Steven and Chiou, Albert S. and Hong, Christy and Roy, Mohana and Gensheimer, Michael F. and Patel, Hinesh and Schulman, Kevin and Dash, Dev and Char, Danton and Downing, Lance and Grolleau, Francois and Black, Kameron and Mieso, Bethel and Zahedivash, Aydin and Yim, Wen-wai and Sharma, Harshita and Lee, Tony and Kirsch, Hannah and Lee, Jennifer and Ambers, Nerissa and Lugtu, Carlene and Sharma, Aditya and Mawji, Bilal and Alekseyev, Alex and Zhou, Vicky and Kakkar, Vikas and Helzer, Jarrod and Revri, Anurang and Bannett, Yair and Daneshjou, Roxana and Chen, Jonathan and Alsentzer, Emily and Morse, Keith and Ravi, Nirmal and Aghaeepour, Nima and Kennedy, Vanessa and Chaudhari, Akshay and Wang, Thomas and Koyejo, Sanmi and Lungren, Matthew P. and Horvitz, Eric and Liang, Percy and Pfeffer, Mike and Shah, Nigam H.},
	month = jun,
	year = {2025},
	note = {arXiv:2505.23802 [cs]},
}

@misc{mistralai_mistralaimistral-7b-instruct-v03_nodate,
	title = {mistralai/{Mistral}-{7B}-{Instruct}-v0.3 · {Hugging} {Face}},
	url = {https://huggingface.co/mistralai/Mistral-7B-Instruct-v0.3},
	urldate = {2025-09-20},
	author = {MistralAI},
}

@misc{qwen_qwenqwen3-next-80b-a3b-instruct_2025,
	title = {Qwen/{Qwen3}-{Next}-{80B}-{A3B}-{Instruct} · {Hugging} {Face}},
	url = {https://huggingface.co/Qwen/Qwen3-Next-80B-A3B-Instruct},
	urldate = {2025-09-20},
	author = {Qwen},
	month = sep,
	year = {2025},
}

@misc{nvidia_nvidianvidia-nemotron-nano-9b-v2_2025,
	title = {nvidia/{NVIDIA}-{Nemotron}-{Nano}-{9B}-v2 · {Hugging} {Face}},
	url = {https://huggingface.co/nvidia/NVIDIA-Nemotron-Nano-9B-v2},
	urldate = {2025-09-20},
	author = {Nvidia},
	month = sep,
	year = {2025},
}

@inproceedings{bengio_curriculum_2009,
	address = {Montreal Quebec Canada},
	title = {Curriculum learning},
	isbn = {978-1-60558-516-1},
	url = {https://dl.acm.org/doi/10.1145/1553374.1553380},
	doi = {10.1145/1553374.1553380},
	language = {en},
	urldate = {2025-09-20},
	booktitle = {Proceedings of the 26th {Annual} {International} {Conference} on {Machine} {Learning}},
	publisher = {ACM},
	author = {Bengio, Yoshua and Louradour, Jérôme and Collobert, Ronan and Weston, Jason},
	month = jun,
	year = {2009},
	pages = {41--48},
}

@misc{fan_irreducible_2023,
	title = {Irreducible {Curriculum} for {Language} {Model} {Pretraining}},
	url = {http://arxiv.org/abs/2310.15389},
	doi = {10.48550/arXiv.2310.15389},
	urldate = {2025-09-20},
	publisher = {arXiv},
	author = {Fan, Simin and Jaggi, Martin},
	month = oct,
	year = {2023},
	note = {arXiv:2310.15389 [cs]},
}

@misc{pouransari_dataset_2025,
	title = {Dataset {Decomposition}: {Faster} {LLM} {Training} with {Variable} {Sequence} {Length} {Curriculum}},
	shorttitle = {Dataset {Decomposition}},
	url = {http://arxiv.org/abs/2405.13226},
	doi = {10.48550/arXiv.2405.13226},
	urldate = {2025-09-20},
	publisher = {arXiv},
	author = {Pouransari, Hadi and Li, Chun-Liang and Chang, Jen-Hao Rick and Vasu, Pavan Kumar Anasosalu and Koc, Cem and Shankar, Vaishaal and Tuzel, Oncel},
	month = jan,
	year = {2025},
	note = {arXiv:2405.13226 [cs]},
}

@misc{wang_dump_2025,
	title = {{DUMP}: {Automated} {Distribution}-{Level} {Curriculum} {Learning} for {RL}-based {LLM} {Post}-training},
	shorttitle = {{DUMP}},
	url = {http://arxiv.org/abs/2504.09710},
	doi = {10.48550/arXiv.2504.09710},
	urldate = {2025-09-20},
	publisher = {arXiv},
	author = {Wang, Zhenting and Cui, Guofeng and Li, Yu-Jhe and Wan, Kun and Zhao, Wentian},
	month = jun,
	year = {2025},
	note = {arXiv:2504.09710 [cs]},
}

@misc{zhang_beyond_2025,
	title = {Beyond {Random} {Sampling}: {Efficient} {Language} {Model} {Pretraining} via {Curriculum} {Learning}},
	shorttitle = {Beyond {Random} {Sampling}},
	url = {http://arxiv.org/abs/2506.11300},
	doi = {10.48550/arXiv.2506.11300},
	urldate = {2025-09-20},
	publisher = {arXiv},
	author = {Zhang, Yang and Mohamed, Amr and Abdine, Hadi and Shang, Guokan and Vazirgiannis, Michalis},
	month = jun,
	year = {2025},
	note = {arXiv:2506.11300 [cs]},
}

@misc{shao_balanced_2024,
	title = {Balanced {Data} {Sampling} for {Language} {Model} {Training} with {Clustering}},
	url = {http://arxiv.org/abs/2402.14526},
	doi = {10.48550/arXiv.2402.14526},
	urldate = {2025-09-20},
	publisher = {arXiv},
	author = {Shao, Yunfan and Li, Linyang and Fei, Zhaoye and Yan, Hang and Lin, Dahua and Qiu, Xipeng},
	month = jun,
	year = {2024},
	note = {arXiv:2402.14526 [cs]},
}

@misc{chang_survey_2023,
	title = {A {Survey} on {Evaluation} of {Large} {Language} {Models}},
	url = {http://arxiv.org/abs/2307.03109},
	doi = {10.48550/arXiv.2307.03109},
	urldate = {2025-09-21},
	publisher = {arXiv},
	author = {Chang, Yupeng and Wang, Xu and Wang, Jindong and Wu, Yuan and Yang, Linyi and Zhu, Kaijie and Chen, Hao and Yi, Xiaoyuan and Wang, Cunxiang and Wang, Yidong and Ye, Wei and Zhang, Yue and Chang, Yi and Yu, Philip S. and Yang, Qiang and Xie, Xing},
	month = dec,
	year = {2023},
	note = {arXiv:2307.03109 [cs]},
}
\bibliographystyle{arxiv}

\clearpage
\appendix

\section{Related Work}\label{app:rel_work}
\textbf{Data sampling methods for LLMs}  Data sampling has long been an important topic in machine learning, and with the rise of LLMs it has gained renewed attention.
Recent research explores diverse strategies for selecting and ordering training data to improve efficiency, scalability, and final model quality.
A first branch of approaches follows \textit{curriculum learning}\citep{bengio_curriculum_2009}, where examples are ordered from easy to hard to smooth optimization. \cite{zhang_beyond_2025} ranks data incrementally by means of different metrics for Language Models pretraining,
\cite{fan_irreducible_2023} quantifies example learnability to prioritize those that may be most beneficial and \cite{pouransari_dataset_2025} uses variable sequence length and batch-size to improve long-context modeling. \cite{wang_dump_2025} applies a similar approach for in-context learning.
Other studies draw on \textit{importance or loss-based sampling}, where training examples are weighted by estimated contribution to learning progress.
At large scale, related ideas appear in corpus-level mixture design, such as the fixed data-mixture weights used to balance sources in GPT-3 \citep{brown_language_2020}, which adjust the proportions of different domains to achieve a desired training distribution.
Some methods address corpus in a heterogeneous way through balanced or cluster-based selection, which preserves diversity by balancing common and rare examples to improve model training\citep{}.
Recently, \cite{shao_balanced_2024} introduces \textit{distribution-level curriculum learning} for LLM post-training, dynamically balancing exploration and exploitation across reward-conditioned datasets.
While these approaches significantly reduce training cost or improve data efficiency, they generally assume offline access to the full corpus or slowly changing data distributions.
They thus differ from our setting, which requires adaptive decisions on streaming data with a limited annotation budget.

\section{Datasets and LLM collections}\label{app:datasets_and_llms}
More than 130 LLMs over 10 different datasets have been employed including different generative tasks. Except for LLMs of MT-Bench, which were evaluated using BERTScore \citep{zhang_bertscore_2020}, the performance of the LLMs on the test sets of the other datasets was assessed using ROUGE-L \citep{lin_rouge_2004}.

\begin{table}[htbp]
\centering
\resizebox{\textwidth}{!}{%
\begin{tabular}{l c c c c c}
\toprule
Dataset & No. of instances & No. of LLMs & LLM scores\\
\midrule
Comprehensive--Arithmetic--Problems & 700 & 13 & $0.17$ - $0.63$\\
Calculus Dataset (FDU)              & 500 & 5  & $0.84$ - $0.95$\\
LongForm--QA (Prompt)               & 850 & 11 & $0.02$ - $0.54$\\
Grammar Correction                  & 2000 & 5 & $0.76$ - $0.84$\\
MedCalc                             & 750 & 9  & $0.12$ - $0.41$\\
MT--Bench                           & 78  & 6  & $0.76$ - $0.79$\\
SQuAD                                & 5000 & 7  & $0.47$ - $0.93$\\
SQuAD v2                             & 5000 & 5  & $0.39$ - $0.77$\\
LongForm--QA (Canonical)             & 600  & 54 & $0.16$ - $0.66$\\
FinQA                                & 500  & 22 & $0.39$ - $0.81$\\
\bottomrule
\end{tabular}%
}
\caption{Overview of datasets and LLM collections, including number of models and the scores achieved on the test set. The number of instances represents the stream length.}
\label{tab:models_datasets}
\end{table}

\begin{figure}[t]
\centering

\begin{minipage}[t]{.19\linewidth}\vspace{0pt}\centering
\includegraphics[width=\linewidth]{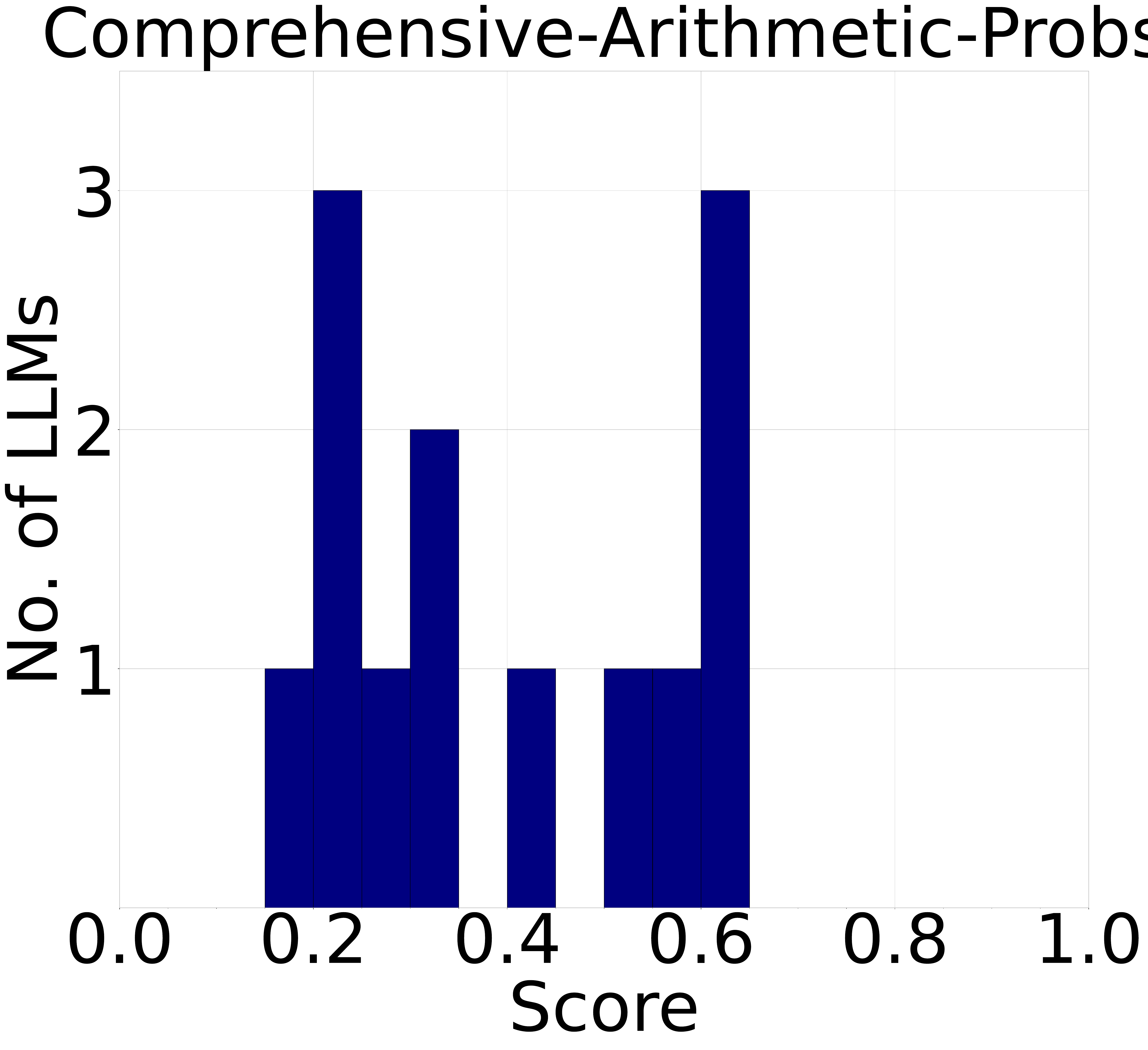}
\end{minipage}\hspace{0pt}%
\begin{minipage}[t]{.19\linewidth}\vspace{0pt}\centering
\includegraphics[width=\linewidth]{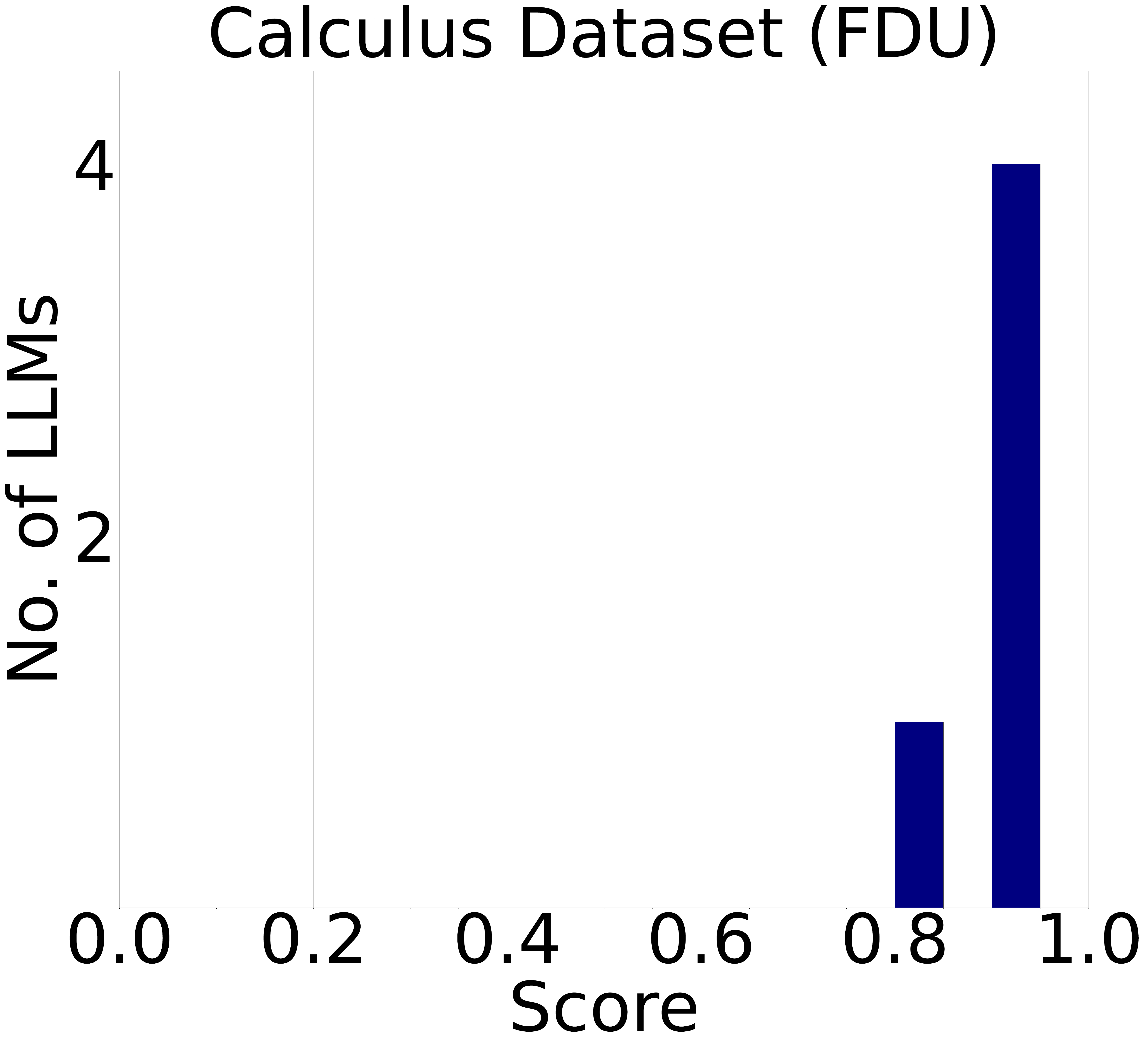}
\end{minipage}\hspace{0pt}%
\begin{minipage}[t]{.19\linewidth}\vspace{0pt}\centering
\includegraphics[width=\linewidth]{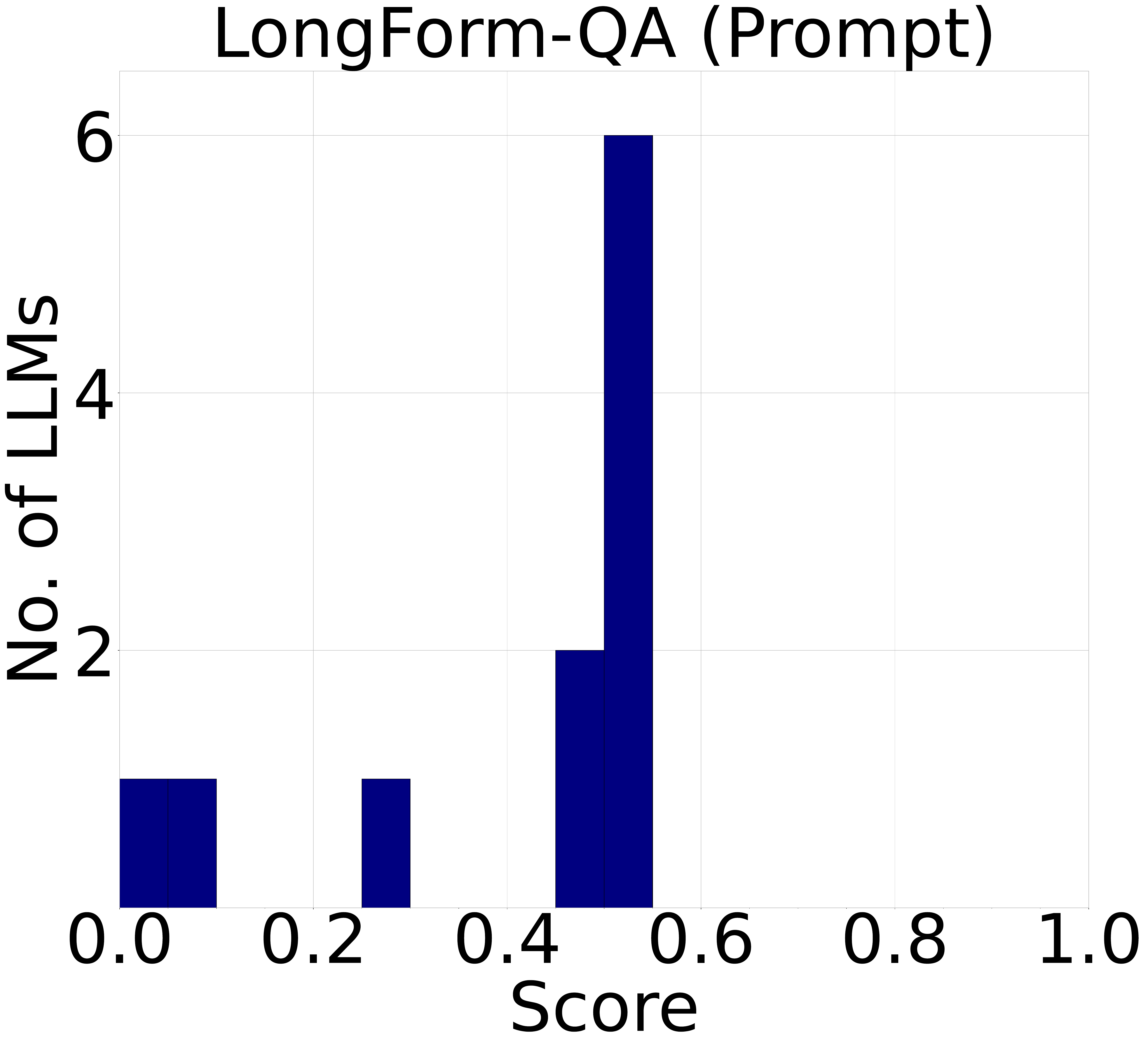}
\end{minipage}\hspace{0pt}%
\begin{minipage}[t]{.19\linewidth}\vspace{0pt}\centering
\includegraphics[width=\linewidth]{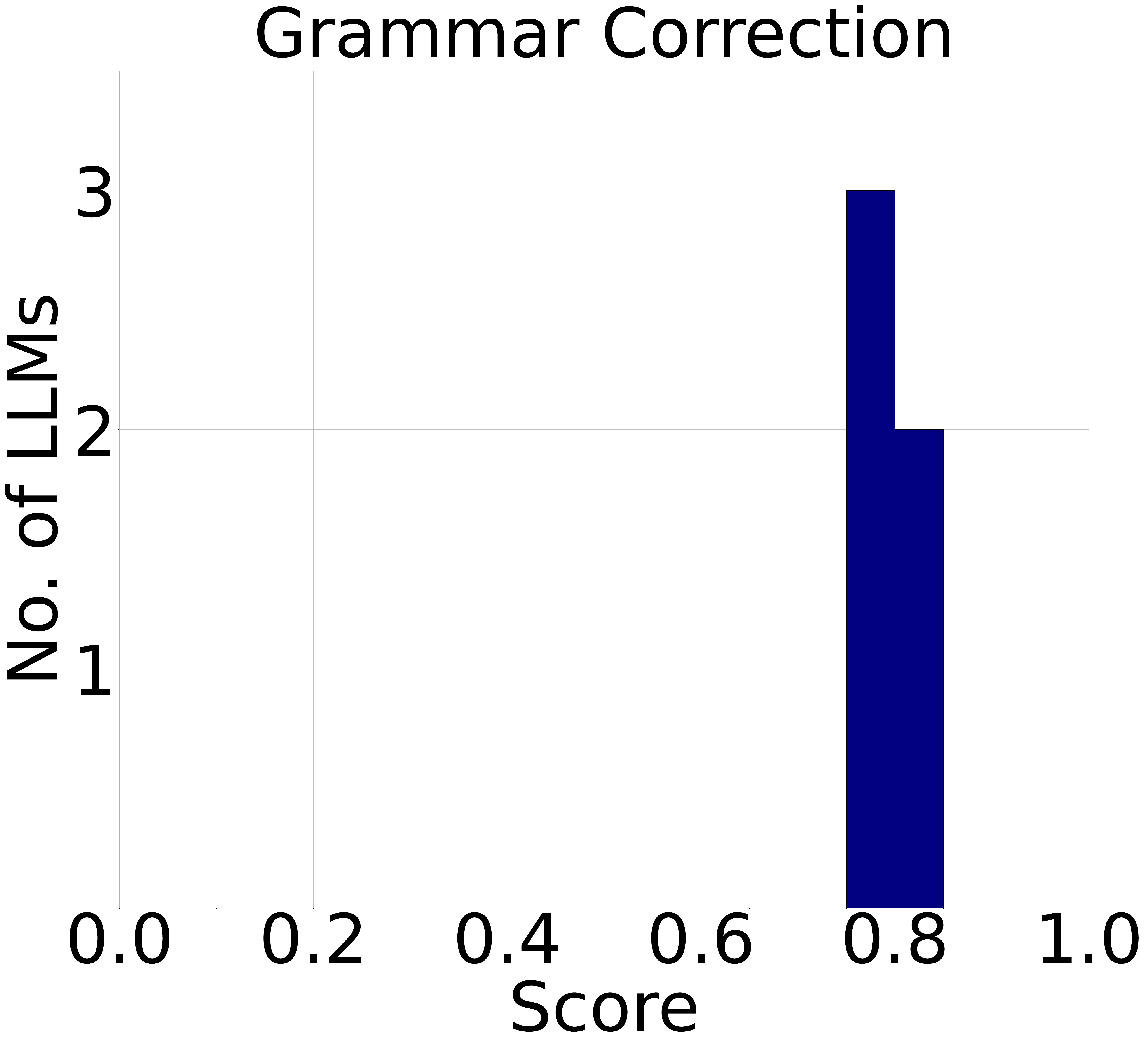}
\end{minipage}\hspace{0pt}%
\begin{minipage}[t]{.19\linewidth}\vspace{0pt}\centering
\includegraphics[width=\linewidth]{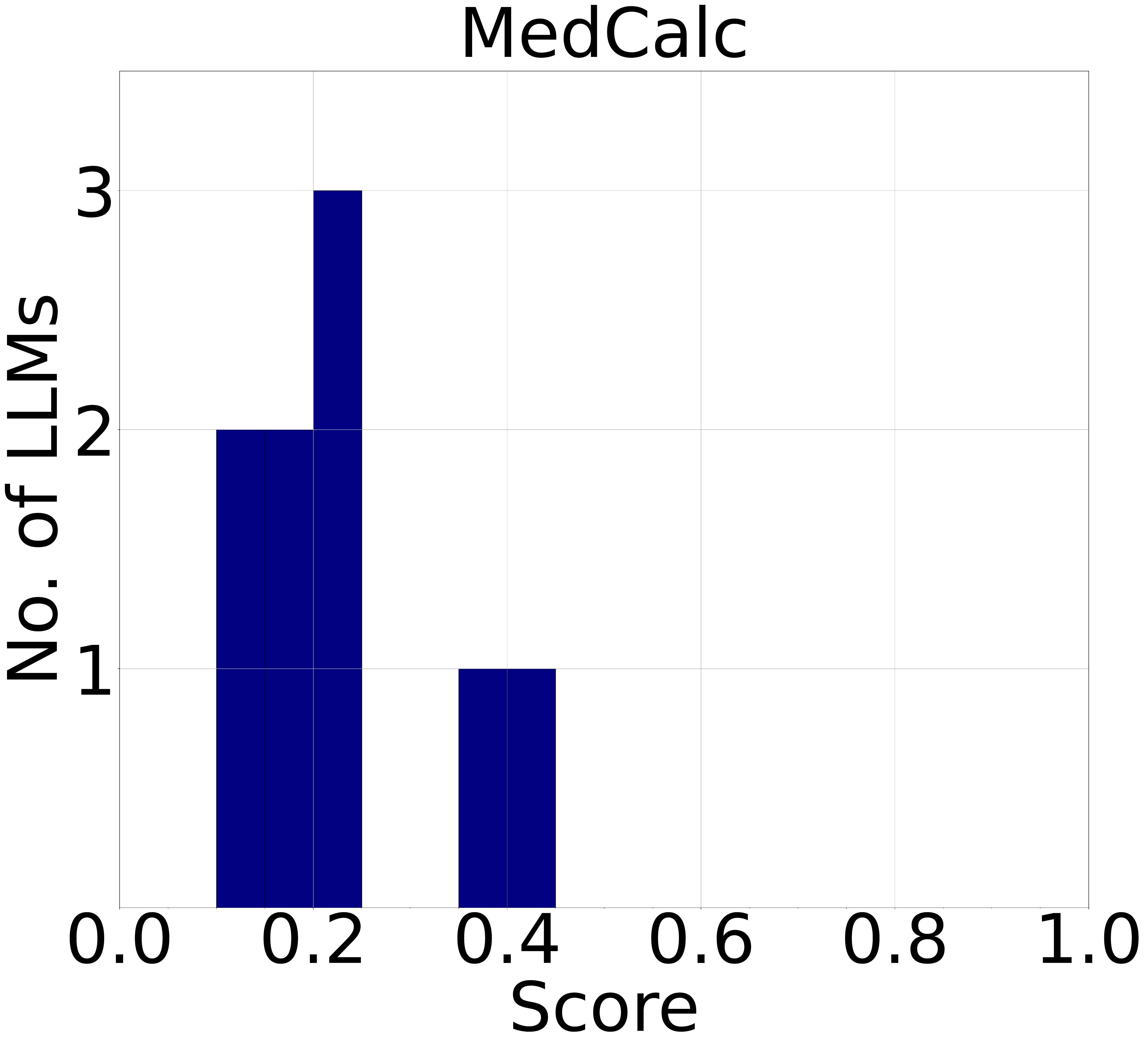}
\end{minipage}

\vspace{2pt}

\begin{minipage}[t]{.19\linewidth}\vspace{0pt}\centering
\includegraphics[width=\linewidth]{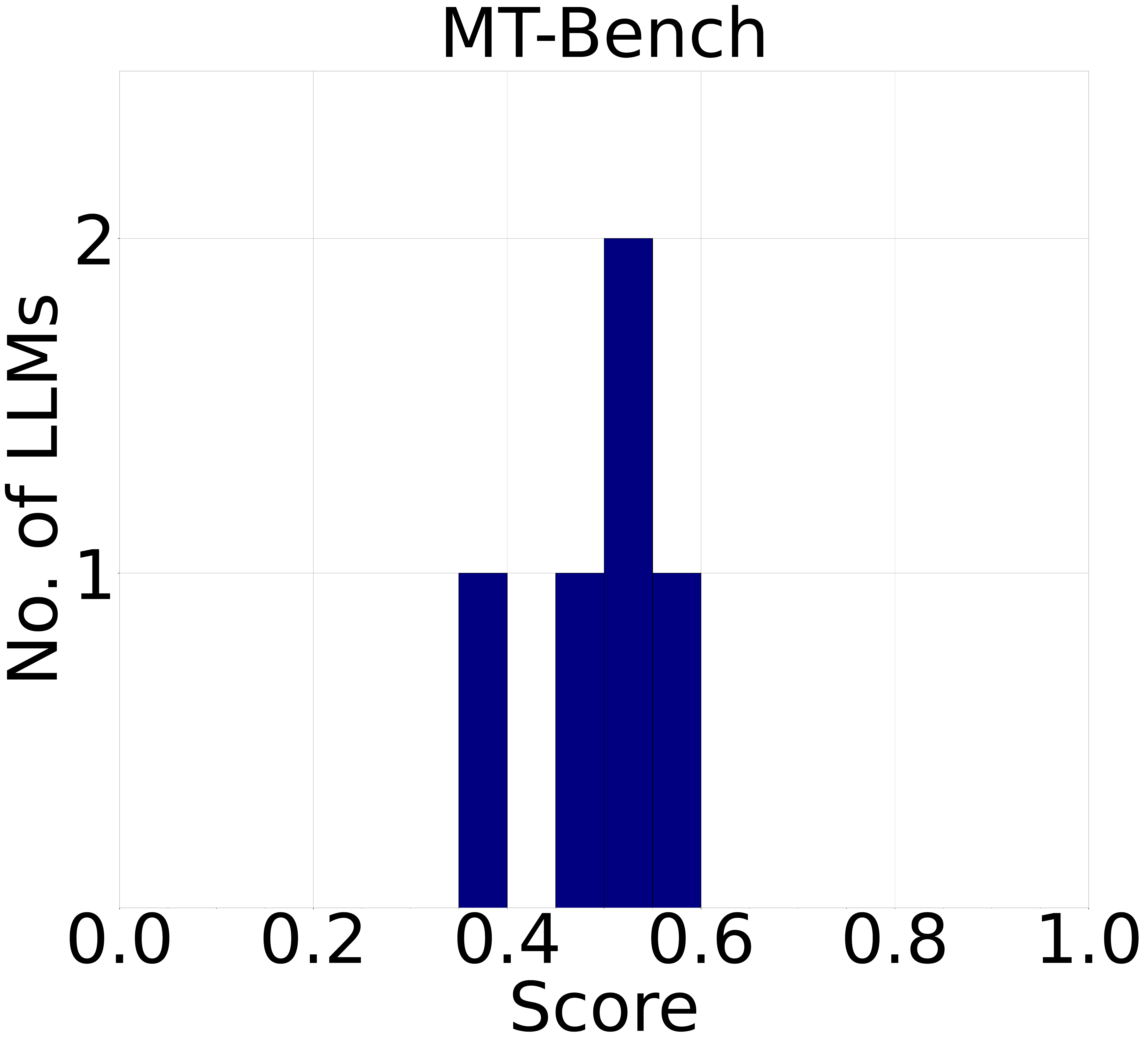}
\end{minipage}\hspace{0pt}%
\begin{minipage}[t]{.19\linewidth}\vspace{0pt}\centering
\includegraphics[width=\linewidth]{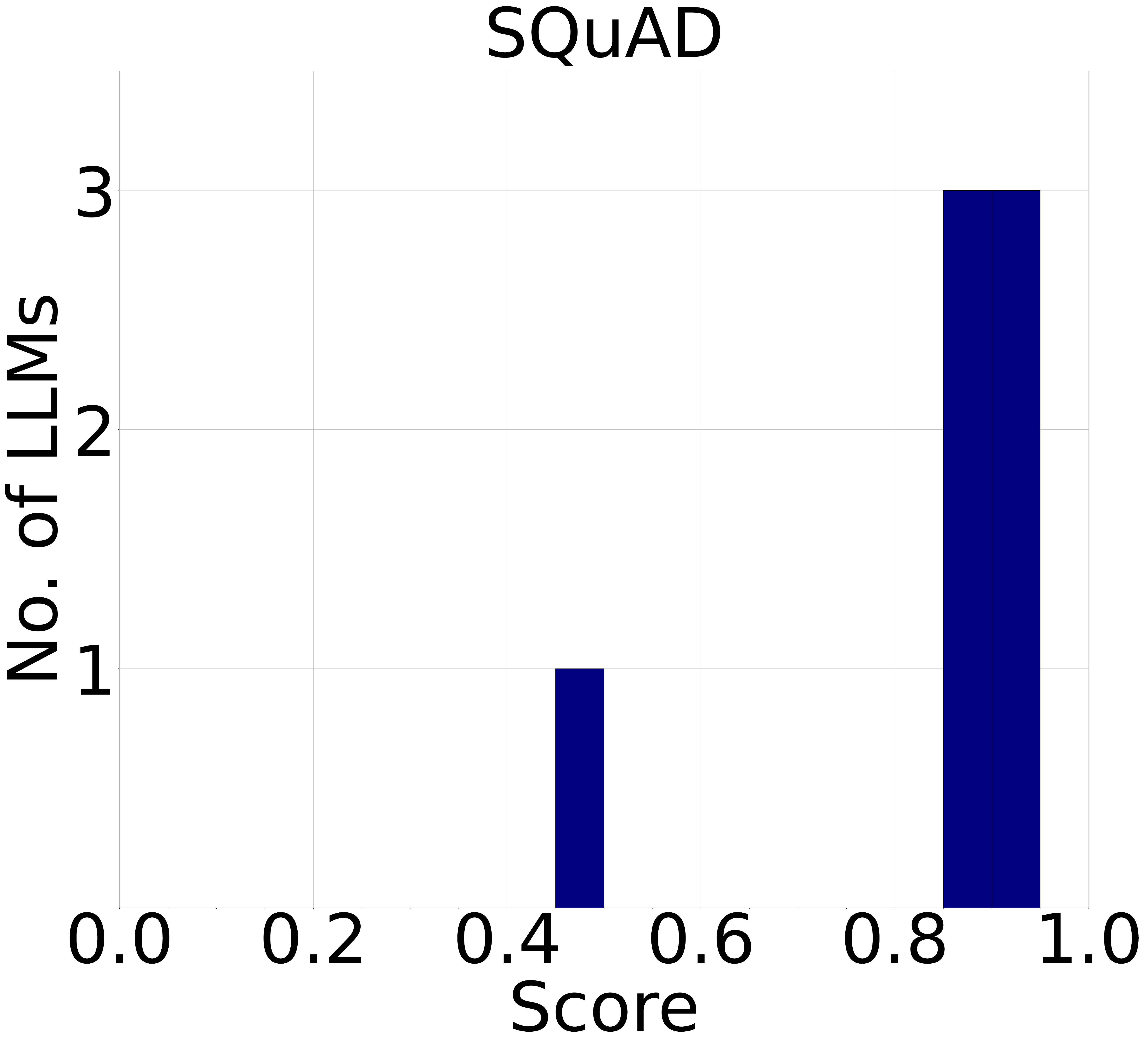}
\end{minipage}\hspace{0pt}%
\begin{minipage}[t]{.19\linewidth}\vspace{0pt}\centering
\includegraphics[width=\linewidth]{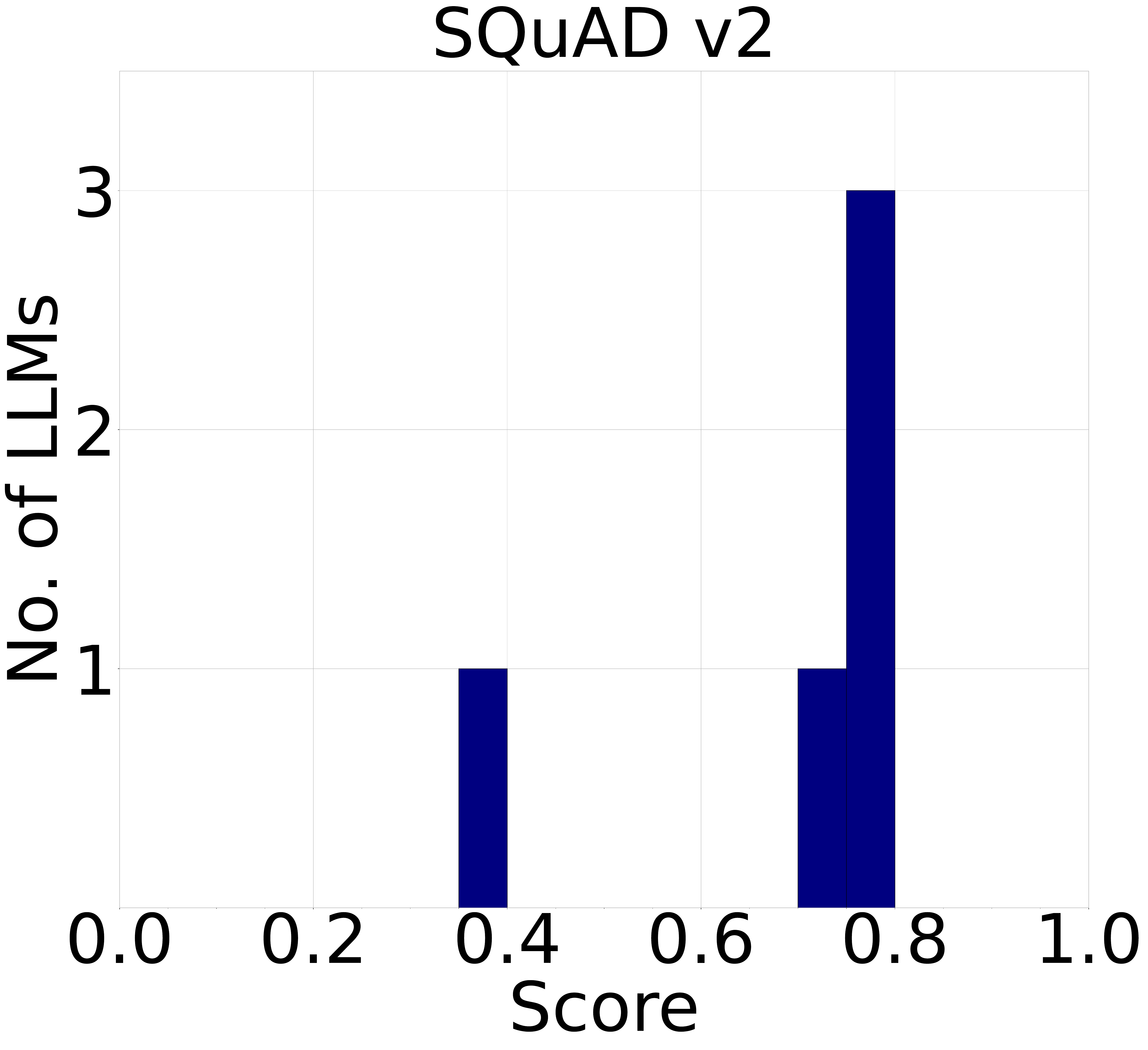}
\end{minipage}\hspace{0pt}%
\begin{minipage}[t]{.19\linewidth}\vspace{0pt}\centering
\includegraphics[width=\linewidth]{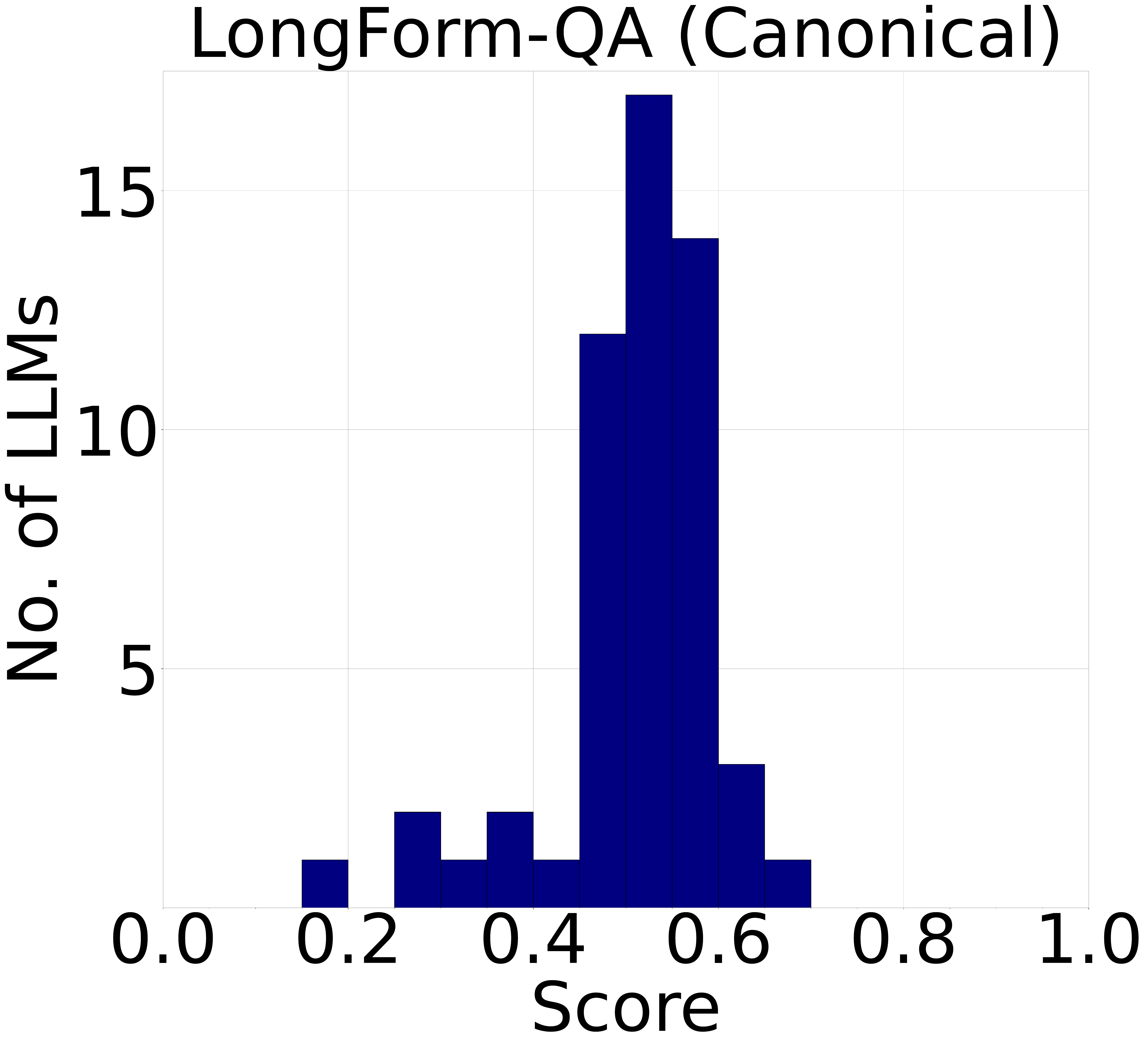}
\end{minipage}\hspace{0pt}%
\begin{minipage}[t]{.19\linewidth}\vspace{0pt}\centering
\includegraphics[width=\linewidth]{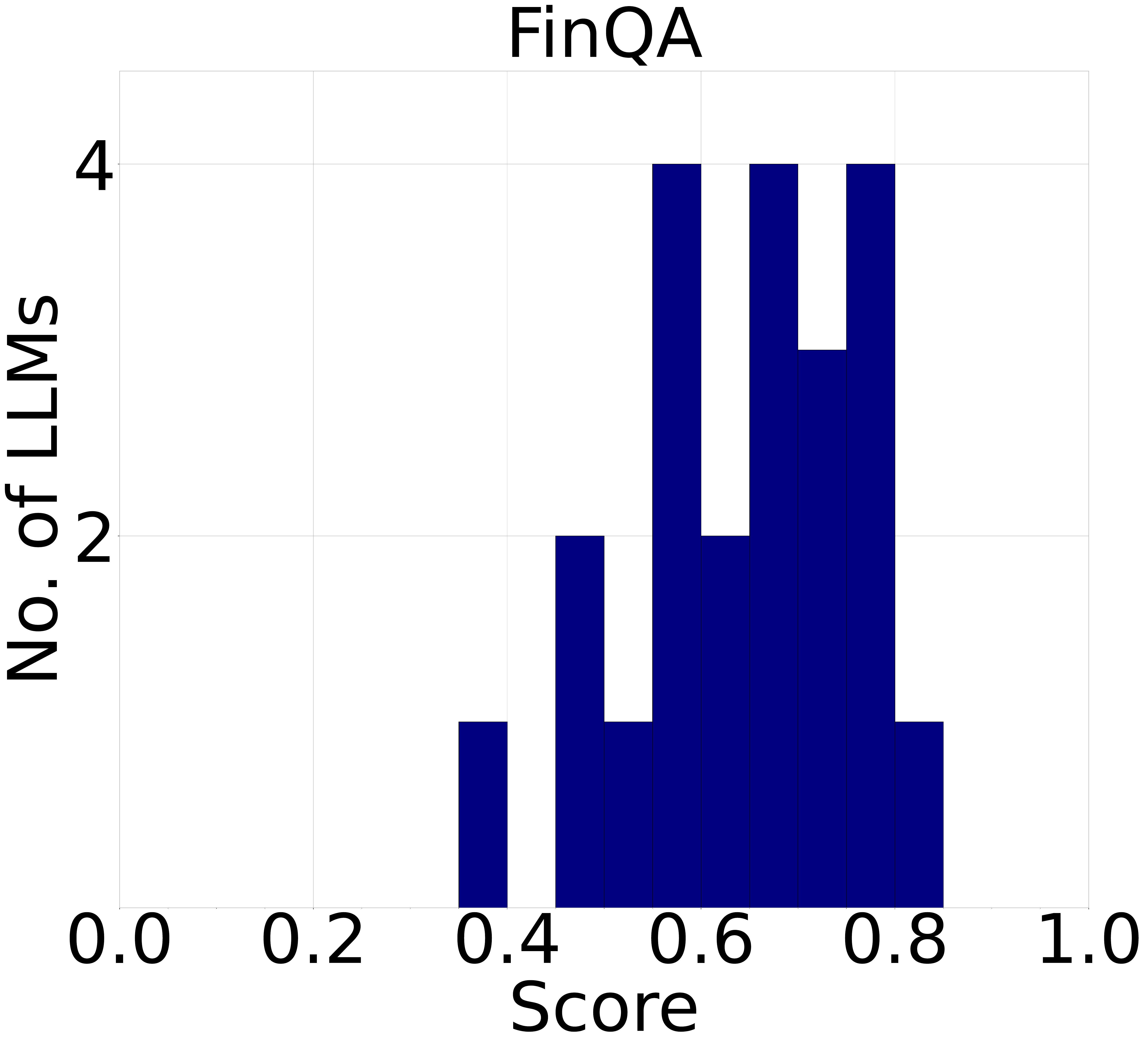}
\end{minipage}

\caption{Score of LLM collections across the datasets.}
\label{fig:acc_hist_all}
\end{figure}

Figure~\ref{fig:acc_hist_all} shows the distribution of LLM scores across datasets. Depending on the task, scores are computed with ROUGE-L~\citep{lin_rouge_2004} or BERTScore~\citep{zhang_bertscore_2020}. We evaluate a diverse collection of LLMs, yielding varied score distributions and enabling a comprehensive assessment. As reported in Table~\ref{tab:models_datasets}, the observed scores span $0.02$–$0.97$ across models and datasets.
Where available, we use benchmarks that release both LLM outputs and oracle annotations. In contrast, for the Calculus Dataset (FDU) \citep{zhang_di-zhang-fducalculus-dataset_2025}, Grammar Correction \citep{agentlans_agentlansgrammar-correction_2024}, Comprehensive Arithmetic Problems \citep{lee_garrethleecomprehensive-arithmetic-problems_2024}, SQuAD \citep{rajpurkar_squad_2016}, and SQuAD v2 \citep{rajpurkar_know_2018}, only oracle annotations are provided; therefore we ran LLMs ourselves to generate the model outputs. As detailed in Section~\ref{sec:experiments}, this includes instruction-tuned LLMs (to reduce output noise and elicit agent-specific behavior), models fine-tuned on the corresponding datasets, and off-the-shelf general LLMs. Our study is model-agnostic: we make no assumptions about LLM architectures or task-specific use cases and assess models in a manner consistent with realistic practitioner workflows.

\section{Baselines}\label{app:baselines}

\textbf{Disagreement} Similar to the Random algorithm, the probability of querying the annotation at each time instance is fixed to $q_t = b/T$. However, we restrict our attention to instances where there is significant disagreement among experts. Specifically, cases in which the answers generated by the LLMs differ substantially from one another. We quantify disagreement at time instance $t$ as the variance—across experts—of each expert’s average pairwise discrepancy from the others in the set of generated answers. Denote this statistic by $\Xi_t$. An instance is deemed to exhibit disagreement whenever $\Xi_t$ exceeds a small near-zero threshold $\delta$.

\textbf{Kullback–Leibler (KL)} We maintain a belief $\bm{p}_t$ over the $m$ models from past annotations (Hedge-style posterior). At each time instance $t$, we turn the mean of the pairwise losses among the answers generated by the LLMs into a normalized predictive loss distribution $\tilde p_t$ (via a softmax over the losses) and measure its mismatch from our belief with $D_t=\mathrm{KL}(\tilde p_t\,\|\,p_t)$ \citep{shlens_notes_2014}. The query probability is set proportional to $D_t$, so we are more likely to annotate when the present loss pattern looks uncertain under our belief and less likely when LLMs agree. We sample a Bernoulli query with bias $D_t$ and spend from a fixed budget $b$ until exhaustion. 

\textbf{Uncertainty} At each time instance \(t\), we form a per–model average loss profile \(p_t\) by averaging the current (unlabeled) pairwise prediction losses across models. We then compute an entropy–like uncertainty score
\(H_t = -\tfrac{1}{m}\sum_i p_{t,i}\log(p_{t,i})\) and set the query probability \(q_t\) proportional to \(H_t\). Thus, instances that bring higher uncertainty are queried more often. We sample a Bernoulli query with bias $H_t$ and spend from a fixed budget $b$ until exhaustion.

\section{Extended Results}\label{app:ext_res}
This section presents additional results and comprehensive data from the experiments described in Section \ref{sec:experiments}. Separately for each dataset and for each budget level, we conduct $500$ independent realizations, ensuring robust and reliable results.

\subsection{Regret}
We evaluate the regret of \framework and baselines at different budget levels where our method identifies the best LLM with high confidence. Table~\ref{tab:regret_framework} reports the budget levels used for each dataset. Across datasets, \framework\ attains consistently lower regret than the baselines. For each dataset, we compare \framework\ against the strongest competing baseline at the same budget and report the reduction factor (best-baseline regret divided by \framework\ regret). As shown in Table~\ref{tab:regret_framework}, \framework\ achieves regret reductions of $2.51\times$, $2.47\times$, $2.44\times$, and $1.97\times$ on Comprehensive--Arithmetic--Problems, SQuAD, SQuAD~v2, and FinQA, respectively.

\begin{table}[htbp]
\centering
\footnotesize
\setlength{\tabcolsep}{6pt}
\begin{tabular}{lrr}
\toprule
Dataset & Budget level & Regret \\
\midrule
Comprehensive--Arithmetic--Problems & 120  & $\downarrow\,\mathbf{2.51}\,\times$ \\
Calculus Dataset (FDU)              & 65   & $\downarrow\,\mathbf{1.93}\,\times$ \\
LongForm--QA (Prompt)               & 450  & $\downarrow\,\mathbf{1.17}\,\times$ \\
Grammar Correction                  & 1200 & $\downarrow\,\mathbf{1.01}\,\times$ \\
MedCalc                             & 400  & $\downarrow\,\mathbf{1.54}\,\times$ \\
MT--Bench                           & 50   & $\downarrow\,\mathbf{1.19}\,\times$ \\
SQuAD                               & 450  & $\downarrow\,\mathbf{2.47}\,\times$ \\
SQuAD v2                            & 800  & $\downarrow\,\mathbf{2.44}\,\times$ \\
LongForm--QA (Canonical)            & 300  & $\downarrow\,\mathbf{1.58}\,\times$ \\
FinQA                               & 200  & $\downarrow\,\mathbf{1.97}\,\times$ \\
\bottomrule
\end{tabular}
\caption{Budget levels and regret analysis.}
\label{tab:regret_framework}
\end{table}

\begin{table}[htbp]
\centering
\resizebox{\textwidth}{!}{%
\begin{tabular}{l r r r r r}
\toprule
Dataset & Online LLM Picker & Kullback--Leibler & Uncertainty & Disagreement & Random \\
\midrule
Comprehensive-Arithmetic-Problems & \textbf{3.67} & 35.39 & 10.09 & 9.89 & \underline{9.22} \\
Calculus Dataset                   & \textbf{0.82} & 7.75  & \underline{1.58} & 3.62 & 3.01 \\
LongForm-QA (Prompt)                       & \textbf{8.45} & 24.06 & 10.31 & \underline{9.88} & 9.97 \\
Grammar Correction                 & \textbf{2.25} & 5.07  & \underline{2.27} & 2.52 & 2.31 \\
MedCalc                            & \textbf{4.57} & 29.18 & 9.67  & \underline{7.03} & 7.07 \\
MT-Bench                           & \textbf{1.72} & 2.71  & \underline{2.06} & 2.29 & 2.14 \\
SQuAD                              & \textbf{8.78} & 79.70 & \underline{21.67} & 28.93 & 29.58 \\
SQuAD v2                           & \textbf{5.98} & 48.45 & \underline{14.58} & 18.83 & 19.31 \\
LongForm--QA (Canonical)           & \textbf{5.39} & 26.98 & 9.63  & 8.88 & \underline{8.53} \\
FinQA                              & \textbf{3.10} & 21.04 & 14.83 & \underline{6.10} & 6.44 \\
\bottomrule
\end{tabular}%
}
\label{tab:ext_regret}
\caption{Regret of \framework and baselines. Bold denotes the best value, underlining denotes the second-best. The regret of \framework is consistently lower with respect to baselines.}
\end{table}

\subsection{Identification Probability}
An overview of the percentage of budget needed for each method to reach the 100\% identification probability for the first time can be found in Table \ref{tab:pctg_annotations}. Note that the percentage of budget for each method is now computed with respect to the total length of the stream. \framework consistently requires fewer annotations compared to baselines to reach the maximum identification probability, showing its effectiveness in properly selecting the most informative annotations to improve the selection strategy.
\begin{table}[htbp]
\centering
\resizebox{\textwidth}{!}{%
\begin{tabular}{l r r r r r}
\toprule
Dataset & Online LLM Picker & Kullback--Leibler & Uncertainty & Disagreement & Random \\
\midrule
Comprehensive-Arithmetic-Problems & \textbf{28.26}\% & 97.10\% & 100.00\% & \underline{91.30}\% & 95.65\% \\
Calculus Dataset & \textbf{15.00}\% & \underline{80.00}\% & 38.00\% & \underline{80.00}\% & 90.00\% \\
LF-QA-Prompt & \textbf{78.82}\% & 100.00\% & \underline{99.41}\% & 100.00\% & \underline{99.41}\% \\
Grammar Correction & \textbf{74.00}\% & 98.00\% & \underline{74.50}\% & 100.00\% & 98.00\% \\
MedCalc & \textbf{78.00}\% & 100.00\% & \underline{98.00}\% & 100.00\% & 100.00\% \\
MT-Bench & \textbf{71.79}\% & 96.15\% & 98.71\% & \underline{92.30}\% & 93.58\% \\
SQuAD & \textbf{13.00}\% & 90.00\% & \underline{29.00}\% & 80.00\% & 58.00\% \\
SQuAD v2 & \textbf{17.00}\% & 100.00\% & \underline{60.00}\% & 90.00\% & 80.00\% \\
LongForm--QA (Canonical)  & \textbf{84.16}\% & 100.00\% & 98.33\% & \underline{96.67}\% & 99.17\% \\
FinQA & \textbf{47.00}\% & 88.00\% & 100.00\% & 84.00\% & \underline{72.00}\% \\
\bottomrule
\end{tabular}%
}
\caption{Percentage of annotations (w.r.t. total stream length) required by \framework and baselines to reach 100\% identification probability. Bold denotes the best value, underlining denotes the second-best.}
\label{tab:pctg_annotations}
\end{table}

 \subsection{Robustness Analysis}
 We compute the 95th percentile accuracy gap at the budget needed by \framework to reach certain identification probability levels.
 Specifically, we focus on 70\%, 80\%, 90\% and 100\% identification probabilities.
 If the exact desired identification probability value is unavailable, the next higher closest value is used.
 Table \ref{tab:rob_analysis} shows the results. Best values are in bold, second-best values are underlined. In $36$ cases out of $40$, \framework achieves the lowest 95th percentile accuracy gap. However, even for the four remaining cases, \framework performs competitively, achieving the second lowest accuracy gap.
\FloatBarrier
\begingroup
\small
\setlength{\tabcolsep}{4pt}
\begin{longtable}{@{}>{\raggedright\arraybackslash}p{.25\linewidth}c*{5}{r}@{}}
\toprule
Dataset & \shortstack{Identification\\probability} & \shortstack{Online LLM\\Picker} & \shortstack{Kullback--\\Leibler} & Uncertainty & Disagreement & Random \\
\midrule
\endfirsthead
\multicolumn{7}{l}{\small\textit{Table \thetable{} (continued)}}\\[4pt]
\toprule
Dataset & \shortstack{Identification\\probability} & \shortstack{Online LLM\\Picker} & \shortstack{Kullback--\\Leibler} & Uncertainty & Disagreement & Random \\
\midrule
\endhead
\midrule
\multicolumn{7}{r}{\small\textit{Continued on the next page}}\\
\endfoot
\endlastfoot
\multirow[t]{4}{=}{Comprehensive-Arithmetic-Problems} & 70\% & \underline{1.81} & 3.71 & 2.10 & \textbf{1.62} & \underline{1.81} \\*
 & 80\% & \textbf{1.33} & 2.67 & 1.71 & \underline{1.62} & 1.71 \\*
 & 90\% & \textbf{0.86} & 2.57 & 1.71 & \underline{1.33} & 1.52 \\*
 & 100\% & \textbf{0.00} & 2.19 & 1.43 & 1.43 & \underline{1.24} \\
\addlinespace[6pt]
\multirow[t]{4}{=}{Calculus Dataset} & 70\% & \underline{1.12} & 3.11 & \textbf{1.02} & 2.92 & 2.81 \\*
 & 80\% & \textbf{0.91} & 3.03 & \underline{0.95} & 2.65 & 2.46 \\*
 & 90\% & \textbf{0.66} & 2.85 & \underline{0.87} & 1.22 & 1.10 \\*
 & 100\% & \textbf{0.00} & 2.76 & \textbf{0.00} & \underline{0.88} & 0.91 \\
\addlinespace[6pt]
\multirow[t]{4}{=}{LongForm--QA (Prompt)} & 70\% & \textbf{1.45} & 2.89 & 1.72 & \underline{1.53} & 1.59 \\*
 & 80\% & \textbf{1.23} & 1.82 & 1.43 & 1.54 & \underline{1.37} \\*
 & 90\% & \textbf{0.63} & 1.68 & 1.35 & \underline{1.26} & 1.30 \\*
 & 100\% & \textbf{0.00} & 0.98 & 0.85 & 0.69 & \underline{0.65} \\
\addlinespace[6pt]
\multirow[t]{4}{=}{Grammar Correction} & 70\% & \textbf{0.11} & \textbf{0.11} & \textbf{0.11} & \textbf{0.11} & \textbf{0.11} \\*
 & 80\% & \textbf{0.11} & \textbf{0.11} & \textbf{0.11} & \textbf{0.11} & \textbf{0.11} \\*
 & 90\% & \textbf{0.11} & \textbf{0.11} & \textbf{0.11} & \textbf{0.11} & \textbf{0.11} \\*
 & 100\% & \textbf{0.00} & \underline{0.11} & \textbf{0.00} & \underline{0.11} & \underline{0.11} \\
\addlinespace[6pt]
\multirow[t]{4}{=}{MedCalc} & 70\% & \underline{2.13} & 4.40 & 2.55 & 2.37 & \textbf{2.10} \\*
 & 80\% & \textbf{1.84} & 2.62 & 2.45 & 2.19 & \underline{1.99} \\*
 & 90\% & \textbf{0.99} & 2.24 & 2.44 & 1.99 & \underline{1.79} \\*
 & 100\% & \textbf{0.00} & \underline{0.48} & 1.59 & 0.89 & \textbf{0.00} \\
\addlinespace[6pt]
\multirow[t]{4}{=}{MT-Bench} & 70\% & \textbf{6.61} & 7.59 & 7.03 & \underline{6.93} & 6.99 \\*
 & 80\% & \textbf{3.86} & 7.50 & 4.78 & 6.42 & \underline{4.61} \\*
 & 90\% & \textbf{3.28} & 6.73 & 3.92 & \underline{3.83} & 3.86 \\*
 & 100\% & \textbf{0.00} & 3.79 & 3.71 & \underline{2.82} & 3.01 \\
\addlinespace[6pt]
\multirow[t]{4}{=}{SQuAD} & 70\% & \underline{1.69} & 2.05 & \textbf{1.54} & 1.74 & 1.76 \\*
 & 80\% & \textbf{1.33} & 1.95 & \underline{1.48} & 1.68 & 1.66 \\*
 & 90\% & \textbf{1.12} & 2.15 & \underline{1.37} & 1.61 & 1.61 \\*
 & 100\% & \textbf{0.00} & 1.72 & \underline{0.57} & 1.24 & 1.24 \\
\addlinespace[6pt]
\multirow[t]{4}{=}{SQuAD v2} & 70\% & \textbf{1.30} & 1.77 & \underline{1.31} & 1.49 & 1.39 \\*
 & 80\% & \textbf{0.92} & 1.58 & \underline{1.21} & 1.28 & 1.29 \\*
 & 90\% & \textbf{0.65} & 1.44 & \underline{0.94} & 1.18 & 1.10 \\*
 & 100\% & \textbf{0.00} & 1.06 & \underline{0.78} & 0.92 & 0.85 \\
\addlinespace[6pt]
\multirow[t]{4}{=}{LongForm--QA (Canonical)} & 70\% & \textbf{2.91} & 7.78 & 4.11 & \underline{3.21} & 3.51 \\*
 & 80\% & \textbf{2.47} & 7.46 & 3.42 & \underline{2.79} & \textbf{3.46} \\*
 & 90\% & \textbf{1.41} & 4.22 & 2.59 & 2.01 & \underline{1.96} \\*
 & 100\% & \textbf{0.00} & \textbf{0.00} & \underline{0.43} & \textbf{0.00} & \textbf{0.00} \\
\addlinespace[6pt]
\multirow[t]{4}{=}{FinQA} & 70\% & \textbf{3.56} & 6.72 & \underline{3.67} & \underline{3.67} & \textbf{3.56} \\*
 & 80\% & \textbf{2.92} & 4.26 & 3.67 & 3.33 & \underline{3.29} \\*
 & 90\% & \textbf{1.99} & 5.21 & 3.83 & \underline{2.74} & 2.98 \\*
 & 100\% & \textbf{0.00} & \underline{2.74} & 3.96 & \textbf{0.00} & \textbf{0.00} \\
\bottomrule
\caption{Robustness analysis: 95th percentile accuracy gap at the budget needed for \framework to reach identification probabilities of 70\%, 80\%, 90\%, and 100\%. Bold denotes the best value, underlining denotes the second-best.}
\label{tab:rob_analysis}\\
\end{longtable}
\endgroup

\subsection{Annotation Efficiency}
For completeness, we show the extended plots related to the results of annotation efficiency described in Table \ref{tab:delta_vicin}.

\begin{figure}[t]
\centering
\includegraphics[width=0.9\linewidth]{images/legend.pdf}\\[-2pt]

\begin{minipage}[t]{.19\linewidth}\vspace{0pt}\centering
\includegraphics[width=\linewidth]{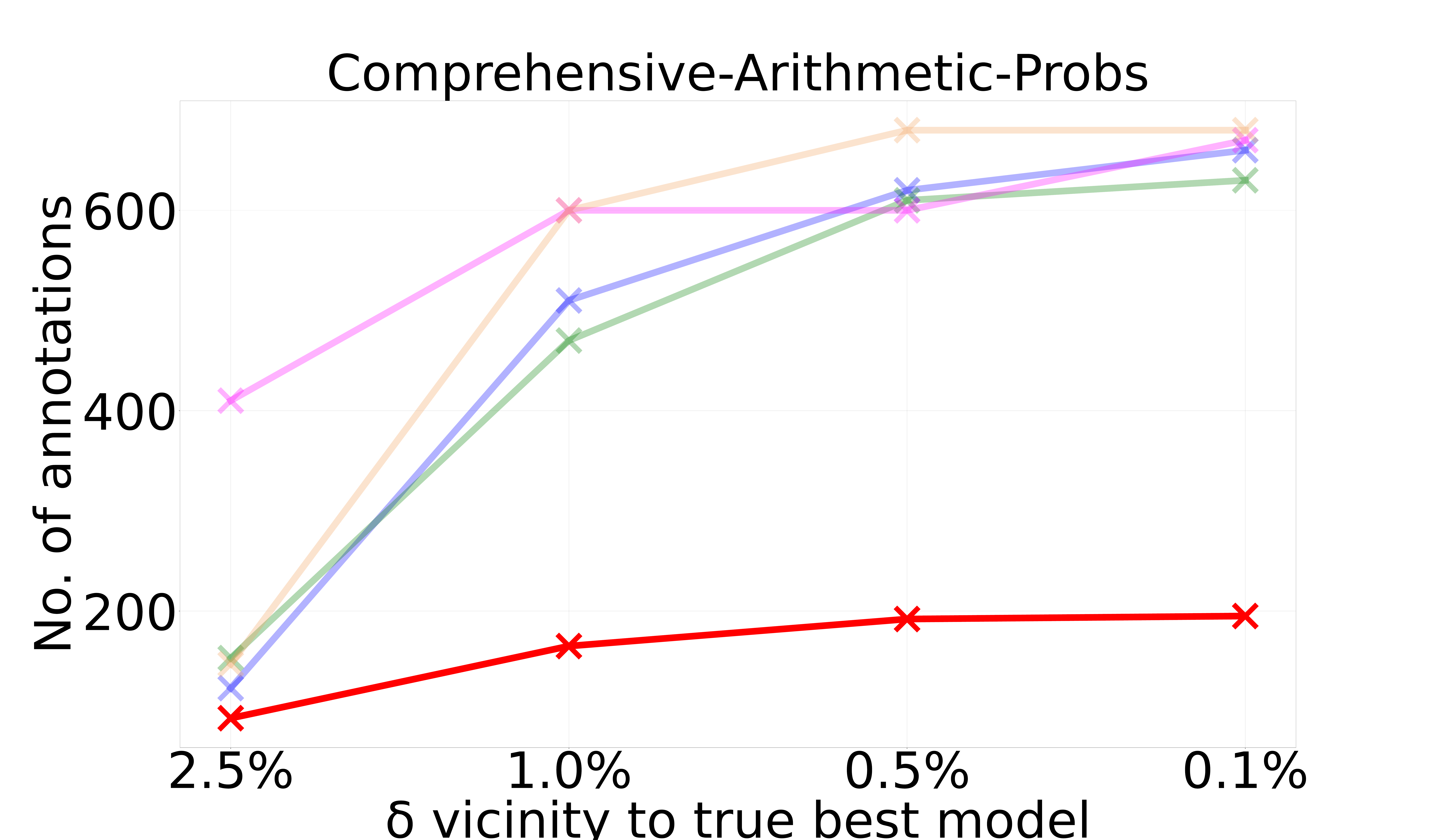}
\end{minipage}\hspace{0pt}%
\begin{minipage}[t]{.19\linewidth}\vspace{0pt}\centering
\includegraphics[width=\linewidth]{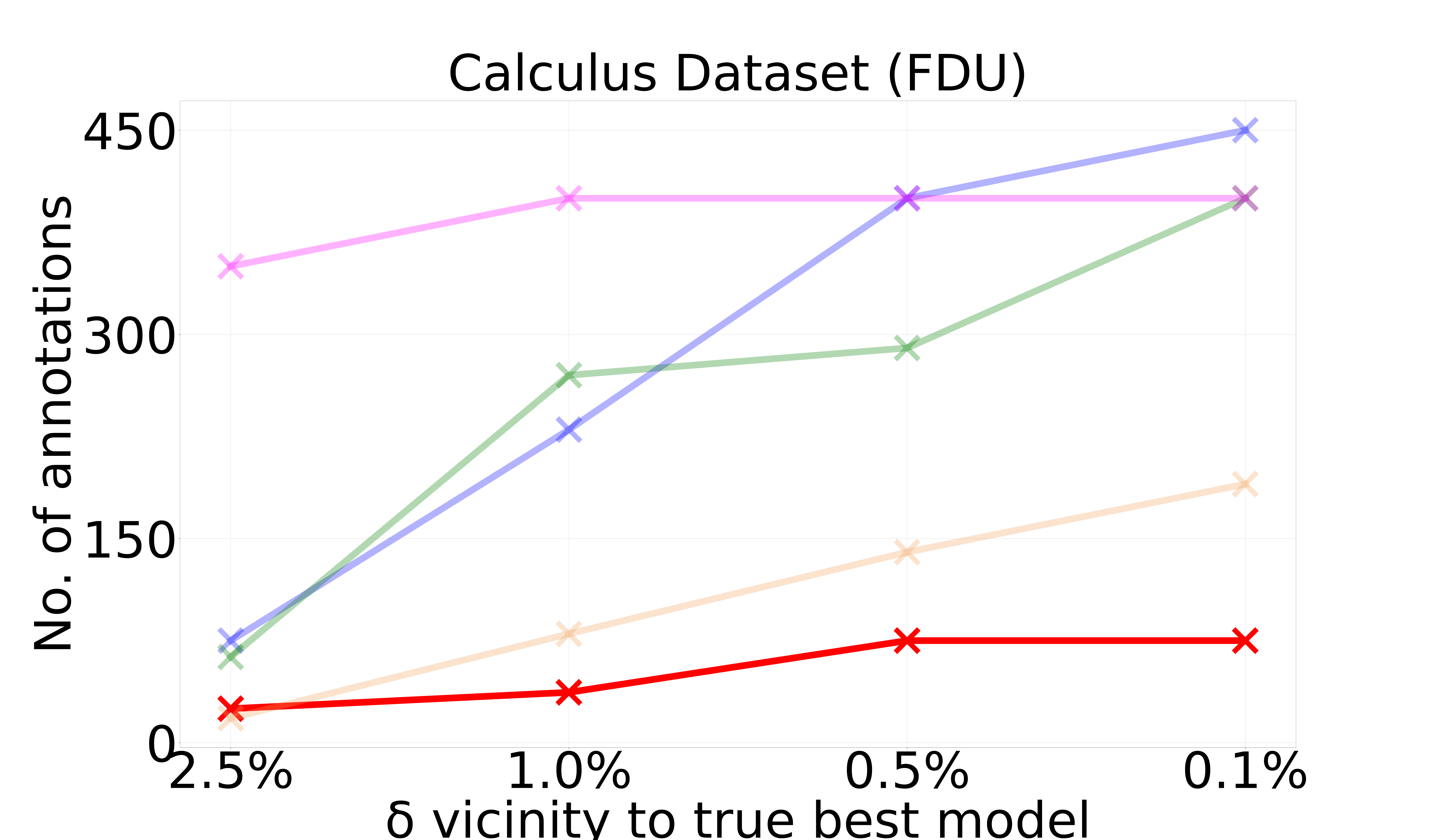}
\end{minipage}\hspace{0pt}%
\begin{minipage}[t]{.19\linewidth}\vspace{0pt}\centering
\includegraphics[width=\linewidth]{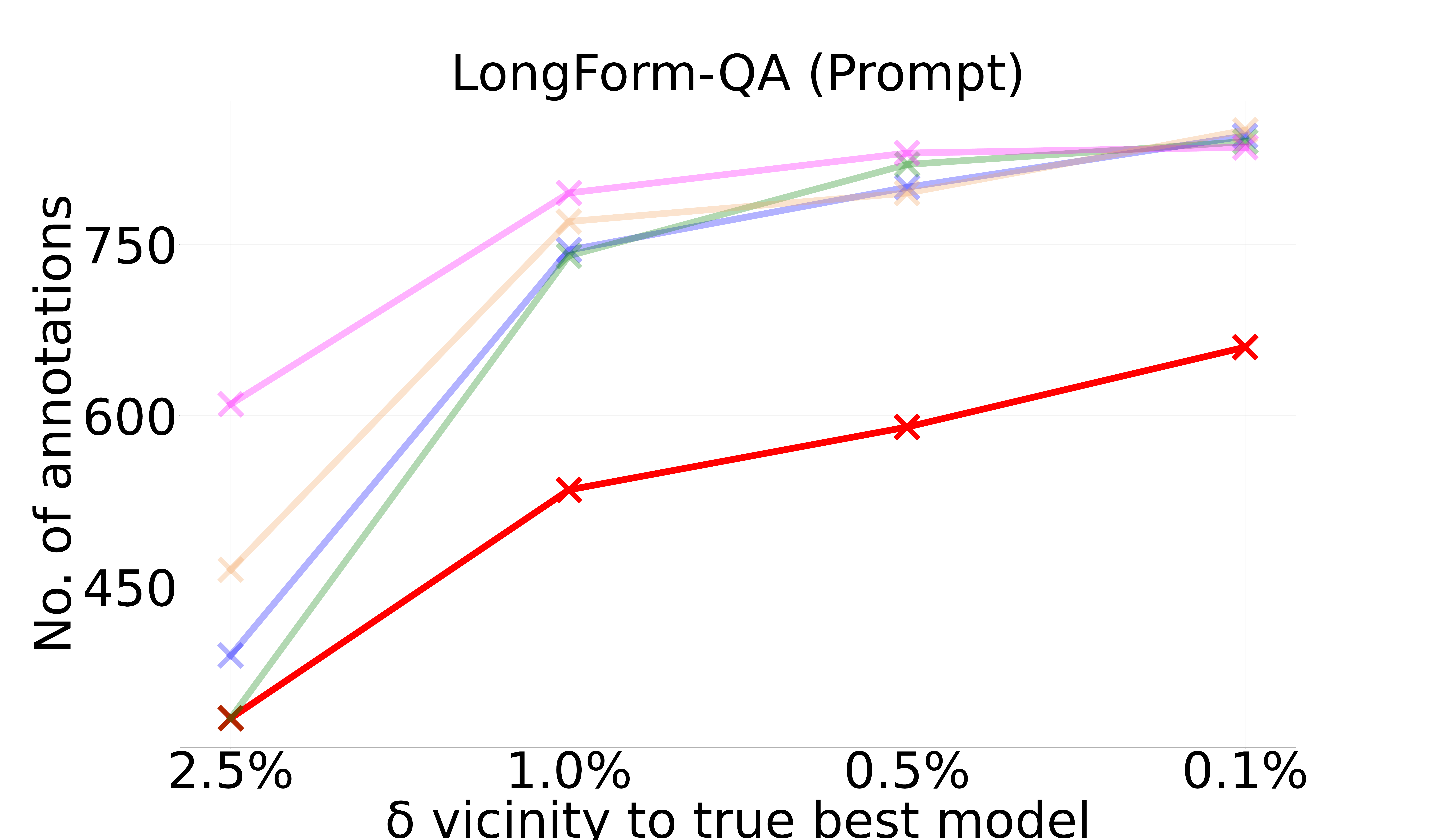}
\end{minipage}\hspace{0pt}%
\begin{minipage}[t]{.19\linewidth}\vspace{0pt}\centering
\includegraphics[width=\linewidth]{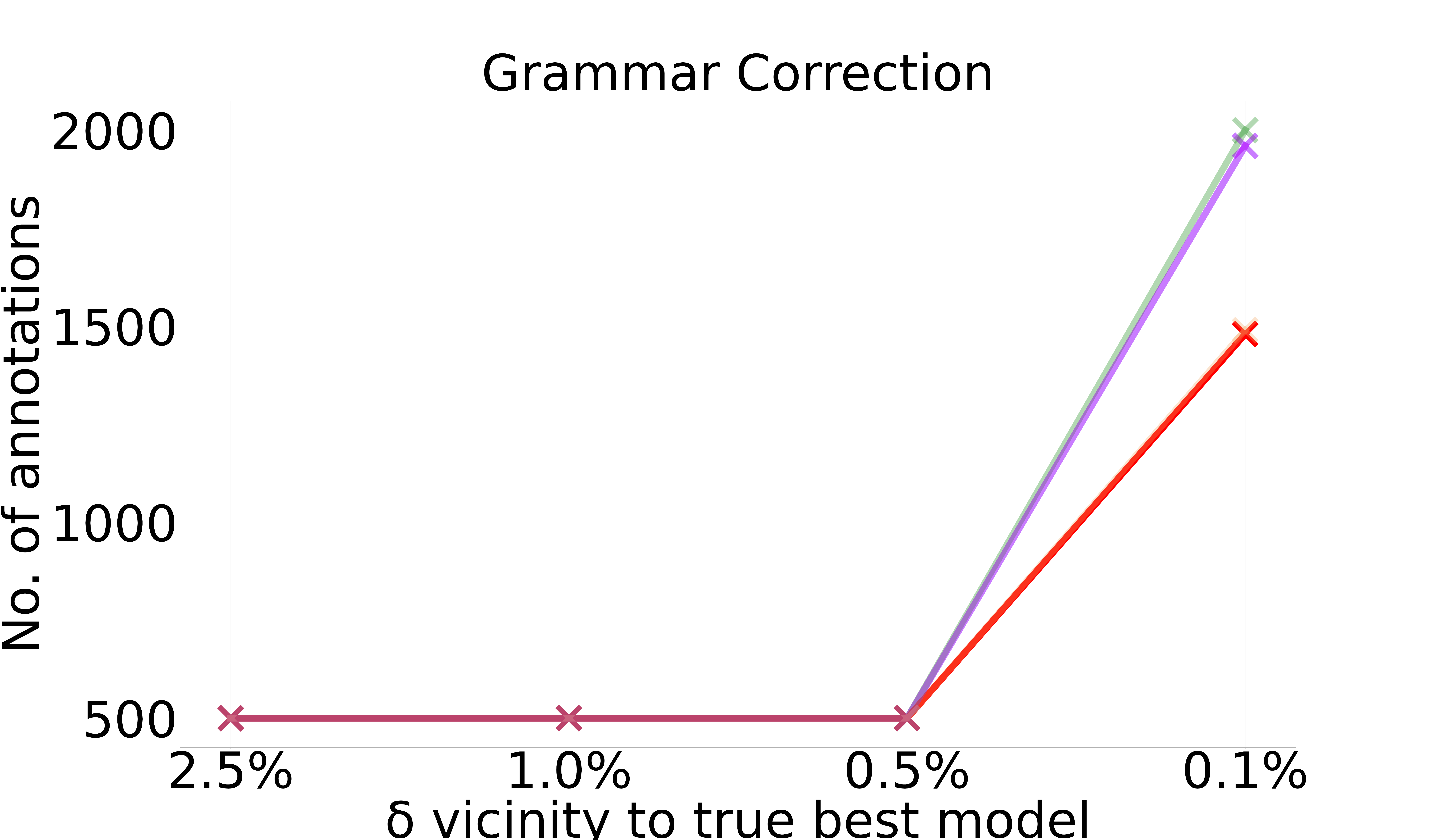}
\end{minipage}\hspace{0pt}%
\begin{minipage}[t]{.19\linewidth}\vspace{0pt}\centering
\includegraphics[width=\linewidth]{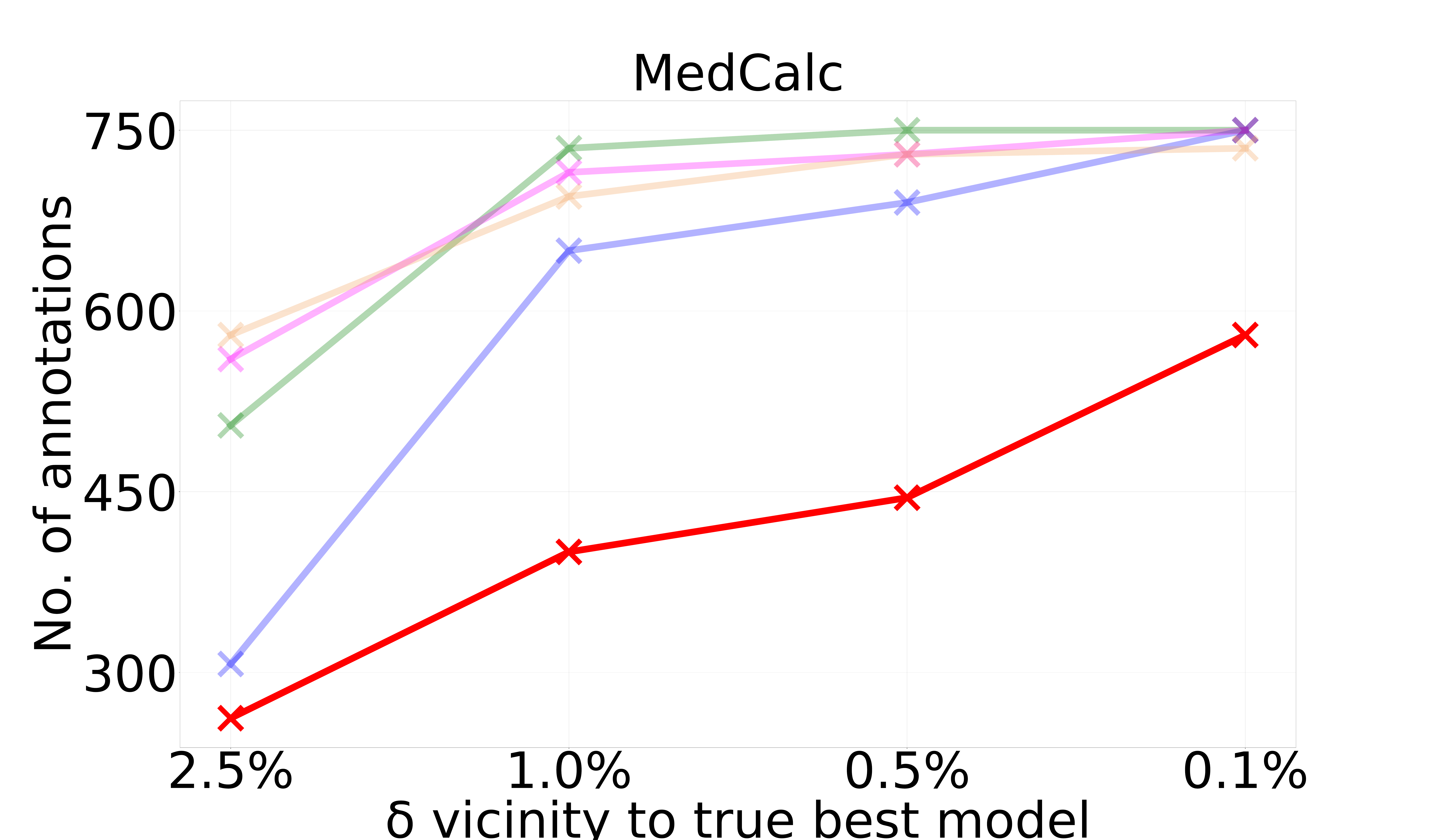}
\end{minipage}

\vspace{2pt}

\begin{minipage}[t]{.19\linewidth}\vspace{0pt}\centering
\includegraphics[width=\linewidth]{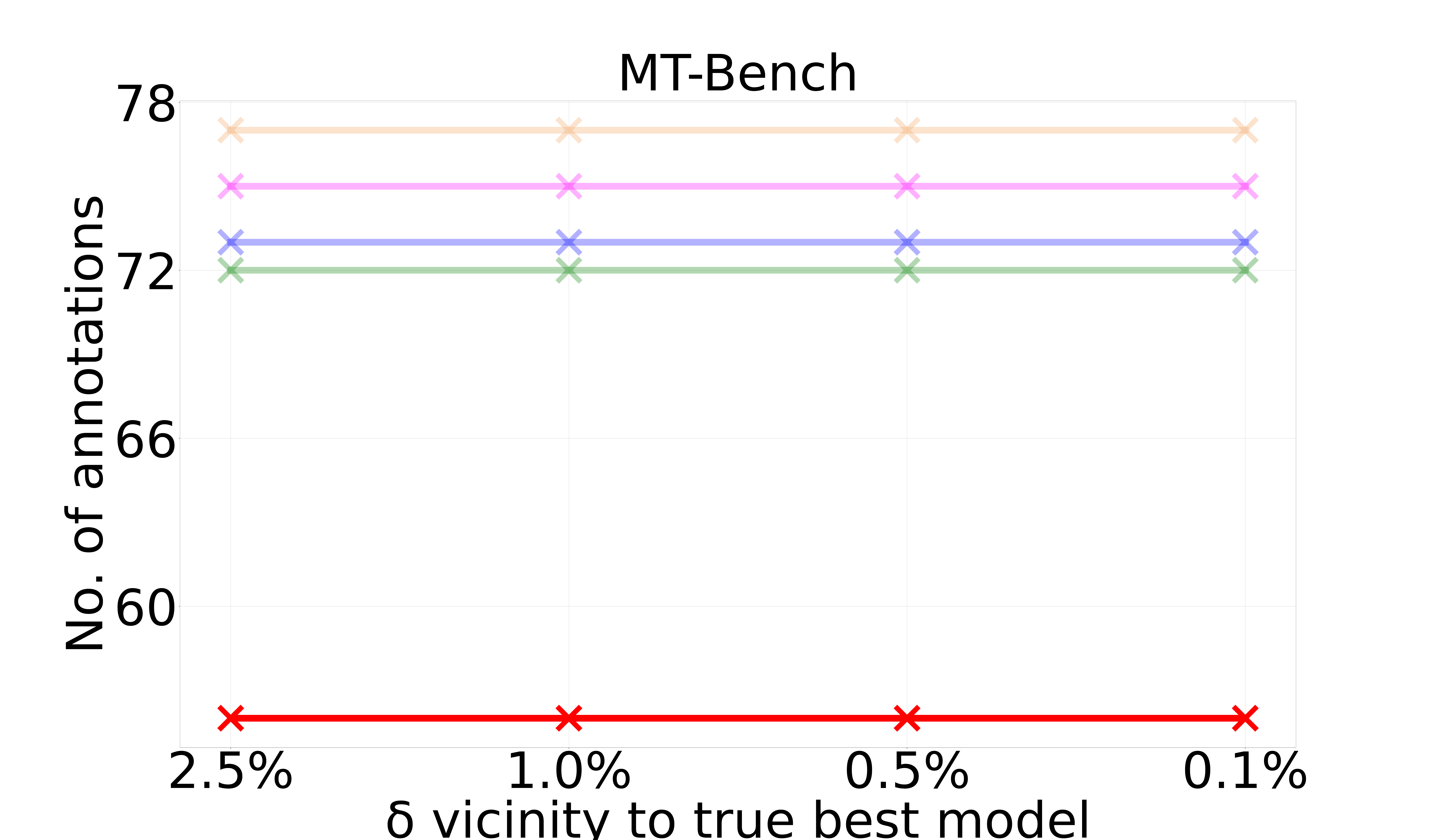}
\end{minipage}\hspace{0pt}%
\begin{minipage}[t]{.19\linewidth}\vspace{0pt}\centering
\includegraphics[width=\linewidth]{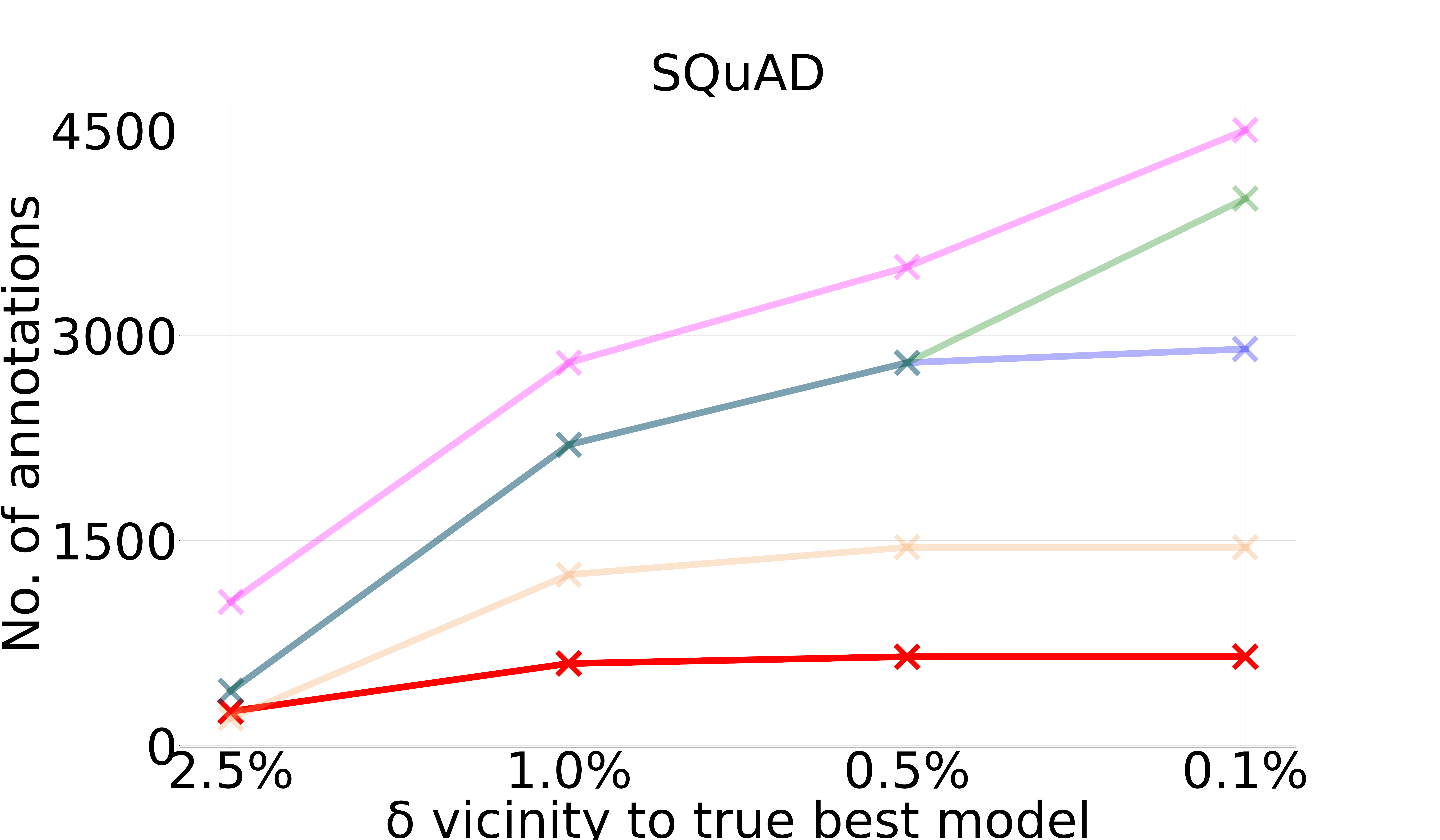}
\end{minipage}\hspace{0pt}%
\begin{minipage}[t]{.19\linewidth}\vspace{0pt}\centering
\includegraphics[width=\linewidth]{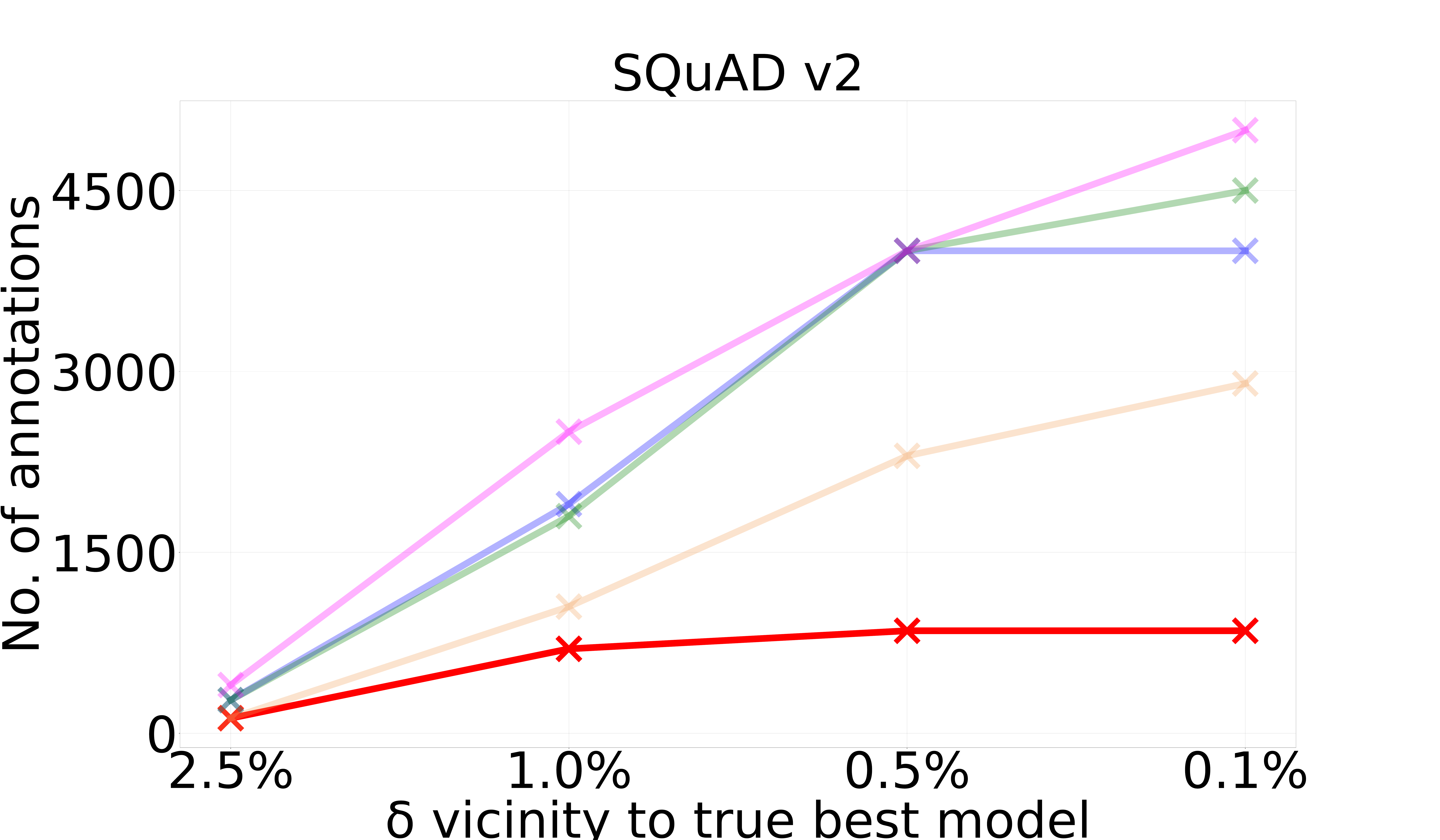}
\end{minipage}\hspace{0pt}%
\begin{minipage}[t]{.19\linewidth}\vspace{0pt}\centering
\includegraphics[width=\linewidth]{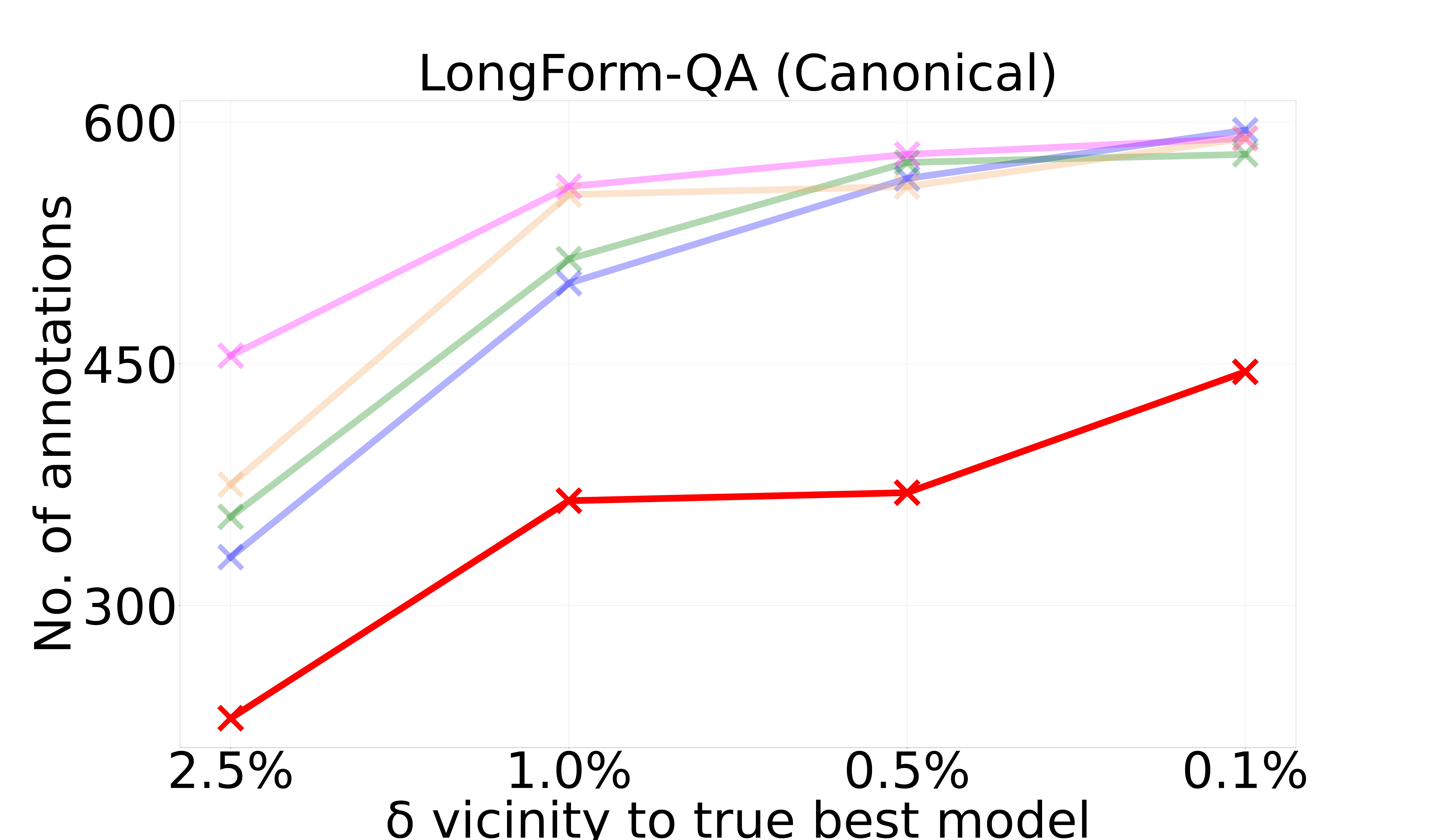}
\end{minipage}\hspace{0pt}%
\begin{minipage}[t]{.19\linewidth}\vspace{0pt}\centering
\includegraphics[width=\linewidth]{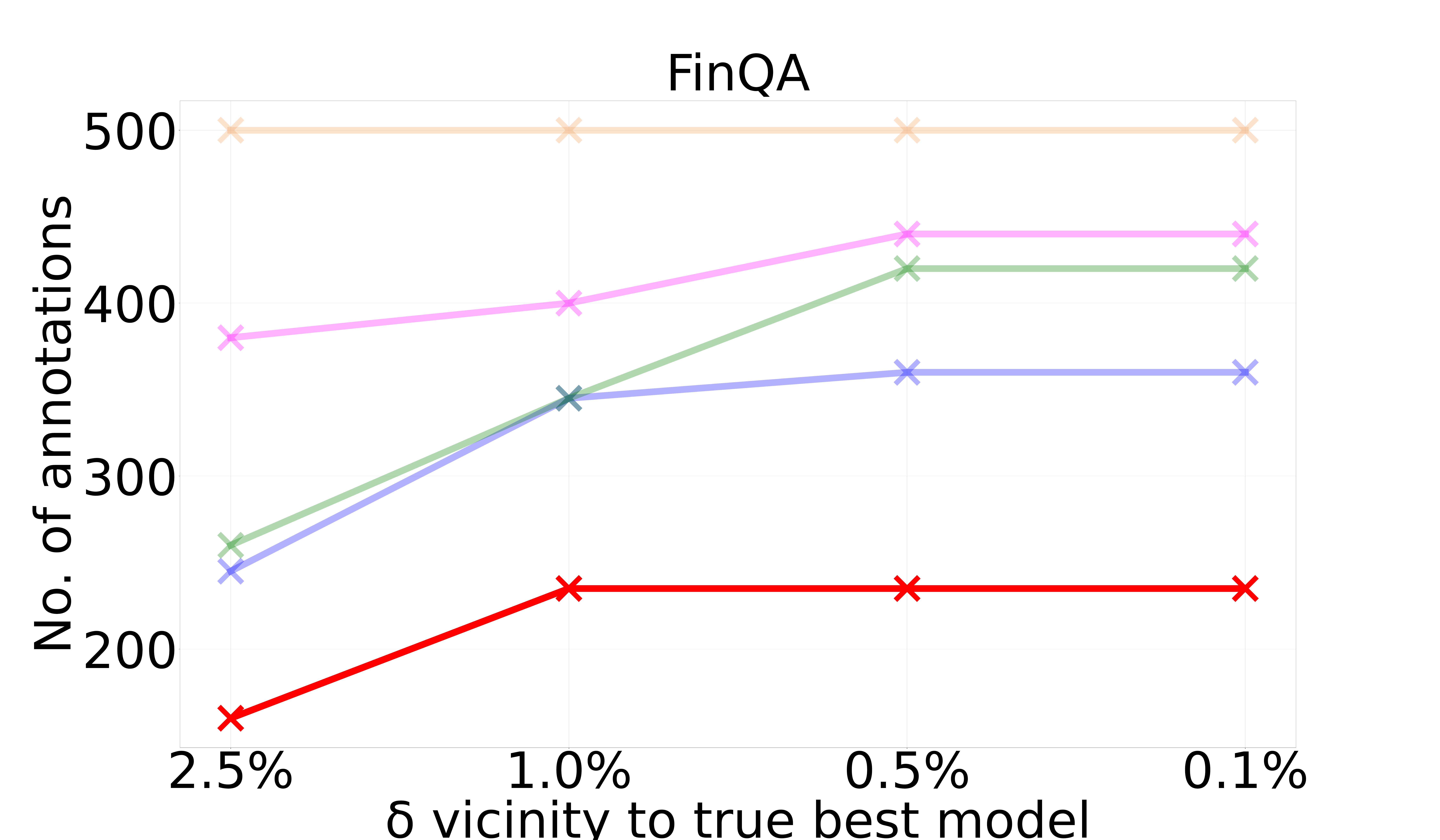}
\end{minipage}

\caption{Annotation efficiency of \framework and baselines across different $\delta$.}
\label{fig:gap_records_plots}
\end{figure}

\section{Ablation Study and Design Choice Analysis}
\label{app:ablation_study}

Looking at the ablation study as a whole, we can frame the results in a way that highlights both the strengths of the method and the rationale behind each design choice. Let’s take a look at the Comprehensive Arithmetic Problems dataset. First, all versions of the algorithm consistently outperform the baseline methods, which already indicates that the overall structure of the approach is sound and effective.

When we break things down further, the largest improvements clearly come from combining variance and entropy. This component has the strongest impact across the board, so emphasizing its contribution feels well-justified.

Regarding the adaptive learning rate, the full \framework reaches 100\% identification probability at around 196 annotations, whereas the version without adaptive LR needs about 202. The difference is small, but since we’re essentially comparing variants of our own method, even slight gains matter in an ablation study. The important point is that the adaptive LR generally leads to improvements. Eventually, we want to remark that the adaptive learning rate depends on the proxy of the variance of those annotations that are observed, thus performance gains also depend on the data distribution under analysis.

When evaluating generative language models, a wide range of metrics can be used. Traditional approaches rely on statistical measures such as BLEU and ROUGE, while more recent methods include sentence-embedding–based metrics and even LLM-as-a-judge evaluations. However, these metrics can become unreliable when the data comes from highly specialized domains. In such cases, general-purpose embedding models or LLM-based judges may fail to capture true semantic similarity, resulting in noisy or misleading scores. In practice, the choice of an appropriate metric is highly context-dependent and closely tied to the underlying data distribution. For example, in our experiments with medical datasets, several open source embedding models produced nearly constant similarity scores across sentences, effectively reducing the algorithms’ behavior to randomness. Identifying robust, domain-appropriate evaluation metrics is an active area of research for many generative tasks, but a detailed investigation of this topic lies beyond the scope of our current work.

\begin{figure}[t]
\centering
\includegraphics[height=3.2cm,keepaspectratio]{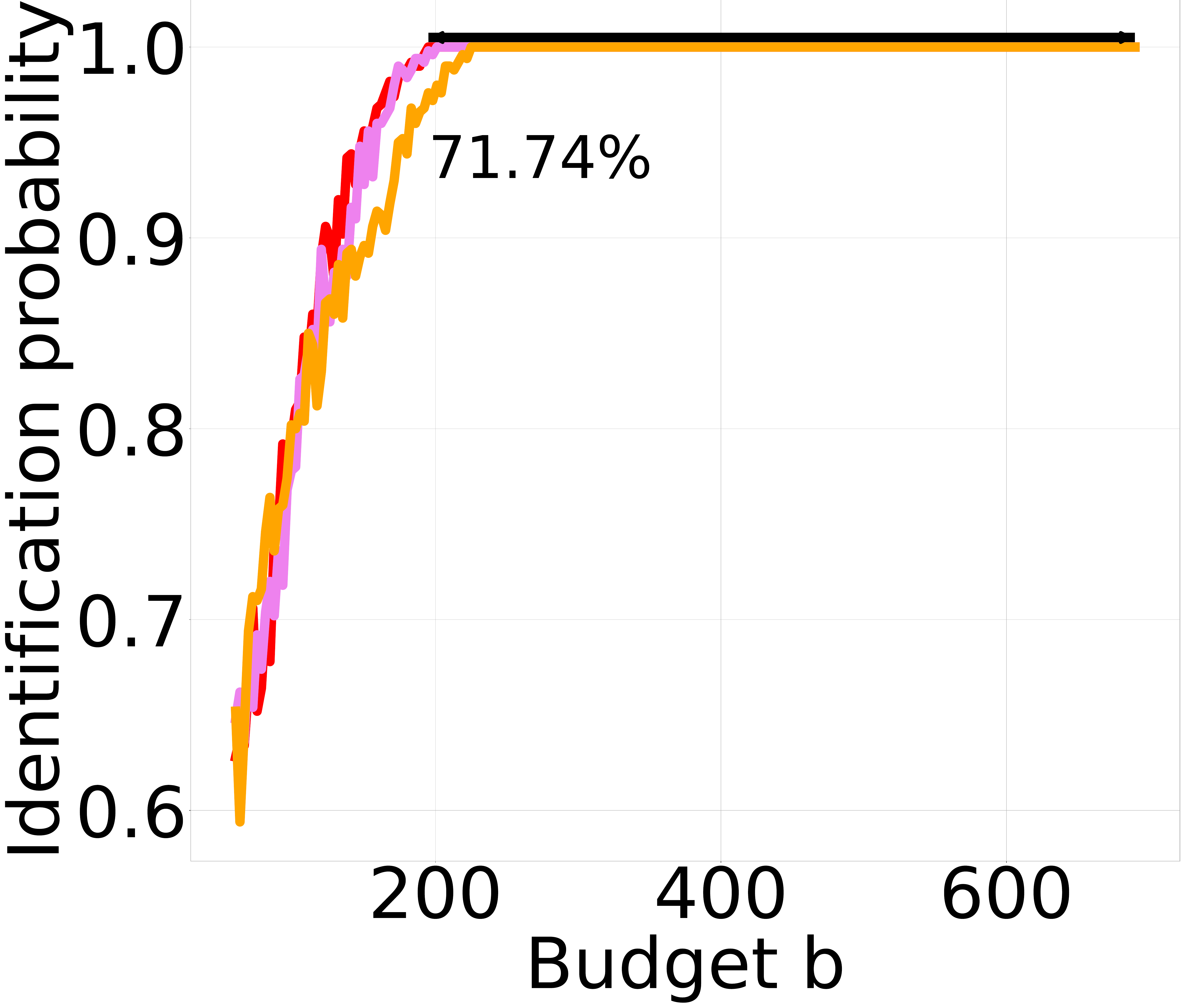}

\caption{Ablation study for different versions of \framework. We compare the identification probability of full \framework (\textit{red}) against the identification probabilities of variance-entropy \framework (\textit{pink}) and adaptive LR \framework (\textit{yellow}).}
\label{fig:ablation_identification}
\end{figure}

\section{Additional Details}\label{app:detailspt}
\subsection{Rationale for Metric Selection}
While more recent evaluation frameworks increasingly utilize Large Language Model (LLM) based metrics due to their higher correlation with human judgment, we intentionally select ROUGE and BERTScore for our evaluation pipeline. This decision is driven by three primary factors: stability, reproducibility, and computational feasibility.

First, standard metrics like ROUGE and BERTScore offer a level of stability and simplicity that remains foundational in generative model evaluation. In contrast, LLM-based evaluators, despite their sophistication, are often susceptible to significant variability, sensitivity to specific prompt phrasing, and inherent model biases. Our preliminary experiments with sentence-embedding-based similarity metrics revealed them to be considerably noisier and less stable across streaming batches. Specifically, we observed that embedding scores tended to cluster tightly with minimal spread, even when model outputs were meaningfully different. This suggests that such models may fail to reliably distinguish between genuinely similar and dissimilar responses in our specific context.

Second, the computational and financial overhead associated with LLM-based evaluators is substantial. Given that our methodology requires calculating pairwise distances across a large collection of language models ($m=54$), the quadratic nature of these comparisons makes LLM-based scoring impractical for large-scale or online evaluation. 

Finally, by utilizing established metrics, we ensure that our results are easily reproducible without requiring access to proprietary or high-cost APIs. While we recognize that evaluating our framework under advanced LLM-based metrics is a valuable direction for future research, the current selection provides a robust and efficient baseline for measuring model performance and hypothetical variance.

\subsection{Computational Overhead Analysis}
To estimate the hypothetical variance at each query step, our method requires evaluating pairwise distances between model outputs. For the LongForm--QA (Canonical) dataset, which represents the largest collection of models in our study ($m = 54$), we quantify the computational requirement as follows.

Given a prompt $x_t$, each model $i$ generates a response $f_i(x_t)$. To estimate pairwise similarity, we compute the distance $d(f_i(x_t), f_j(x_t))$ for all model pairs $(i, j)$. Since the distance metric is symmetric, where $d(f_i(x_t), f_j(x_t)) = d(f_j(x_t), f_i(x_t))$, we only need to evaluate the upper triangular portion of the distance matrix, excluding the diagonal. This results in:
\begin{equation}
\frac{m \times (m - 1)}{2} = \frac{54 \times 53}{2} = 1431 \text{ distinct pairwise comparisons.}
\end{equation}
Additionally, we compute the oracle similarity for each model output relative to the ground truth annotation $y_t$, denoted as $d(y_t, f_i(x_t))$, adding 54 evaluations per prompt. This brings the total number of distance computations per prompt to $1431 + 54 = 1485$.

In practice, this overhead is negligible. By preprocessing the dataset to compute all pairwise similarities before running the experiments, we significantly improve efficiency. On a local machine (Apple MacBook M3 Pro, 11-core CPU, 18GB unified memory), processing 1000 prompts requires an end-to-end time of 2 minutes and 36 seconds. This equates to approximately 0.156 seconds per prompt, a latency low enough to support real-time or interactive applications even when scaling to dozens of models.

\subsection{Posterior Distribution}
Let $\hat{L}_{t-1,i}$ be the cumulative loss estimate and $\eta_t$ the adaptive learning rate. The posterior distribution $\mathbf{p}_t=[p_{t, i}]_{i\in\mathcal{M}}$ is updated as:
\begin{equation}
    p_{t,i} = \frac{\exp{\left\{-\eta_t\hat{L}_{t-1,i}\right\}}}{\sum_{j\in\mathcal{M}}\exp{\left\{-\eta_t\hat{L}_{t-1,j}\right\}}}
\end{equation}

\end{document}